\documentclass{article}

\usepackage{PRIMEarxiv}

\usepackage[utf8]{inputenc} 
\usepackage[T1]{fontenc}    
\usepackage{hyperref}       
\usepackage{url}            
\usepackage{booktabs}       
\usepackage{amsmath}
\usepackage{amsfonts}       
\usepackage{nicefrac}       
\usepackage{microtype}      
\usepackage{graphicx}
\usepackage{subcaption}
\usepackage{xcolor}
\usepackage{listings}
\usepackage{float}
\usepackage{placeins}
\usepackage{tikz}
\usepackage{enumitem}
\graphicspath{{./}{assets/}{assets/logo/}}
\definecolor{codebg}{RGB}{248,248,248}
\definecolor{codekw}{RGB}{0,92,175}
\definecolor{codecom}{RGB}{0,128,0}
\definecolor{codestr}{RGB}{163,21,21}

\lstdefinestyle{cleanpseudo}{
  backgroundcolor=\color{codebg},
  basicstyle=\ttfamily\small,
  keywordstyle=\color{codekw}\bfseries,
  commentstyle=\color{codecom},
  stringstyle=\color{codestr},
  numbers=left,
  numberstyle=\tiny,
  numbersep=8pt,
  frame=single,
  breaklines=true,
  showstringspaces=false,
  tabsize=2
}

\newcommand{\roundedtitlelogo}[1]{%
  \tikz[baseline=(logo.base)] \node[rounded corners=3pt, inner sep=0pt, clip] (logo)
  {\includegraphics[width=1.1cm,height=1.1cm]{#1}};%
}

\newcommand{\roundedinlinelogo}[2][1.1em]{%
  \tikz[baseline=(logo.base)] \node[rounded corners=1.5pt, inner sep=0pt, clip] (logo)
  {\includegraphics[height=#1]{#2}};%
}

\title{\raisebox{-0.30\height}{\roundedtitlelogo{\detokenize{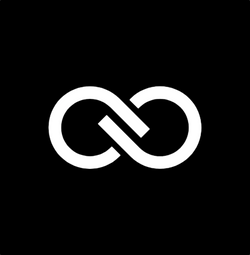}}}\hspace{0.55em}Nucleus-Image: Sparse MoE for Image Generation
}

\author{Nucleus AI Team}
\date{}

\begin{document}
\maketitle

\begin{abstract}
  We present Nucleus-Image, a text-to-image generation model that establishes a new Pareto frontier in quality-versus-efficiency by matching or exceeding leading models on GenEval, DPG-Bench, and OneIG-Bench while activating only approximately 2B parameters per forward pass. Nucleus-Image employs a sparse mixture-of-experts (MoE) diffusion transformer architecture with Expert-Choice Routing that scales total model capacity to 17B parameters across 64 routed experts per layer. We adopt a streamlined architecture optimized for inference efficiency by excluding text tokens from the transformer backbone entirely and using joint attention that enables text KV sharing across timesteps. To improve routing stability when using timestep modulation, we introduce a decoupled routing design that separates timestep-aware expert assignment from timestep-conditioned expert computation. We construct a large-scale training corpus of 1.5B high-quality training pairs spanning 700M unique images through multi-stage filtering, deduplication, aesthetic tiering, and caption curation. Training follows a progressive resolution curriculum (256 to 512 to 1024) with multi-aspect-ratio bucketing at every stage, coupled with progressive sparsification of the expert capacity factor. We adopt the Muon optimizer and share our parameter grouping recipe tailored for diffusion models with timestep modulation. Combined with a Warmup-Stable-Merge learning rate schedule, this eliminates the need for an EMA shadow copy of model weights. Nucleus-Image demonstrates that sparse MoE scaling is a highly effective path to high-quality image generation, reaching the performance of models with significantly larger active parameter budgets at a fraction of the inference cost. These results are achieved without post-training optimization of any kind: no reinforcement learning, no direct preference optimization~\cite{rafailov2023direct}, and no human preference tuning. We release the full model weights, training code, and dataset to the community, making Nucleus-Image the first fully open-source MoE diffusion model at this quality tier.\\[0.8em]
  \raisebox{-0.2em}{\roundedinlinelogo{\detokenize{assets/logo/OpsAI_Logo.png}}}~\href{https://withnucleus.ai/image}{https://withnucleus.ai/image}\\[0.3em]
  \raisebox{-0.2em}{\roundedinlinelogo{\detokenize{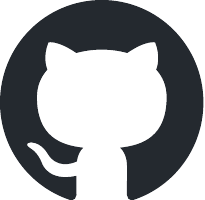}}}~\href{https://github.com/WithNucleusAI/Nucleus-Image}{https://github.com/WithNucleusAI/Nucleus-Image}\\[0.3em]
  \raisebox{-0.2em}{\includegraphics[height=1.1em]{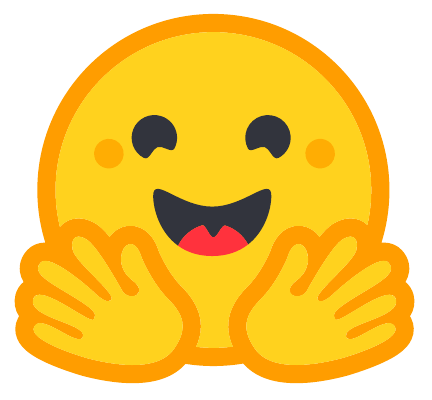}}~\href{https://huggingface.co/NucleusAI/NucleusMoE-Image}{https://huggingface.co/NucleusAI/NucleusMoE-Image}
\end{abstract}

\newpage
\tableofcontents
\newpage

\begin{figure}[H]
    \centering
    \includegraphics[width=\linewidth]{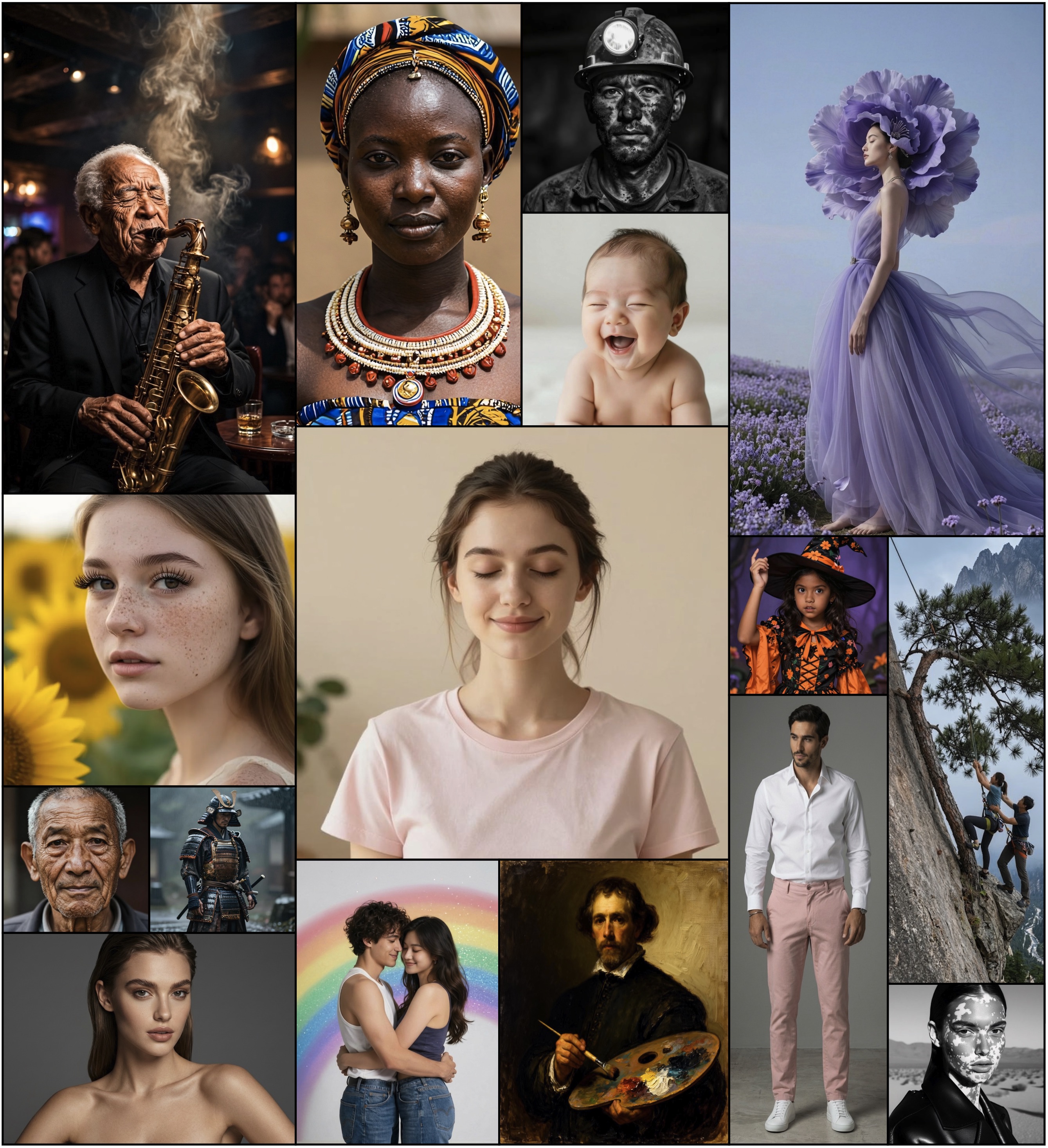}\vspace{-2pt}
    \includegraphics[width=\linewidth]{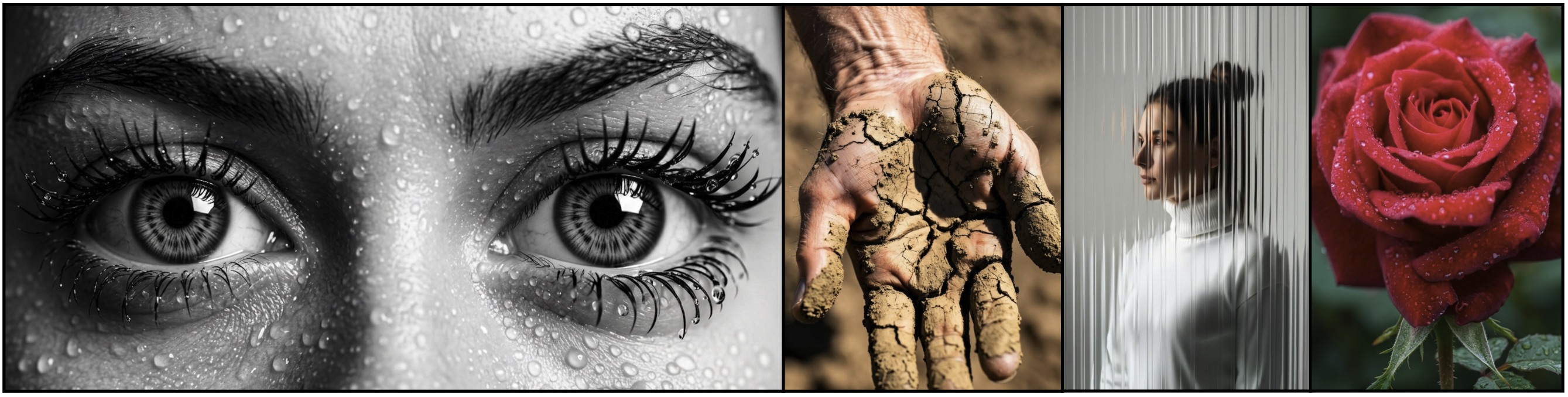}
    \caption{Nucleus-Image generations of human subjects and portraits, spanning diverse cultures, ages, and artistic styles. From expressive character studies to fine-grained close-ups with intricate skin texture and detail.}
    \label{fig:collage-1}
\end{figure}
\newpage

\begin{figure}[H]
    \centering
    \includegraphics[width=\linewidth]{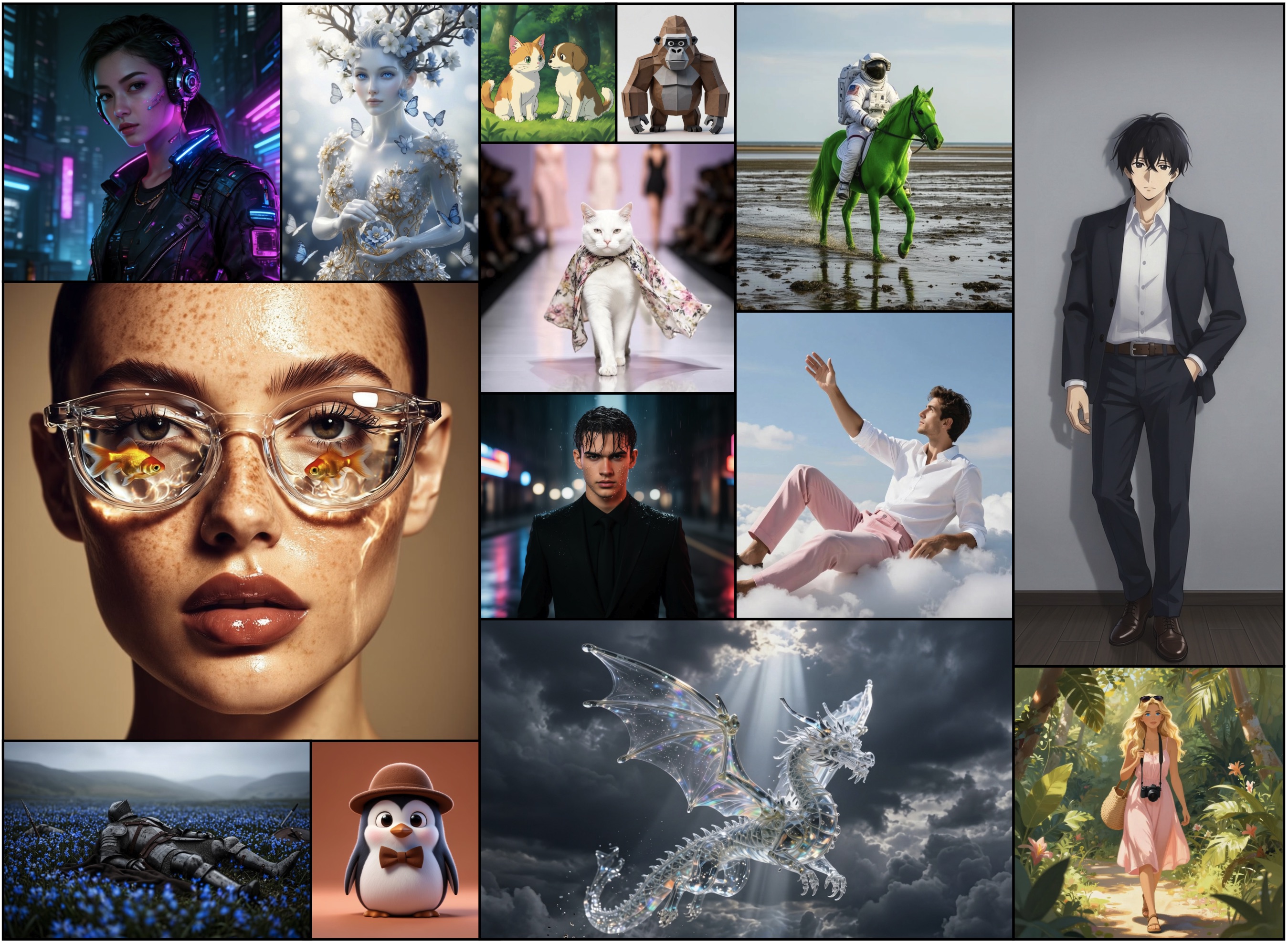}\vspace{-2pt}
    \includegraphics[width=\linewidth]{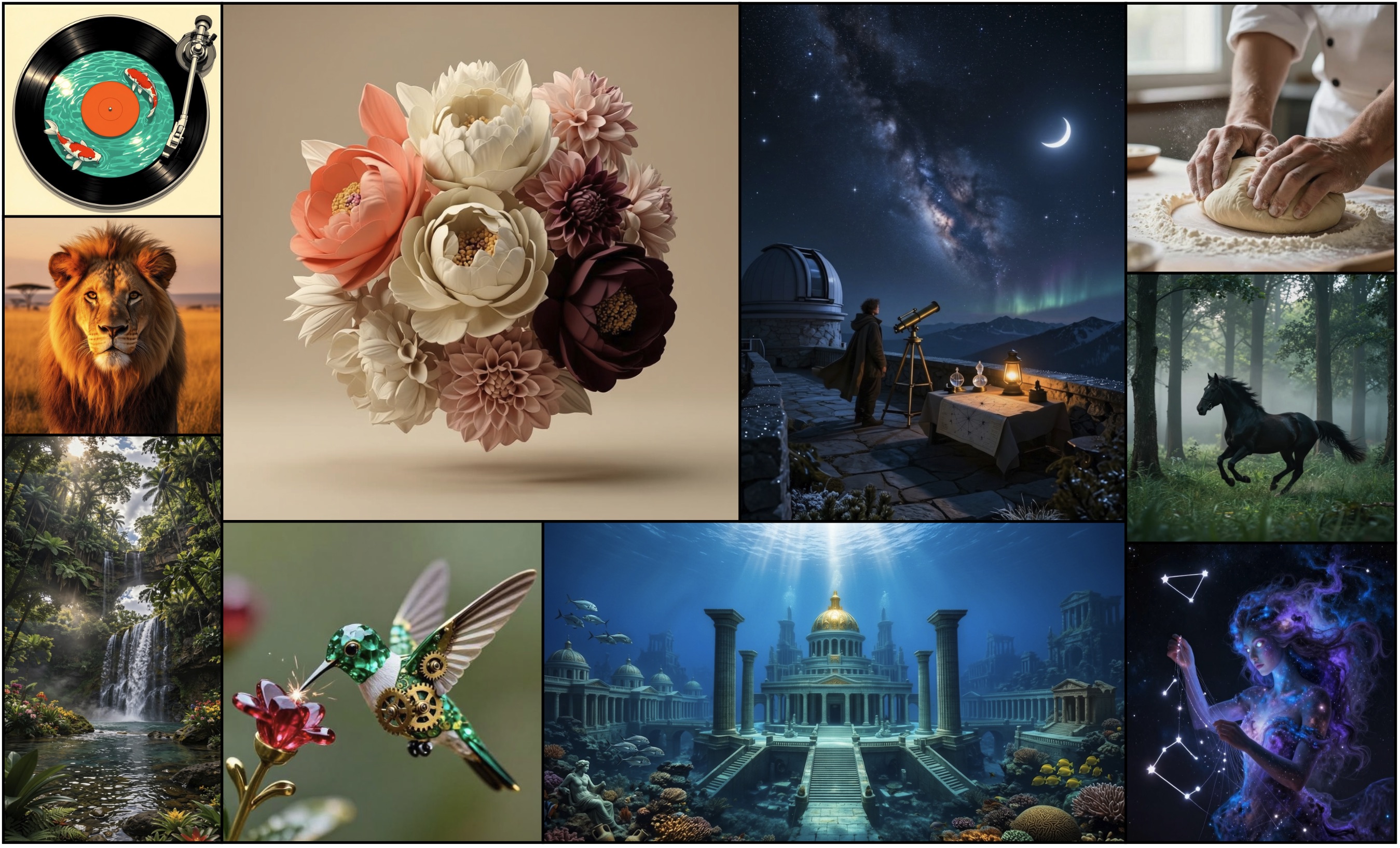}
    \caption{Nucleus-Image generations spanning fantasy, surrealism, animation, and the natural world.}
    \label{fig:collage-2}
\end{figure}
\newpage

\begin{figure}[H]
    \centering
    \includegraphics[width=\linewidth]{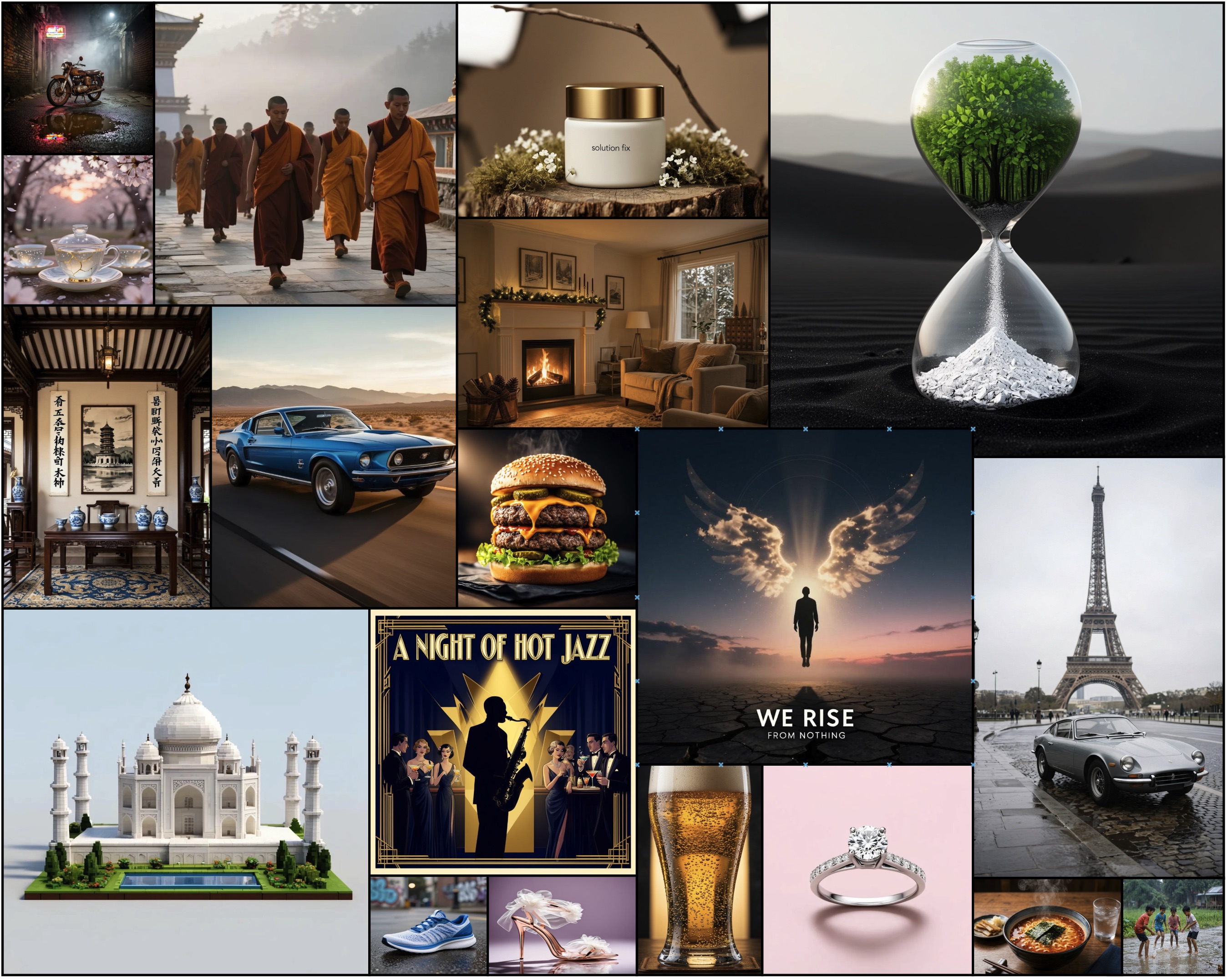}\vspace{-2pt}
    \includegraphics[width=\linewidth]{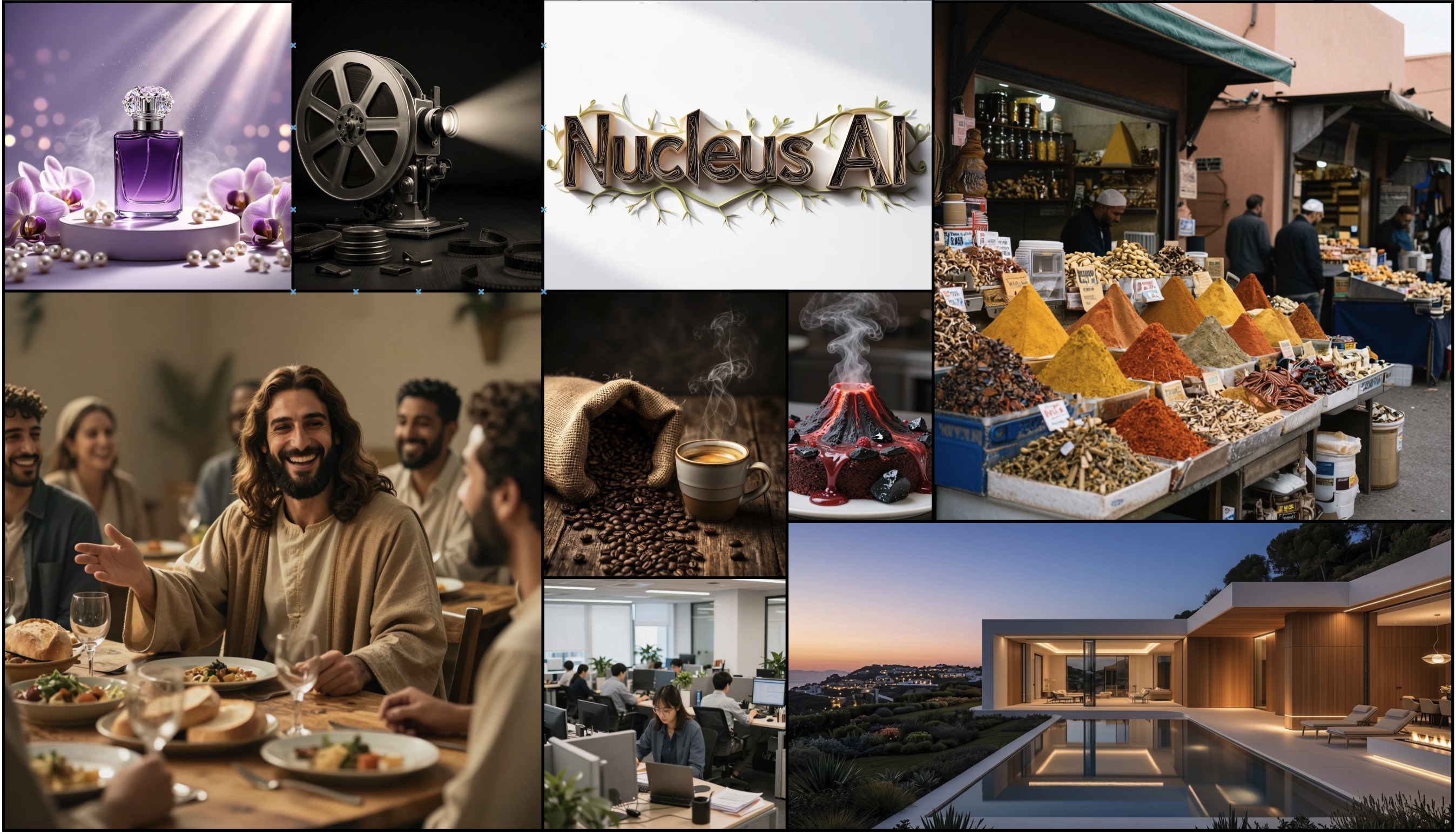}
    \caption{Nucleus-Image generations across product photography, architecture, typography, food, and world culture demonstrating versatility in commercial, conceptual, and everyday imagery.}
    \label{fig:collage-3}
\end{figure}
\newpage

\begin{figure}[H]
  \centering
  \includegraphics[width=\linewidth,keepaspectratio]{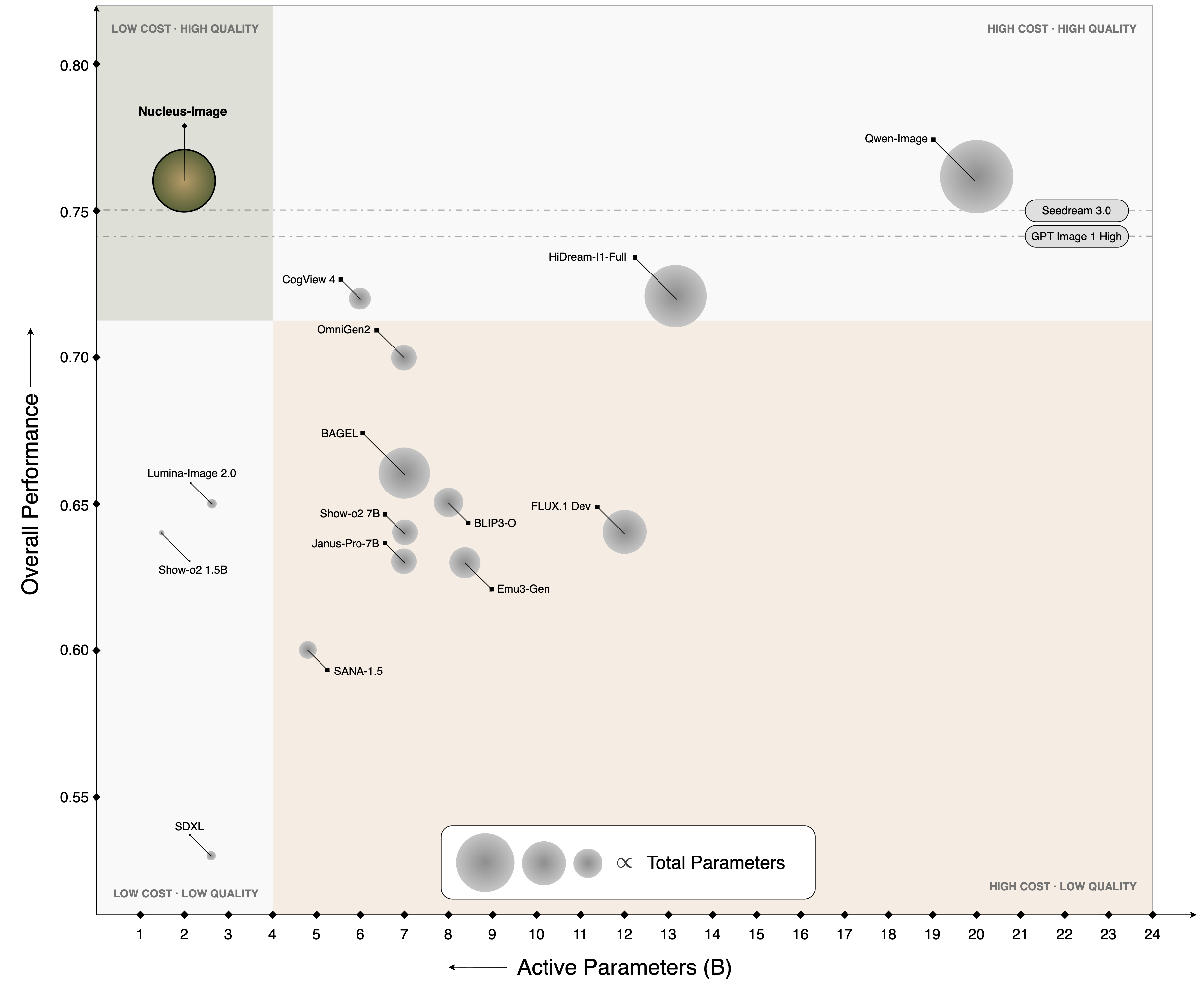}
  \caption{Overall performance computed as the average of GenEval, DPG-Bench, and OneIG-Bench benchmark scores}
  \label{fig:overall-performance}
\end{figure}

\section{Introduction}
Nucleus-Image targets high-quality text-to-image generation while keeping compute and memory costs tractable. The model adopts sparse MoE layers so that only a fraction of parameters are active per token, and uses expert-choice routing to keep expert utilization balanced without large auxiliary load-balancing losses. The training stack pairs a large curated dataset with rectified flow supervision and a multi-stage curriculum that emphasizes higher-quality data later in training. The remainder of this report describes the data pipeline, MoE kernel design, model architecture, routing and optimization choices, distributed training setup, and inference-time optimizations.

\section{Data Pipeline}
We build the Nucleus-Image training corpus from web-sourced candidates that are progressively validated, deduplicated, scored, captioned, and partitioned into a final curated mixture of around 700M images. Each retained image stores caption candidates at varying granularities, producing roughly 1.5B image-caption supervision pairs. The pipeline is optimized for rapid iteration along two axes: increasing the supply of high-quality images and increasing the reliability and diversity of textual supervision.

At a high level, the dataset is a metadata log coupled to a large asset store. Each training row references a pixel asset, a set of caption candidates with explicit provenance, and a small set of derived control signals (image quality tiers, caption alignment and provenance, curricula, and task routing) that deterministically map the row into training mixtures used by the sampler.

\textbf{Terminology.} In this section, \emph{quality tiers} refer to discretized bands of continuous quality scores. \emph{Episodic buckets} refer to the \(K=8\) curriculum partitions consumed progressively during training, while \emph{aspect-ratio buckets} refer to training-time batch groupings keyed by target crop shape. The remainder of this section describes how we produce these rows and the signals attached to them.

\begin{figure}[t]
  \centering
  \includegraphics[width=0.92\linewidth]{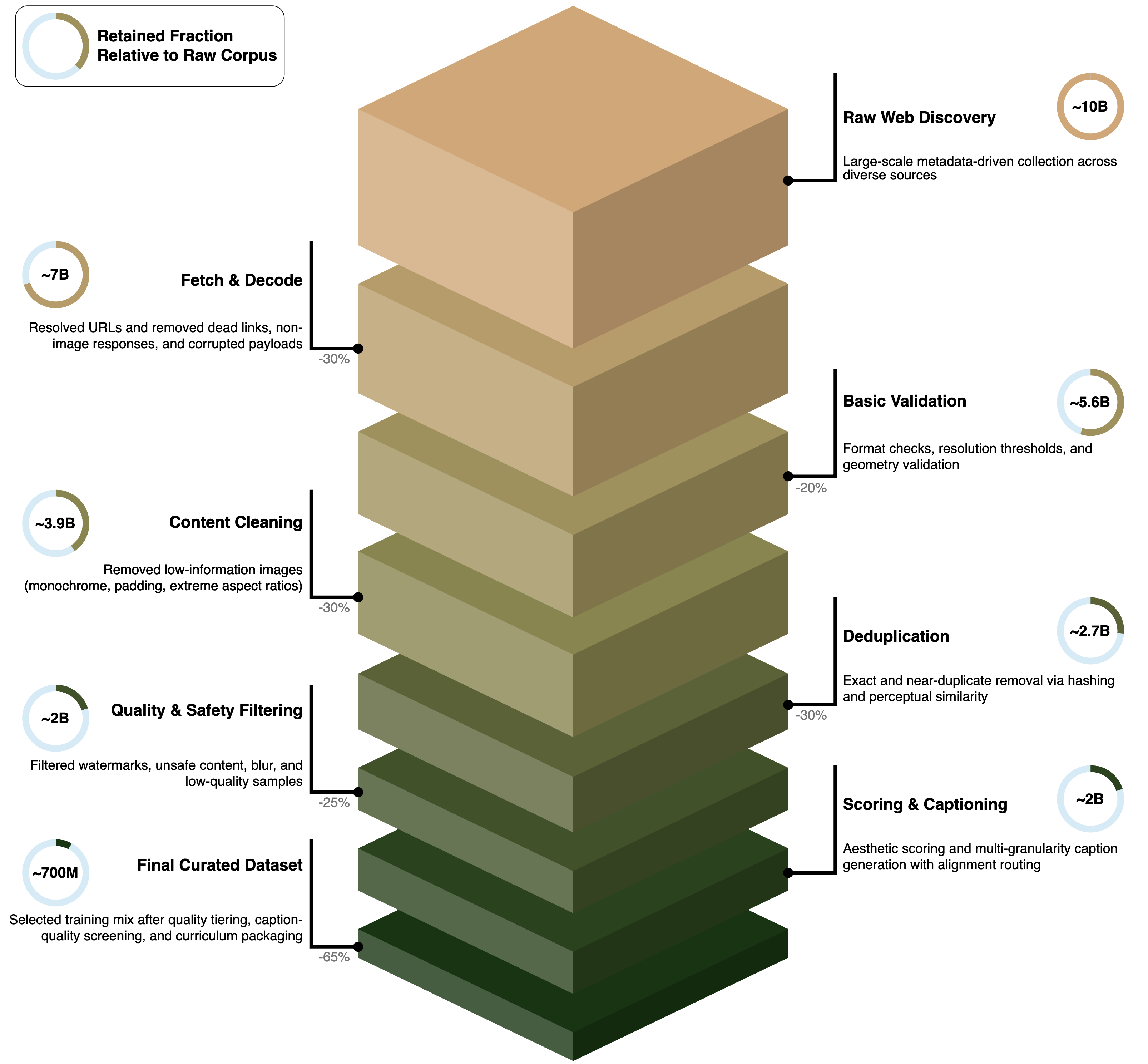}
  \caption{Dataset retention across the data pipeline. Block height indicates retained corpus size, and ring markers indicate retained fraction relative to the raw corpus.}
  \label{fig:data-funnel}
\end{figure}

Figure~\ref{fig:data-funnel} provides a high-level view of dataset retention across the pipeline. Starting from large-scale web discovery, the corpus is progressively reduced by fetch-time failures, basic validity checks, heuristic cleaning, deduplication, and quality-and-safety filtering before entering scoring, caption enrichment, and final curriculum-aware curation. The subsequent sections describe these stages in detail and introduce the metadata signals that control retention and downstream sampling.

\subsection{Collection Pipeline}
\textbf{Web crawling and storage.} We assembled the raw corpus by crawling public web sources at scale following the metadata-driven collection strategy popularized by CLIP and subsequent work \cite{radford2021learning,metaclip}. We resolved, fetched, and filtered candidate image URLs to remove dead links and non-image responses, stored the surviving assets in Google Cloud Storage (GCS), and indexed them via a crawl manifest. Each ingested image is assigned a stable content-addressed key and a corresponding metadata row that records its storage locator, source URL, original geometry, and a content hash. We deliberately decouple pixel storage from metadata early in the pipeline. Images reside in object storage as opaque blobs, while all downstream signals (geometry, quality scores, captions, provenance, and routing tags) are accumulated in the metadata layer. This separation enables rapid iteration on annotation models and filtering policies without re-downloading or re-decoding pixels.

\textbf{Metadata representation.} The training dataloader consumes a flat Parquet metadata table with a consistent row schema (Table~\ref{tab:dataset-core-schema}). We favor a columnar representation over a relational schema to maximize predicate pushdown and enable large-scale analytics. Dataset versions are published as immutable snapshots plus per-run manifests that select eligible rows and columns.

\begin{table}[H]
  \caption{Core training-facing metadata schema.}
  \centering
  \begin{tabular}{lll}
    \toprule
    Field                           & Type         & Purpose                                                            \\
    \midrule
    \texttt{id}                     & string       & Stable record identifier for tracing and debugging.                \\
    \texttt{media\_path}            & string       & Content-addressed locator in object storage.                       \\
    \texttt{width}, \texttt{height} & int          & Original geometry; used for aspect-ratio bucketing and validation. \\
    \texttt{captions}               & list[string] & Candidate captions at multiple granularities.                      \\
    \texttt{caption\_sources}       & list[string] & Provenance tag per caption, aligned with \texttt{captions}.        \\
    \texttt{caption\_lengths}       & list[int]    & Token counts per caption.                                          \\
    \texttt{media\_source}          & string       & Coarse origin label.                                               \\
    \texttt{aesthetic\_score}       & float        & Real-image aesthetic score; nullable for synthetic samples.        \\
    \texttt{quality\_tier}          & int          & Static A1-A5 quality tier used for sampling.                       \\
    \texttt{episodic\_bucket}       & int          & Deterministic B1–B8 curriculum label.                              \\
    \texttt{sha256}                 & string       & Content hash for exact deduplication and integrity verification.   \\
    \bottomrule
  \end{tabular}
  \label{tab:dataset-core-schema}
\end{table}

\textbf{Schema versioning and invariants.} Metadata acts as the control plane for training and is published as immutable snapshots plus per-run manifests that select the rows and columns eligible for a given run. This design makes regressions debuggable: when metrics shift, we can attribute changes to a specific annotation stage or filter policy without re-downloading or re-decoding pixels.

The training loader enforces simple row-level invariants on every eligible example. Geometry must be present and positive. Caption arrays must be aligned (\texttt{captions}, \texttt{caption\_sources}, and \texttt{caption\_lengths} have the same length). Schema evolution must preserve required column names and types so that columnar predicate pushdown remains valid across dataset versions.

\subsection{Cleaning and Validation}
Cleaning is applied immediately after crawling, before any expensive annotation work. Validation is tiered to reject failures cheaply on CPU and reserve GPU compute for samples likely to survive.

\textbf{Tier 1 (CPU, cheap).} The first tier performs format decoding, geometry validation, and header/magic-byte consistency checks, reflecting prior practical guidance that careful input sanitization improves training stability in image generation pipelines \cite{hinz2020}. Images that fail to decode in any supported format (JPEG, PNG, WebP), report zero or negative dimensions, or whose declared content-type disagrees with the file header are rejected outright. This tier also drops images below a minimum resolution threshold, as sub-threshold images contribute negligible training signal at target generation resolutions.

\textbf{Tier 2 (CPU, moderate).} The second tier applies histogram-based sanity checks, orientation normalization, and lightweight content-level filters. Near-monochrome images (solid black, solid white, or near-zero variance across channels), images with excessive letterboxing or padding, and images with extreme aspect ratios are discarded. EXIF orientation metadata is normalized at this stage to prevent a single scene from appearing as multiple distinct training modes due to rotation tags.

\textbf{Tier 3 (GPU, expensive).} The final tier runs GPU-accelerated filters for safety classification, watermark detection, blur detection via frequency-domain metrics, and embedding-based quality scoring. Images flagged as unsafe, heavily watermarked, or severely degraded are removed. Surviving real images receive a continuous aesthetic quality score (Section~\ref{subsec:quality-scoring}), and image-text alignment scores (Section~\ref{subsec:image-text-alignment}) are computed for samples with candidate captions and written back to metadata.

We orchestrate the GPU-resident stages of this pipeline using NVIDIA DALI \cite{dali} for accelerated image decode and batch transforms, and leverage NVIDIA NeMo Curator~\cite{nemocurator} for scalable GPU-accelerated inference across the heavy scoring passes like CLIP-based embedding, safety classification, and aesthetic quality heads. This combination avoids repeated CPU-to-GPU transfers and lets the Tier~3 filters operate on already-resident tensors while NeMo Curator manages distributed execution of the embedding and scoring fleet. In a representative production deployment, the ingestion pipeline sustained throughput in the order of \(10^3\) images/sec across 8\,\(\times\) H100 nodes, with the bottleneck shifting from raw download bandwidth to GPU-bound quality scoring and captioning as scale increased.
\subsection{Deduplication}
Duplicates are a dominant failure mode in web-scale corpora: repeated supervision reduces effective data diversity and increases the risk of overfitting in late-stage training \cite{ren2025}. We apply a two-stage deduplication strategy.

First, exact duplicates are eliminated via SHA-256 content hashes computed during ingestion. Second, near-duplicates are identified using perceptual hashing (pHash) with a conservative Hamming radius; within each cluster, a single representative is retained based on resolution and quality score. This setup is aligned with recent practice in memorization mitigation and large-scale dataset curation \cite{zhang2025,sugiura2025}. In a typical crawl-derived subset, exact-hash deduplication removes approximately 3-5\% of records, and near-duplicate clustering removes an additional \textasciitilde8\%, yielding an improvement in effective diversity per unit compute.

\subsection{Quality Scoring and Tiering}
\label{subsec:quality-scoring}

Validity checks alone do not produce a useful training set. Each surviving sample is assigned one or more quality signals that serve either as hard filters or as soft training-curriculum axes. Real images receive an explicit aesthetic score, while synthetic images are ranked using non-aesthetic signals such as caption quality, caption length, and provenance.

\textbf{Aesthetic scoring.} For the real-image subset, a lightweight regression head on top of a frozen image encoder predicts a scalar aesthetic quality score, calibrated against human preference annotations in the spirit of prior work on aesthetic scoring for generative models \cite{taylor2026}. The score is stored in metadata and used as one input into downstream quality tiering. In a representative run, aesthetic scoring on real images retained approximately \textasciitilde580M samples above a floor of 5.0; these were then combined with approximately \textasciitilde115M synthetic images routed through non-aesthetic quality heuristics, yielding a final tiered corpus of roughly 700M images. Tier assignments (A1-A5) are then used to shape the training distribution (Section~\ref{subsec:curriculum}) rather than selecting a single "best" slice. This late-stage reduction corresponds to the bottom of the retention funnel in Figure~\ref{fig:data-funnel}, where large-scale crawl candidates are converted into the final tiered training mixture.

\begin{figure}[t]
  \centering
  \includegraphics[width=\linewidth]{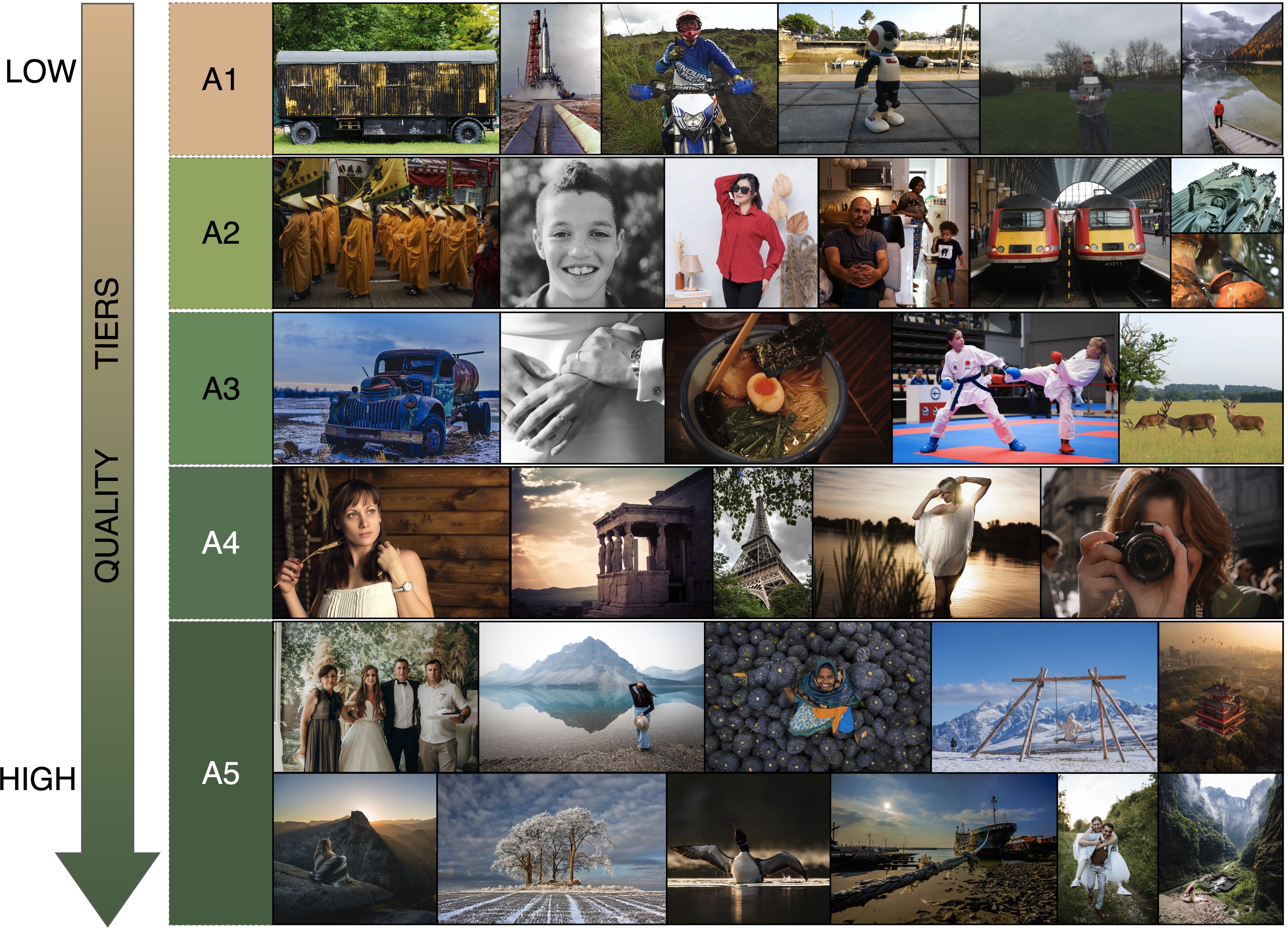}
  \caption{Representative quality tiers for real images. Real-image samples are ranked using aesthetic scoring together with downstream quality signals, then grouped into the shared A1-A5 quality taxonomy used by the sampler.}
  \label{fig:quality-tiers-real}
\end{figure}

Figure~\ref{fig:quality-tiers-real} shows representative examples from the A1-A5 ladder for real images. The progression runs from marginal or weakly composed samples to visually strong photographs, illustrating how aesthetic scoring contributes to the real-image quality taxonomy.

\begin{figure}[t]
  \centering
    \includegraphics[width=\linewidth]{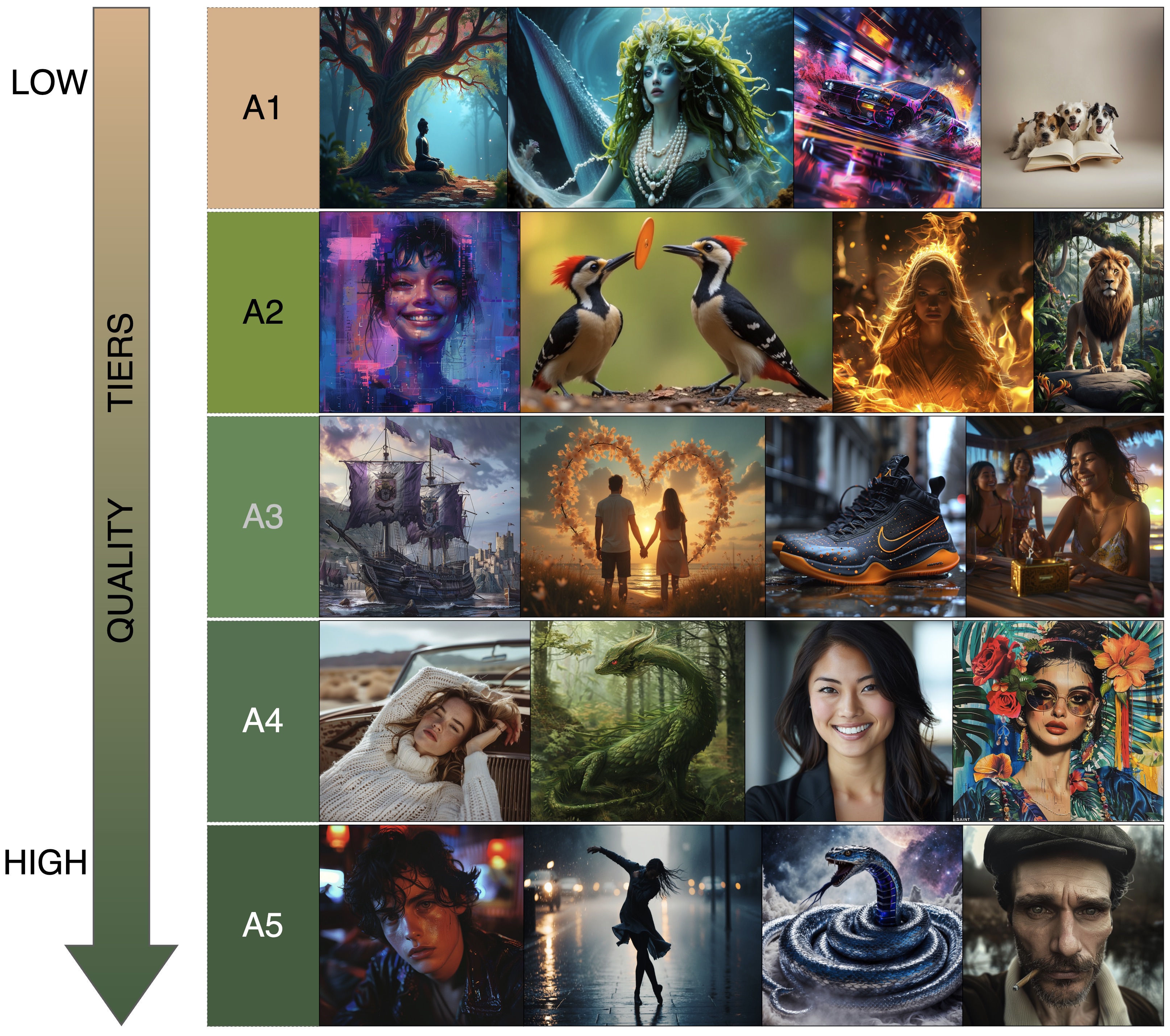}
  \caption{Representative quality tiers for synthetic images. Synthetic samples bypass aesthetic scoring and are assigned to the same A1-A5 taxonomy using non-aesthetic quality heuristics}
  \label{fig:quality-tiers-syn}
\end{figure}

Figure~\ref{fig:quality-tiers-syn} shows the corresponding A1-A5 ladder for synthetic images. Although synthetic samples do not receive aesthetic scores directly, they are mapped into the same quality-tier framework using non-aesthetic heuristics such as caption quality, caption length, provenance, and related metadata signals.

\textbf{Hard quality filters.} In addition to scalar aesthetic scoring and downstream quality tiering, binary drop filters target failure modes that are poorly captured by a single scalar score. These include images with prominent watermarks or site-specific overlays, compression artifacts, and images whose dominant content is a border, frame, or UI chrome rather than a scene. These filters are applied in the Tier 3 GPU stage.

\textbf{Composite quality slices.} After scoring and filtering, each image is assigned to a composite \emph{quality slice} derived from its image quality tier, resolution tier, and image-text alignment score. Even the lowest-quality slice enforces a minimum image-quality threshold, ensuring that no training data falls below a baseline perceptual standard. The highest-quality slice contains images that are high-resolution and high-quality, paired with well-aligned captions, and receives the largest sampling weight during late-stage training.

\subsection{Caption Curation}
\label{subsec:image-text-alignment}

Textual supervision is the primary signal for a text-to-image model. We maintained multiple candidate captions per image with explicit provenance tags rather than collapsing each image to a single caption. This enabled post-hoc selection, analysis, and mixing at training time.

\textbf{Multi-granularity captioning.} The dataset is captioned at three levels of granularity: short (concise single-sentence descriptions), medium (paragraphs capturing salient objects, actions, and spatial relationships), and detailed (comprehensive descriptions including background, style, mood, and fine-grained attributes). Each granularity is produced by a dedicated captioning pass, and all variants are stored as separate entries in the \texttt{captions} list with corresponding \texttt{caption\_sources} provenance tags. This strategy is motivated by recent findings that caption detail significantly impacts generation quality. Lumina-Image 2.0 shows that more precise and detailed captions accelerate training convergence and effectively increase model capacity without additional parameters, while Playground-v3 demonstrates that training with captions at multiple levels of detail enables a richer linguistic concept hierarchy.\cite{lumina-image-2,playground-v3}

\textbf{Image-Text alignment routing.} For images that arrive with existing web-scraped alt-text or metadata captions, an image-text alignment score \(s\) (CLIP-based image-text alignment score) is computed and used to decide whether the original text is suitable for training. A three-way routing policy is applied:
\begin{itemize}
  \item Preserve: if \(s > 0.65\), keep the original caption as a candidate.
  \item Refine: if \(0.30 \leq s \leq 0.65\), generate an improved caption conditioned on the original text and image.
  \item Synthesize: if \(s < 0.30\), generate a new caption from pixels and provenance-only context.
\end{itemize}
Thresholds are calibrated against a manually labeled validation slice, ensuring that high-quality web captions are preserved (avoiding unnecessary re-captioning) while low-alignment alt-text is replaced. These caption-derived signals also feed back into corpus ranking and tiering, especially for synthetic samples where aesthetic scores are unavailable.

\textbf{Captioning models and web-context conditioning.} We implement captioning as an ensemble routing problem rather than relying on a single captioner. A fast, general-purpose captioner handles the bulk of generic imagery. Higher-precision VLMs are reserved for domains where hallucination risk is elevated (diagrams, scientific figures, OCR-heavy content, and images containing embedded text). When an image arrives with accompanying web text (e.g., alt-text or other crawl-time captions), we treat this text as provenance-bearing context rather than ground truth. For samples routed to \emph{refine} (moderate image-text alignment), we condition the recaptioner on both the pixels and the available web text so that names, specific entities, and contextual details are preserved while the caption is expanded and better grounded in visual evidence. The original web caption is retained as a separate candidate with its own provenance tag. Refinement captions are appended with provenance indicating they were conditioned on crawl-time text.

\subsection{Synthetic text rendering}

Accurate text rendering is a well-known challenge for diffusion-based image generation models. To improve the model's ability to produce legible in-image text across languages and layouts, we incorporated a dedicated synthetic text-data pipeline that generates images containing rendered text with controlled variations, similar to the multi-stage text rendering strategies used in Qwen-Image.\cite{qwen-image}

The pipeline renders text onto clean background images or templates using a variety of fonts, sizes, colors, layouts, and languages. To maximize diversity, we vary typographic parameters (font family, weight, kerning, alignment), visual parameters (color, opacity, shadow, outline), and compositional parameters (text placement, curvature, perspective distortion). A curriculum strategy progresses from simple single-word renders to multi-line paragraph-level compositions, and from common Latin scripts to multilingual text including logographic writings. The resulting synthetic images are captioned with ground-truth text content and added to the training set with appropriate provenance tags and controllable sampling weights. This text-rendering stream is a targeted subset of the broader synthetic corpus rather than the entirety of synthetic data.

\subsection{Dataset distribution and curriculum}
\label{subsec:curriculum}

The curated dataset exposes orthogonal axes of control over the training distribution: geometry (via aspect-ratio bucketing), quality (via static quality tiers and sampling weights), caption granularity (via provenance-based selection over short/medium/detailed captions), and supervision type (via a weighted task sampler across text-to-image and auxiliary tasks). Each axis is represented as metadata fields via sampler policies, allowing schedules to be updated without rewriting the pixel store.

\begin{figure}[t]
  \centering
  \includegraphics[width=\linewidth]{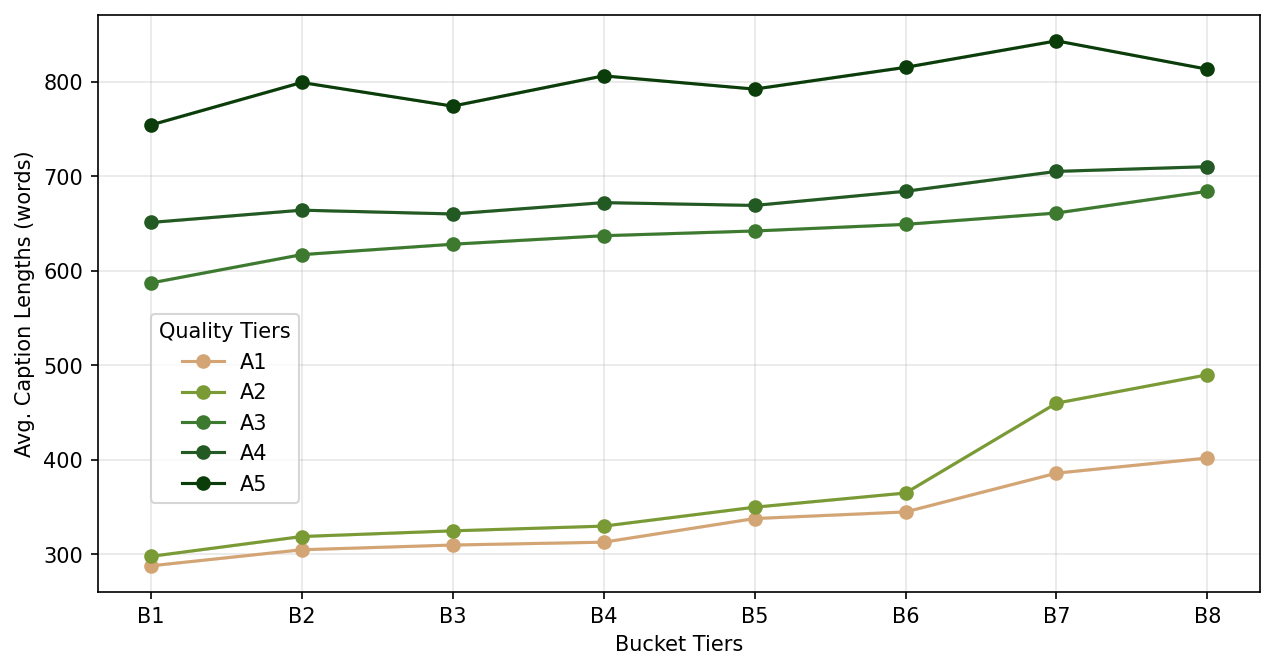}
  \caption{Average caption length across quality tiers and episodic buckets. Higher quality tiers generally carry longer captions, and later buckets trend toward richer textual supervision.}
  \label{fig:caption-lengths}
\end{figure}

Figure~\ref{fig:caption-lengths} shows that caption length varies systematically across both static quality tiers and episodic curriculum buckets. Higher-quality tiers generally carry longer captions, while later buckets skew toward richer textual supervision, making caption length a useful secondary signal in curriculum design. Caption lengths are measured in tokens using the model's training tokenizer, ensuring that length-based analyses reflect the effective text budget seen during training.

We implement curriculum learning by assigning each example to one of \(K=8\) \emph{episodic buckets} derived from a composite curriculum score that combines image quality tier and resolution tier. In practice, we compute a scalar curriculum score per image, rank the corpus by this score, and partition it into eight buckets from broad/early supervision to high-fidelity/late supervision. Training then progressively shifts sampling mass toward later buckets while maintaining a small allocation to earlier buckets to preserve diversity. Episodic buckets are materialized as deterministic metadata predicates via \texttt{episodic\_bucket} field, allowing schedules and mixture weights to be updated without rewriting the pixel store.

\begin{figure}[t]
  \centering
  \includegraphics[width=\linewidth]{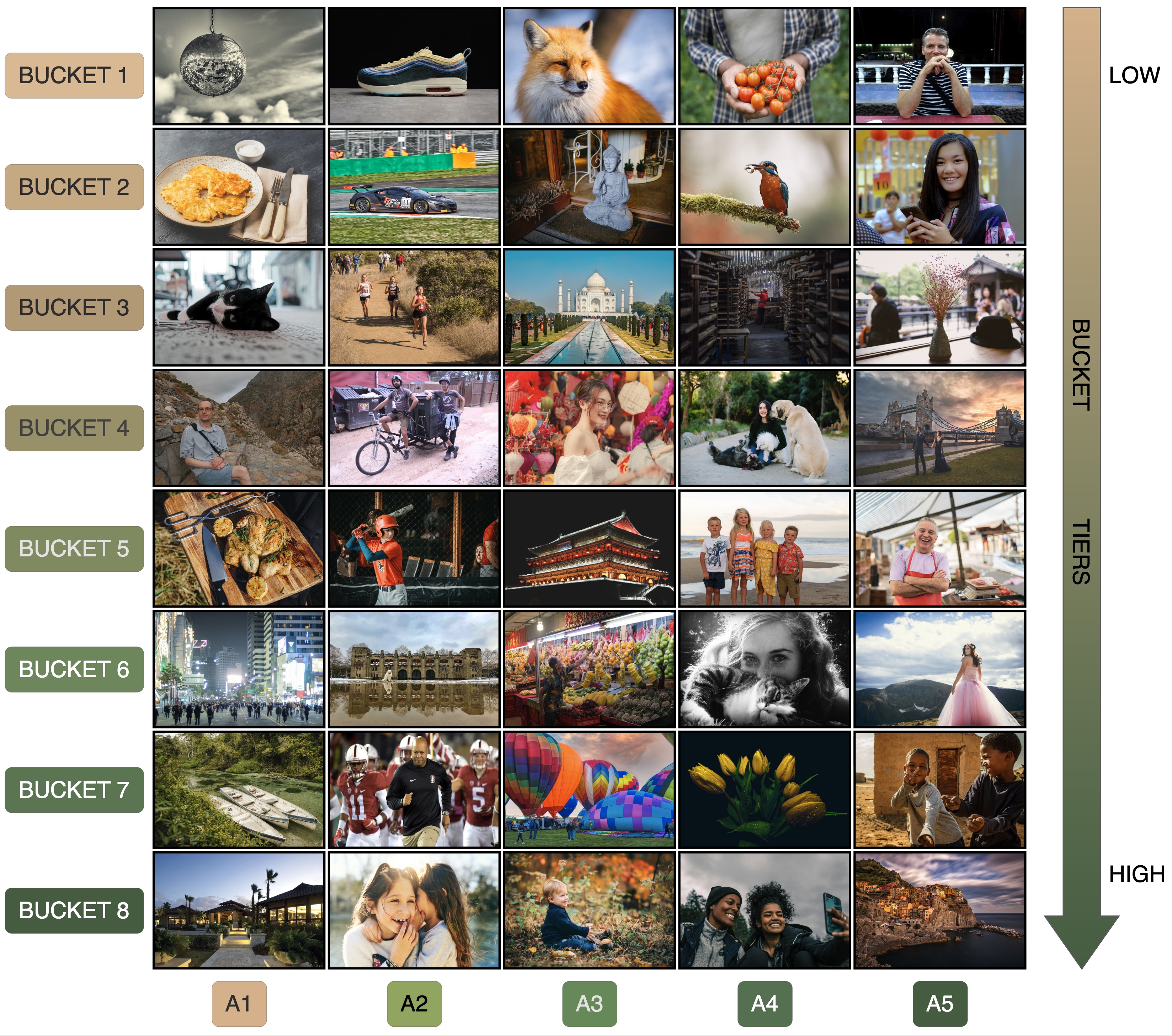}
  \caption{Joint view of static quality tiers and episodic buckets. Columns denote quality tiers A1-A5 and rows denote curriculum buckets B1-B8, illustrating that episodic bucketing acts as a second control axis layered on top of quality tiering.}
  \label{fig:bucket-tier-grid}
\end{figure}

Figure~\ref{fig:bucket-tier-grid} visualizes the two-axis organization of the training corpus. Columns correspond to static quality tiers A1-A5, while rows correspond to episodic buckets B1-B8, illustrating that curriculum buckets cut across multiple quality bands rather than replacing them.

\subsection{Training-time data consumption}
While the preceding subsections describe how the dataset is constructed and organized, this subsection documents the mechanics by which the training loop consumes it.

\textbf{Multi-resolution training.} Unlike the common practice of training at a fixed square resolution and introducing variable aspect ratios only at later stages, we train with multiple aspect ratios from the outset at every resolution stage. Training follows a progressive resolution curriculum (256\,\(\rightarrow\)\,512\,\(\rightarrow\)\,1024; see Section~\ref{sec:training-stages}), and at each stage the dataloader assigns each image to the valid crop size that maximizes retained area under that stage's token budget and aspect-ratio constraints.

\textbf{Aspect-ratio bucketing and batch construction.} Efficient VAE encoding and diffusion training require that all images in a batch share the same spatial dimensions. We therefore assign each example to an \emph{aspect-ratio bucket} keyed by a target crop size, and we construct each batch entirely within a single bucket.

The bucketing contract is: given a fixed token budget and a finite set of valid crop sizes (subject to aspect-ratio limits and divisibility constraints), we choose for each image the crop size that maximizes the fraction of the original image retained after center-cropping. The selected crop size becomes the example's bucket key, and the example index is appended to that bucket.

At dataset initialization, the loader enumerates valid crop sizes for each base resolution under (i) a token budget constraint, (ii) an aspect-ratio ceiling, and (iii) VAE compatibility constraints (e.g., divisibility by the patchification stride). A \texttt{ResolutionBucketSampler} shuffles indices within each bucket, packs them into fixed-size batches, and then shuffles batches across buckets before distributing them to data-parallel workers. This two-level shuffle ensures shape-uniform batches while avoiding long runs of a single aspect ratio during training. The sampler is stateful and supports exact resumption via checkpointed batch offsets.

\textbf{Task-conditioned supervision and filtering.} Each metadata row can be mapped into multiple task types (colorization, in-painting, zoom-in/out etc.). The training distribution over tasks is controlled by a weighted sampler, allowing the supervision mix to be adjusted per phase without rewriting metadata. The loader supports two complementary filtering mechanisms: simple scalar predicates (\texttt{min}, \texttt{max}, \texttt{equal}, \texttt{include}, \texttt{exclude}) are pushed down into the Parquet reader via columnar filters, while more complex predicates (regex matches, substring tests, and caption-array transforms such as excluding machine-translated captions) are applied in-memory after load. This split preserves I/O performance while supporting expressive filtering over caption provenance and length distributions.

\textbf{Failure handling and observability.} At training time, a non-trivial fraction of asset fetches fail due to transient storage errors or corrupted payloads. The dataloader handles failures by substituting a placeholder image that matches the expected bucket shape, ensuring tensor stackability without stalling the training step. Placeholder images are periodically refreshed with recently fetched real images so that placeholder content remains representative of the training distribution rather than a degenerate constant. Each sample carries an explicit success flag that propagates through the batch, enabling per-step and per-epoch failure-rate analytics and alerting when the fraction of placeholders exceeds a configured threshold. This metric is treated as a first-class training health signal where runs that exceed the configured threshold are investigated as data quality incidents rather than allowed to silently continue. In the event of a training crash or preemption, the sampler's checkpointed state enables exact resumption from the last completed batch without repeating or skipping data.

\section{Model Architecture}
\begin{figure}[t]
  \centering
  \includegraphics[width=\linewidth,height=0.60\textheight,keepaspectratio]{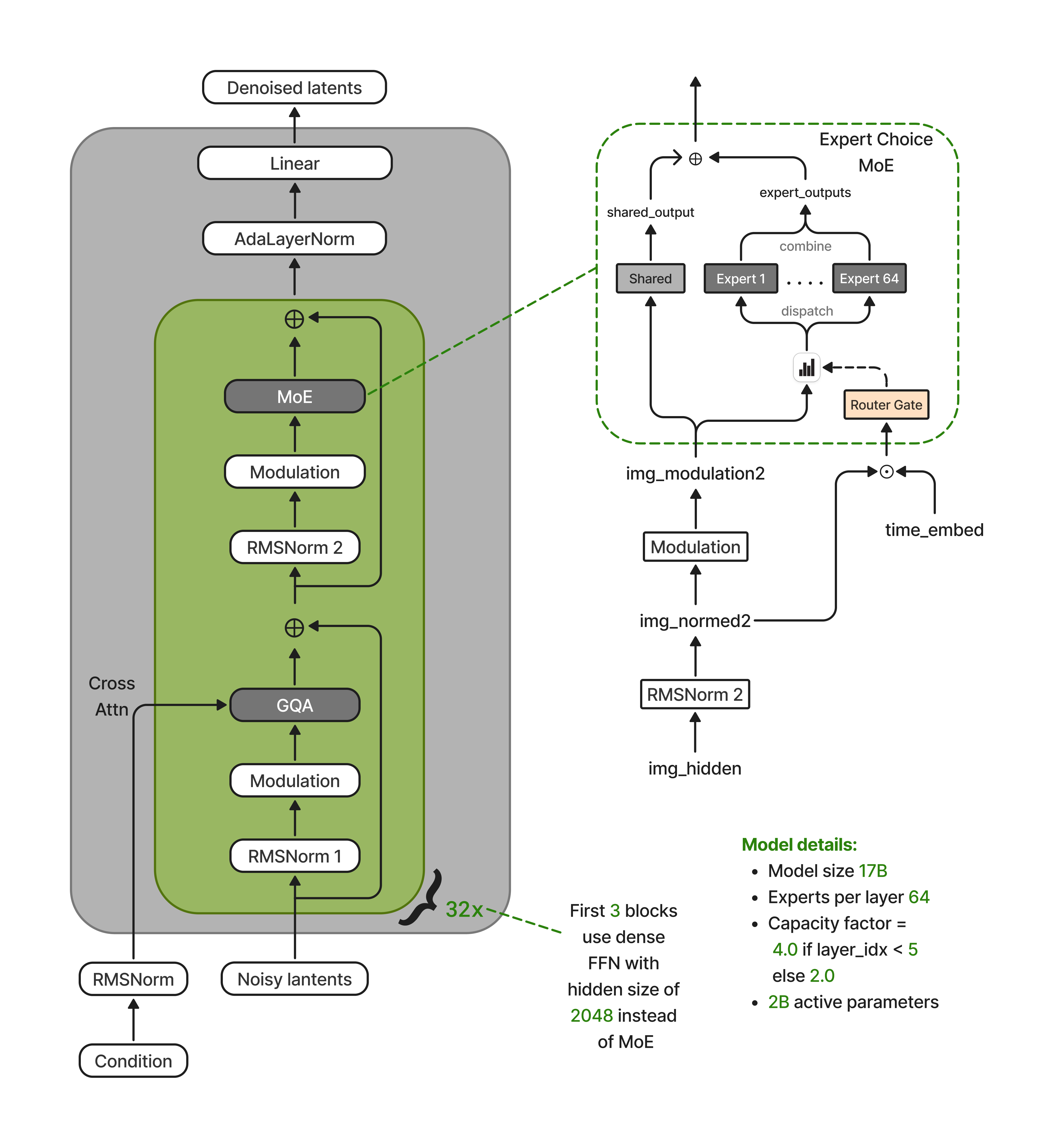}
  \caption{Overview of the Nucleus-Image architecture. \textbf{Left}: A single transformer block consisting of cross-attention (GQA), modulation, and MoE layers, repeated 32 times. The first 3 blocks use dense FFN with hidden size 2{,}048 instead of MoE. \textbf{Right}: The expert-choice MoE module with 64 routed experts and a shared expert, where the router selects tokens based on the unmodulated input and timestep.}
  \label{fig:architecture}
\end{figure}

Nucleus-Image is a diffusion transformer that employs a sparse mixture-of-experts (MoE) architecture to scale model capacity while maintaining computational efficiency. The model comprises 32 transformer layers with a hidden dimension of 2048, totaling approximately 17 billion parameters. Through expert-choice routing, only approximately 2 billion parameters are activated per forward pass. Figure~\ref{fig:architecture} provides an overview of the model architecture, and Table~\ref{tab:arch} summarizes the key architectural specifications.

\begin{table}[H]
  \caption{Nucleus-Image architectural specifications.}
  \centering
  \begin{tabular}{ll}
    \toprule
    Specification            & Value             \\
    \midrule
    Total parameters         & 17B               \\
    Active parameters        & \textasciitilde2B \\
    Layers                   & 32                \\
    Hidden dimension         & 2048              \\
    Attention heads (Q / KV) & 16 / 4            \\
    Head dimension           & 128               \\
    Experts per MoE layer    & 64 + shared       \\
    Expert hidden dimension  & 1344              \\
    Text encoder             & Qwen3-VL-8B-Instruct       \\
    Image tokenizer          & Qwen-Image VAE (16ch)       \\
    \bottomrule
  \end{tabular}
  \label{tab:arch}
\end{table}

\subsection{Image Tokenizer (VAE)}
The Variational Autoencoder (VAE) serves as the image tokenizer, compressing input images into compact latent representations for diffusion training and decoding them back to pixel space during inference. We use the pre-trained Qwen-Image VAE~\cite{qwen-image} without modification. This VAE extends the Wan-2.1 VAE~\cite{wan2025wan} with improved image decoding, particularly for text-rich content and fine-grained detail, and produces 16-channel latent representations that form the input to our diffusion transformer.

\subsection{Transformer Architecture}
Each transformer block follows a pre-normalization structure with adaptive modulation~\cite{peebles2023dit} conditioned on the diffusion timestep. The timestep embedding is produced by a sinusoidal encoding followed by a two-layer MLP, which then generates per-layer scale and gate parameters for modulating both the attention and feed-forward sub-layers.

For text conditioning, we employ direct cross-attention from the text encoder's hidden states rather than processing text tokens through a separate stream. Specifically, image queries attend to concatenated image and text key-value pairs, where text keys and values are projected from the Qwen3-VL~\cite{yang2025qwen3} encoder output. This design reduces computational overhead compared to bidirectional attention schemes while maintaining strong text-image alignment.

We adopt Grouped Query Attention (GQA)~\cite{ainslie2023gqa} with a 4:1 ratio, using 16 query heads and 4 key-value heads. This reduces the key-value cache by 4$\times$ during inference with negligible impact on generation quality. Following recent work on attention stability~\cite{lumina-image-2}, we apply RMSNorm to queries and keys prior to computing attention scores (QK-Norm).

For positional encoding, we follow the multi-dimensional rotary position embedding (mRoPE) formulation from Qwen-VL~\cite{qwen-image}, which encodes separate frequency components for temporal, height, and width axes. This enables the model to generalize across different image resolutions.

\subsection{Mixture of Experts}
We replace the dense feed-forward network with a sparse mixture-of-experts layer in 29 of the 32 transformer blocks, following recent work on scaling diffusion transformers with MoE~\cite{fei2024ditmoe}. Each MoE layer consists of 64 routed experts and one shared expert, where each expert is a SwiGLU~\cite{shazeer2020glu} network with hidden dimension 1344. For training stability, we find having at least three dense FFN layers before the MoE layers is essential.

\subsubsection{Expert-Choice Routing}
Unlike conventional token-choice routing where each token selects its top-$k$ experts, we employ expert-choice routing~\cite{zhou2022expertchoice,sun2024ecdit} where each expert selects its top-$k$ tokens. This guarantees uniform expert utilization and eliminates the need for auxiliary load-balancing losses during training.

A key design choice is that routing decisions are \emph{decoupled} from expert computation: the router receives unmodulated token representations concatenated with the timestep embedding, while expert MLPs receive the fully modulated representation. Section~\ref{sec:decoupled-routing} describes the motivation and design of this decoupling in detail.





\begin{lstlisting}[style=cleanpseudo,language=Python,float=t,caption={Expert-choice routing pseudocode.},label={lst:expert-choice-routing}]
# Concatenate unmodulated tokens with broadcasted timestep
t_expanded = broadcast(t, dim=1, size=S)           # [B, S, d]
router_input = concat(X_u, t_expanded, dim=-1)     # [B, S, 2d]

# Compute routing scores
logits = router_input @ W_r                         # [B, S, E]
scores = softmax(logits, dim=-1)                    # [B, S, E]

# Expert-choice: each expert selects top-k tokens
affinity = transpose(scores, dims=(1, 2))           # [B, E, S]
capacity = ceil(C * S / E)
top_indices = topk(affinity, k=capacity, dim=-1)    # [B, E, capacity]
gate_values = gather(affinity, top_indices)         # [B, E, capacity]

# Normalize gates per token (handle multi-expert selection)
token_totals = scatter_sum(gate_values, top_indices)
gate_normalized = gate_values / (token_totals + epsilon)
gate_scaled = gate_normalized * alpha

# Route tokens to experts
routed_tokens = gather(X, top_indices)              # [E x B x capacity, d]
expert_outputs = GroupedExperts(routed_tokens)      # [E x B x capacity, d]
expert_outputs = expert_outputs * gate_scaled

# Combine with shared expert
shared_output = SharedExpert(X)                     # [B, S, d]
output = shared_output + scatter_add(expert_outputs, top_indices)

return output
\end{lstlisting}

\subsubsection{Progressive Capacity Factor}
\label{sec:progressive-cf}
The capacity factor controls how many tokens each expert may select and therefore governs the density of expert participation. We apply capacity factor schedules along two orthogonal axes: \emph{across layers} within a single forward pass, and \emph{across training stages} as the resolution increases (Section~\ref{sec:training-stages}).

\paragraph{Per-layer schedule.}
We observe that early transformer layers benefit from broader token-expert interaction, while later layers can operate effectively with sparser routing. Table~\ref{tab:capacity-schedule} shows the per-layer schedule used during the final 1024-resolution stage.

\begin{table}[H]
  \caption{Per-layer capacity factor schedule (1024-resolution stage).}
  \centering
  \begin{tabular}{lll}
    \toprule
    Layer Index & Capacity Factor & Effective Experts per Token \\
    \midrule
    3--4        & 4.0             & \textasciitilde4            \\
    5--31       & 2.0             & \textasciitilde2            \\
    \bottomrule
  \end{tabular}
  \label{tab:capacity-schedule}
\end{table}

This schedule reduces computational cost in the deeper layers where representations are more specialized, while maintaining sufficient routing diversity in early layers for feature extraction.

\paragraph{Per-stage schedule.}
In addition to the per-layer variation, the capacity factor is progressively reduced across training stages as the model transitions from lower to higher resolutions. Table~\ref{tab:capacity-stage} summarizes this progression; Section~\ref{sec:training-stages} describes the full multi-stage curriculum.

\begin{table}[H]
  \caption{Capacity factor across training resolution stages.}
  \centering
  \begin{tabular}{lll}
    \toprule
    Resolution Stage & Capacity Factor       & Rationale                                                     \\
    \midrule
    256              & 8.0 (uniform)         & Dense gradients: only 256 image tokens across 64 experts      \\
    512              & 4.0 (uniform)         & Moderate sparsity as representations mature                   \\
    1024             & 4.0 / 2.0 (per-layer) & Full progressive schedule (Table~\ref{tab:capacity-schedule}) \\
    \bottomrule
  \end{tabular}
  \label{tab:capacity-stage}
\end{table}

The primary motivation for the high capacity factor at low resolutions is gradient density. Since text tokens participate only via cross-attention and do not pass through the MoE layers, the token sequence seen by each MoE layer consists solely of image tokens. At 256 resolution a square crop produces only $256$ tokens; with 64 experts and a capacity factor of 2, each expert would select just 8 tokens per sample---far too few for stable gradient estimates. A capacity factor of 8 raises this to 32 tokens per expert. As resolution increases, the longer token sequences provide sufficient per-expert counts at lower capacity factors. Section~\ref{sec:training-stages} discusses the per-stage schedule in detail.

\subsubsection{Shared Expert}
Each MoE layer includes a shared expert that processes all tokens unconditionally. The shared expert output is added to the sparse expert outputs before the residual connection. This design serves two purposes: (1) it provides a baseline computation path that ensures no token receives zero expert contribution, and (2) it allows the routed experts to specialize on complementary computations rather than redundantly learning common transformations.

\subsubsection{Decoupled Routing and Computation}
\label{sec:decoupled-routing}

A distinctive feature of our MoE design is the \emph{decoupling} of the representations used for routing decisions and expert computation. Each DiT block derives two representations from the post-attention residual $\mathbf{x}$:

\[
  \mathbf{x}_{\text{norm}} = \frac{\text{RMSNorm}(\mathbf{x})}{\sqrt{\ell}}, \qquad \mathbf{x}_{\text{mod}} = \mathbf{x}_{\text{norm}} \odot (1 + \mathbf{s}_2)
\]

where $\ell$ is the layer index (under progressive layer-norm scaling) and $\mathbf{s}_2$ is the timestep-derived modulation scale. The router receives the unmodulated representation $\mathbf{x}_{\text{norm}}$ concatenated with the timestep embedding, while the expert MLPs receive the fully modulated representation $\mathbf{x}_{\text{mod}}$.

\paragraph{Motivation.}
This decoupling addresses a failure mode specific to diffusion transformers that has no analogue in MoE language models. In LLMs, the router and experts see the same hidden state---there is no per-step extrinsic signal that rescales the router input. In diffusion transformers, however, the adaptive modulation scale $(1 + \mathbf{s}_2)$ varies substantially across timesteps: at convergence, the norm of $\mathbf{x}_{\text{mod}}$ at $t = 0.01$ can differ from $t = 0.99$ by an order of magnitude. When the router sees modulated inputs, its logits are dominated by the conditioning signal rather than token content, causing all tokens at a given timestep to be routed to the same subset of experts regardless of spatial or semantic identity. This effectively collapses expert-choice routing into \emph{timestep-choice} routing, destroying the spatial and semantic specialisation that makes MoE effective.

Routing purely on $\mathbf{x}_{\text{norm}}$ resolves stability but discards all timestep information from routing decisions. This is suboptimal because the optimal expert assignment \emph{should} depend on the denoising stage---high-noise steps benefit from experts specialising in coarse structure, while low-noise steps benefit from fine-detail experts.

\paragraph{Design.}
We resolve this tension by providing the timestep as a \emph{separate, normalised signal} concatenated with the unmodulated representation: $\text{router\_input} = [\mathbf{x}_{\text{norm}} \;\|\; \mathbf{t}]$. This preserves routing stability---the token component is invariant to modulation scale---while allowing the router to learn timestep-dependent expert preferences through the additive timestep channel. The expert MLPs receive $\mathbf{x}_{\text{mod}}$, retaining full conditioning expressivity for the actual computation. The result is a clean separation: \textbf{route by content and timestep, compute with conditioning}.

This design also provides a clean gradient path: router gradients flow through $\mathbf{x}_{\text{norm}}$ and do not carry modulation-scale gradients, while expert gradients flow through $\mathbf{x}_{\text{mod}}$ with the full conditioning gradient. This prevents the router from receiving conflicting gradient signals from the expert loss and the modulation scale.

\subsubsection{Expert Parallelism}
The MoE layers are parallelized across devices using expert parallelism, where each device holds a subset of experts. We utilize PyTorch's grouped matrix multiplication (\texttt{torch.\_grouped\_mm}) for efficient batched computation across experts with varying token counts. Further details on our parallelism strategy are provided in Section~\ref{sec:parallelism}.

\subsection{Fused Triton Kernels}
We implement custom Triton kernels for several operations that are memory-bandwidth bound in the standard PyTorch implementation.

\textbf{Fused Gated Residual.} The gated residual connection $\mathbf{x} \leftarrow \mathbf{x} + \tanh(\mathbf{g}) \odot \mathbf{r}$ is fused into a single kernel that avoids materializing the intermediate tanh output, reducing memory traffic by approximately 33\%.

\textbf{Fused LayerNorm-Scale.} The adaptive modulation pattern---layer normalization followed by element-wise scaling---is fused to eliminate an intermediate tensor write. This kernel computes $\text{LayerNorm}(\mathbf{x}) \odot (1 + \mathbf{s})$ in a single pass.

\textbf{Fused Gate-Residual-LayerNorm-Scale.} The complete sequence of gated residual addition, layer normalization, and scale modulation is fused into a single kernel, which is applied between the attention and MoE sub-layers in each block.

We additionally leverage the Liger Kernel library~\cite{hsu2024liger} for fused SwiGLU activations and RMSNorm operations, and Flash Attention 3~\cite{flash3} for variable-length attention computation.

\subsection{Architectural Choices for Training Stability}
Several architectural choices are critical for stable training of deep MoE diffusion transformers.

\textbf{Dense Initial Layers.} The first three transformer layers (indices 0, 1, 2) use dense feed-forward networks rather than MoE. We found this essential for training stability: without dense initial layers, the layer-wise activation norms grow rapidly in early training, leading to divergence. We hypothesize that routing decisions require meaningful token representations to be effective, which are not yet available in the earliest layers.

\textbf{QK-Normalization.} We apply RMSNorm~\cite{zhang2019root} to query and key projections before computing attention scores. The normalization weights are initialized to produce identity behavior and are frozen during training. This prevents attention logit explosion without introducing additional learnable parameters.

\textbf{Tanh Gating.} Residual connections use tanh-bounded gates: $\mathbf{x} \leftarrow \mathbf{x} + \tanh(\mathbf{g}) \odot \mathbf{r}$. The tanh nonlinearity constrains gate magnitudes to $[-1, 1]$, preventing unbounded residual accumulation. Gate parameters are initialized to zero, yielding identity residual connections at initialization.

\textbf{Decoupled Routing.} As described in Section~\ref{sec:decoupled-routing}, the router operates on the \emph{unmodulated} token representation concatenated with the timestep embedding, while expert computation uses the fully modulated representation. This decoupling prevents the adaptive modulation scale---which varies by an order of magnitude across timesteps---from dominating routing decisions and collapsing expert specialisation into timestep-dependent selection. The design is particularly important given our logit-normal timestep sampling distribution, which concentrates samples near intermediate noise levels and would otherwise leave extreme timesteps with degraded routing diversity.

\section{MoE Computational Kernels}

Efficient execution of mixture-of-experts layers requires careful attention to the computational kernels that underlie expert computation. Unlike dense layers where all tokens pass through identical weights, MoE layers route different subsets of tokens to different experts, creating irregular workloads that can be challenging to execute efficiently on modern accelerators. This section describes the kernel-level optimizations that enable high-throughput MoE computation in Nucleus-Image.

\subsection{Grouped Matrix Multiplication}

The core computation in each expert is a SwiGLU feed-forward network, which requires three matrix multiplications per expert. A naive implementation would either (1) loop over experts sequentially, underutilizing GPU parallelism, or (2) pad all experts to the maximum token count and use batched matrix multiplication, wasting computation on padding tokens.

We instead leverage grouped matrix multiplication, which performs multiple matrix multiplications of potentially different sizes in a single kernel launch. Given a sequence of token counts $[n_1, n_2, \ldots, n_E]$ for $E$ experts, grouped matrix multiplication computes:

\[
  \mathbf{Y}_e = \mathbf{X}_e \mathbf{W}_e^\top \quad \text{for } e = 1, \ldots, E
\]

where $\mathbf{X}_e \in \mathbb{R}^{n_e \times d}$ and $\mathbf{W}_e \in \mathbb{R}^{h \times d}$. The input tokens are concatenated into a single 2D tensor, and an offset array specifies the boundary between each expert's tokens. This approach achieves near-optimal hardware utilization without padding overhead, as the kernel schedules work across all experts simultaneously based on actual token counts.

\subsection{Fused SwiGLU Activation}

Each expert computes the SwiGLU activation:

\[
  \mathbf{h} = \text{SiLU}(\mathbf{X}\mathbf{W}_1^\top) \odot (\mathbf{X}\mathbf{W}_3^\top)
\]

where $\odot$ denotes element-wise multiplication. A naive implementation would materialize two intermediate tensors of shape $(n, h)$ before computing the element-wise product. We employ Liger kernel that computes the SiLU activation and element-wise multiplication in a single pass, reading inputs once and writing the final result directly. This reduces memory traffic by eliminating one intermediate tensor write and read cycle, which is significant given that MoE layers are often memory-bandwidth bound due to the large number of expert parameters.

\subsection{Token Permutation Kernel}

Expert-parallel execution (described in Section~\ref{sec:parallelism}) distributes experts across devices, requiring all-to-all communication to route tokens to their assigned experts. After the all-to-all exchange, tokens arrive ordered by their source device rather than by their target expert. However, grouped matrix multiplication requires tokens to be contiguous by expert.

A straightforward solution would compute the permutation indices on the CPU and transfer them to the GPU, but this incurs a device-to-host synchronization that serializes execution. We adapted the Triton-based token permutation kernel from TorchTitan~\cite{torchtitan2024}, which generates permutation indices entirely on the GPU without any host synchronization.

The kernel operates as follows. Given the token counts per expert from each source device (a tensor of shape $[R \times E]$ where $R$ is the number of devices and $E$ is the number of local experts), it computes:

\begin{enumerate}
  \item \textbf{Prefix sums} over the token counts to determine the starting position of each (device, expert) block in the input tensor.
  \item \textbf{Per-expert write offsets} by summing token counts across source devices for each expert.
  \item \textbf{Index generation} where each thread block handles one expert, iterating over source devices and writing sequential indices to the appropriate output positions.
\end{enumerate}

The resulting permutation maps the all-to-all output (ordered by source device) to the grouped-MM input (ordered by target expert). After expert computation, an inverse permutation restores the original ordering for the all-to-all combine operation.





\subsection{Integration with Expert Parallelism}

The kernels described in this section form the computational core of the MoE layer, but their efficiency depends critically on the data layout established by the parallelism strategy. Section~\ref{sec:parallelism} describes how tokens are distributed across devices via all-to-all communication, how expert weights are sharded, and how the forward and backward passes are orchestrated to minimize communication overhead. The permutation kernel serves as the bridge between the communication pattern (device-ordered tokens) and the computation pattern (expert-ordered tokens), enabling efficient grouped matrix multiplication in a distributed setting.

\section{Training}
\label{sec:training}

\subsection{Loss Functions}
We train Nucleus-Image using the rectified flow framework~\cite{liu2022flow,esser2024scaling}, which formulates image generation as learning a velocity field that transports samples along straight paths from noise to data. The primary training objective is a mean squared error loss on the predicted velocity. To promote balanced expert utilization in the MoE layers, we apply two auxiliary regularizers---z-loss and orthogonal loss---that act on the router weights and logits. In the final high-resolution fine-tuning stage on 1K images, we additionally introduce a wavelet-domain loss that emphasizes high-frequency detail reconstruction. Table~\ref{tab:loss} summarizes the loss components and their coefficients.

\subsubsection{Rectified Flow Loss}
The primary training objective follows the rectified flow formulation~\cite{liu2022flow,esser2024scaling}, which learns a velocity field that transports samples from a noise distribution to the data distribution along straight paths.

\paragraph{Formulation.}
Given a clean latent $\mathbf{x}_0$ encoded by the VAE and Gaussian noise $\boldsymbol{\epsilon} \sim \mathcal{N}(0, \mathbf{I})$, we construct a noisy latent at timestep $t \in [0, 1]$ via linear interpolation:

\[
  \mathbf{z}_t = (1 - t)\,\mathbf{x}_0 + t\,\boldsymbol{\epsilon}
\]

The model predicts the velocity $\mathbf{v}_\theta(\mathbf{z}_t, t)$, and the training objective minimizes the mean squared error against the ground-truth velocity:

\[
  \mathcal{L}_{\text{flow}} = \mathbb{E}_{t, \mathbf{x}_0, \boldsymbol{\epsilon}} \left[ \left\| \mathbf{v}_\theta(\mathbf{z}_t, t) - (\mathbf{x}_0 - \boldsymbol{\epsilon}) \right\|_2^2 \right]
\]

Under our convention, $t = 0$ corresponds to clean data and $t = 1$ corresponds to pure noise. The loss is computed in float32 for numerical stability, averaged first over the spatial and channel dimensions per sample, then across the batch.

\paragraph{Timestep Sampling.}
The distribution over timesteps $t$ significantly impacts training dynamics. We adopt a resolution-aware logit-normal sampling strategy inspired by Stable Diffusion 3~\cite{esser2024scaling}, with a 10\% uniform mixture for coverage.

The sampling procedure is as follows:
\begin{enumerate}
  \item Draw $z \sim \mathcal{N}(0, 1)$ and compute $t = \sigma(z)$ via the sigmoid function.
  \item Apply a resolution-dependent time shift:
        \[
          t \leftarrow \frac{e^\mu}{e^\mu + (1/t - 1)^\sigma}
        \]
        where $\sigma = 1.0$ and $\mu$ is a linear function of the image token count $N$:
        \[
          \mu(N) = 0.5 + \frac{1.15 - 0.5}{4096 - 256} \cdot (N - 256)
        \]
        This maps $\mu = 0.5$ at 256 tokens (low resolution) to $\mu = 1.15$ at 4096 tokens (high resolution), concentrating more probability mass at higher noise levels for larger images where the denoising task is harder.
  \item With probability 0.1, replace the logit-normal sample with a uniform draw $t \sim \mathcal{U}(0, 1)$. This mixture ensures adequate gradient signal at the extremes of the timestep range where the logit-normal density is low.
\end{enumerate}

\subsubsection{MoE Load Balancing Losses}
Expert-choice routing (Section 3.2.1) inherently guarantees uniform token allocation across experts, eliminating the need for standard load-balancing auxiliary losses used in token-choice MoE architectures. However, we find that two complementary regularizers on the router itself are important for training stability and expert specialization: a z-loss that prevents router logit explosion, and an orthogonal loss that encourages diverse routing directions. The z-loss is injected into the router computation graph via auxiliary gradient injection, while the orthogonal loss is applied via a direct post-backward weight update that bypasses optimizer state; neither mechanism alters the forward-pass routing decisions.

\paragraph{Z-Loss.}
The z-loss~\cite{zoph2022stmoe} penalizes the magnitude of router logits to prevent them from growing unboundedly during training. Large router logits produce increasingly peaked softmax distributions, which reduces the effective expert capacity and can destabilize optimization. Given router logits $\mathbf{l} \in \mathbb{R}^{S \times E}$ for $S$ tokens and $E$ experts, the z-loss is defined as:

\[
  \mathcal{L}_z = \frac{1}{S} \sum_{i=1}^{S} \left( \log \sum_{j=1}^{E} e^{l_{ij}} \right)^2
\]

The z-loss is applied before the softmax normalization of router scores. It is injected into the computation graph via a custom \texttt{AddAuxiliaryLoss} autograd function that passes gradients through during backpropagation without modifying the forward activations. The coefficient is set to $\lambda_z = 5 \times 10^{-7}$, which is small enough to avoid interfering with the primary training objective while providing sufficient regularization to prevent logit explosion across the 29 MoE layers.

\paragraph{Orthogonal Loss.}
The orthogonal loss regularizes the router weight matrix to encourage expert diversity. As analyzed by Guo et al.~\cite{guo2025advancing}, standard load-balancing losses can induce a self-reinforcing degradation loop in MoE models: auxiliary losses homogenize token-to-expert assignments, causing experts to receive overlapping training signals and converge to similar representations. As expert outputs become less distinguishable, the router receives weaker discriminative signals, leading to increasingly uniform routing decisions---which in turn further reduces expert specialization. Orthogonal regularization breaks this cycle by explicitly encouraging the router to maintain maximally differentiated routing directions for each expert.

Our approach follows the ERNIE 4.5~\cite{ernie2025report} design and applies orthogonality constraints directly to the router weight matrix rather than to expert outputs~\cite{guo2025advancing}. Given the router weight matrix $\mathbf{W}_r \in \mathbb{R}^{2d \times E}$ (where $2d$ is the routing input dimension and $E$ is the number of experts), we first column-normalize the expert vectors:

\[
  \hat{\mathbf{w}}_{r,j} = \frac{\mathbf{w}_{r,j}}{\|\mathbf{w}_{r,j}\|_2 + \epsilon}
\]

The orthogonal loss is then the normalized Frobenius distance between the Gram matrix and the identity:

\[
  \mathcal{L}_{\text{ortho}} = \frac{1}{E^2} \left\| \hat{\mathbf{W}}_r^\top \hat{\mathbf{W}}_r - \mathbf{I} \right\|_F^2
\]

This penalty decomposes into two components: the sum of squared off-diagonal entries, which penalizes correlation between routing directions of different experts, and the sum of squared deviations of diagonal entries from unity, which maintains unit-norm routing vectors. By driving the off-diagonal elements of $\hat{\mathbf{W}}_r^\top \hat{\mathbf{W}}_r$ toward zero, the loss ensures that each expert's routing vector spans a distinct subspace, providing the router with maximally informative projections for token assignment. This is complementary to the expert-choice routing mechanism (Section 3.2.1): while expert-choice routing guarantees balanced token counts per expert, the orthogonal loss ensures that the routing decisions themselves are semantically diverse.

A key distinction from the z-loss is how the orthogonal loss gradient is applied. Following the ERNIE 4.5 prescription~\cite{ernie2025report}, the gradient is used for a direct weight update that bypasses Adam's momentum and variance tracking:

\[
  \mathbf{W}_r \leftarrow \mathbf{W}_r - \lambda_{\text{ortho}} \cdot \nabla_{\mathbf{W}_r} \mathcal{L}_{\text{ortho}}
\]

with $\lambda_{\text{ortho}} = 1 \times 10^{-7}$. This update is executed after the backward pass but before the optimizer step, applied independently to each MoE layer's router weights. The gradient computation uses a separate \texttt{torch.enable\_grad()} context with detached weights to avoid interfering with the main computation graph. This direct-update approach ensures a consistent regularization magnitude independent of the optimizer's adaptive learning rate, which is important because adaptive optimizers can amplify or attenuate auxiliary gradients unpredictably depending on their accumulated statistics.

\subsubsection{Wavelet Loss}
At the end of 1K training, we incorporate the wavelet-domain loss introduced by \cite{wang2025diffusion4k}, which provides frequency-aware supervision. The standard MSE loss in latent space weights all spatial frequencies equally, which can yield blurry outputs as the model prioritizes low-frequency structure over fine-grained texture---a problem exacerbated at higher resolutions where perceptually important details span a wider frequency range.

\paragraph{Wavelet Transform.}
Following \cite{wang2025diffusion4k}, a single-level Haar discrete wavelet transform (DWT) decomposes each channel of the latent tensor into four sub-bands:

\begin{itemize}
  \item \textbf{LL} (low-low): Approximation coefficients capturing overall structure.
  \item \textbf{LH} (low-high): Horizontal detail coefficients.
  \item \textbf{HL} (high-low): Vertical detail coefficients.
  \item \textbf{HH} (high-high): Diagonal detail coefficients.
\end{itemize}

The Haar wavelet is chosen for its computational simplicity and exact invertibility, which are desirable properties for latent-space operations where the VAE has already performed substantial spatial compression.

\paragraph{Formulation.}
The wavelet loss is applied to the reconstructed clean latent rather than the velocity prediction. Given the model's velocity output $\mathbf{v}_\theta$ and the noise $\boldsymbol{\epsilon}$, the predicted clean latent is recovered as $\hat{\mathbf{x}}_0 = \mathbf{v}_\theta + \boldsymbol{\epsilon}$, which is then compared against the ground-truth clean latent $\mathbf{x}_0$ in the wavelet domain:

\[
  \mathcal{L}_{\text{wavelet}} = \left\| \text{LL}(\hat{\mathbf{x}}_0) - \text{LL}(\mathbf{x}_0) \right\|_2^2 + \alpha_{\text{hf}} \left( \left\| \text{LH}(\hat{\mathbf{x}}_0) - \text{LH}(\mathbf{x}_0) \right\|_2^2 + \left\| \text{HL}(\hat{\mathbf{x}}_0) - \text{HL}(\mathbf{x}_0) \right\|_2^2 + \left\| \text{HH}(\hat{\mathbf{x}}_0) - \text{HH}(\mathbf{x}_0) \right\|_2^2 \right)
\]

where $\alpha_{\text{hf}} = 1.0$ is the high-frequency sub-band weight. The DWT is computed in float32 for numerical stability.

\paragraph{Coefficient.}
The wavelet loss weight is set to $\lambda_{\text{wavelet}} = 0.1$. This loss is disabled ($\lambda_{\text{wavelet}} = 0$) during the initial pretraining phases and is activated only during the high-resolution fine-tuning stage, where the model has already acquired strong coarse structure and benefits from explicit high-frequency supervision to produce sharper 2K outputs.

\subsubsection{Total Loss}
The total training objective is:

\[
  \mathcal{L}_{\text{total}} = \mathcal{L}_{\text{flow}} + \lambda_{\text{wavelet}} \cdot \mathcal{L}_{\text{wavelet}}
\]

where $\lambda_{\text{wavelet}} = 0$ during pretraining and $\lambda_{\text{wavelet}} = 0.1$ during high-resolution fine-tuning. The MoE load balancing losses operate outside this scalar objective: the z-loss contributes gradients via the auxiliary loss injection mechanism during backpropagation, while the orthogonal loss is applied as a direct weight update after the backward pass. This separation ensures that the auxiliary regularizers do not distort the loss landscape seen by the primary optimizer.

\begin{table}[H]
  \caption{Loss components and coefficients.}
  \centering
  \begin{tabular}{lll}
    \toprule
    Loss component       & Coefficient      & Stage                             \\
    \midrule
    Rectified flow (MSE) & 1.0              & All phases                        \\
    Z-loss               & $5\times10^{-7}$ & All phases                        \\
    Orthogonal loss      & $1\times10^{-7}$ & All phases (direct weight update) \\
    Wavelet loss         & 0.1              & High-resolution fine-tuning       \\
    \bottomrule
  \end{tabular}
  \label{tab:loss}
\end{table}

\subsection{Optimizer}
Nucleus-Image is trained using Muon~\cite{jordan2024muon}, combined with a Warmup-Stable-Merge (WSM) learning rate schedule~\cite{tian2025wsm}. Muon orthogonalizes the Nesterov momentum buffer via Newton-Schulz iteration before each weight update, producing updates with near-unit spectral norm and more uniform effective step sizes across matrix dimensions compared to Adam-family optimizers. We apply RMS-norm learning rate adjustment following~\cite{liu2025muonscalable}, which allows Muon and AdamW parameter groups to share a single base learning rate and weight decay without per-group tuning.

\subsubsection{Parameter Groups}
Not all parameters are well-suited to orthogonalized updates. Parameters involved in input patchification, timestep conditioning, and output projection interface between heterogeneous representation spaces and empirically train more stably under per-element adaptive scaling. We therefore partition model parameters into three optimizer groups.

\textbf{Muon group.} All weight matrices with rank $\geq 2$ not listed below. This includes all attention projection weights ($\mathbf{W}_Q, \mathbf{W}_K, \mathbf{W}_V, \mathbf{W}_O$) and the MoE expert FFN weight matrices. These parameters receive Muon updates with momentum $\mu = 0.95$, Nesterov lookahead, and RMS-norm-scaled learning rate. Crucially, we keep Q, K, V as separate projections rather than a fused QKV matrix, and similarly keep the SwiGLU gate and up projections unfused. This ensures each matrix retains a well-conditioned aspect ratio for Newton-Schulz orthogonalization, and that each matrix receives an individually calibrated RMS-norm learning rate adjustment.

\textbf{AdamW group.} The patch embedding (\texttt{patch\_embed}), timestep and image modulation projections (\texttt{time\_embed}, \texttt{img\_mod}), and the output head (\texttt{final\_norm}, \texttt{final\_proj}). Our initial experiments found modulation parameters unstable under Muon; we place them under AdamW instead. These parameters are optimized with AdamW~\cite{adamw} at $\beta_1 = 0.9$, $\beta_2 = 0.95$, and weight decay $\lambda = 0.01$.

\textbf{No-decay AdamW group.} All 1-dimensional parameters (biases, normalization scales) and router gate weights (\texttt{router.gate}). Router weights are excluded from weight decay to avoid penalizing the learned routing distribution. AdamW is applied with $\lambda = 0$ and the same $\beta$ values.

All three groups share a base learning rate of $\eta = 1 \times 10^{-4}$ and weight decay $\lambda = 0.01$ (where applicable).

\begin{table}[H]
  \caption{Optimizer groups and hyperparameters.}
  \centering
  \begin{tabular}{llllll}
    \toprule
    Group    & Parameters                           & Optimizer & $\eta$           & $\beta_1 / \beta_2$ or $\mu$ & $\lambda$ \\
    \midrule
    Muon     & Attention \& expert FFN weights      & Muon      & $1\times10^{-4}$ & $\mu = 0.95$                 & 0.01      \\
    AdamW    & Patch embed, modulation, output head & AdamW     & $1\times10^{-4}$ & 0.9 / 0.95                   & 0.01      \\
    No-decay & Biases, norms, router gates          & AdamW     & $1\times10^{-4}$ & 0.9 / 0.95                   & 0         \\
    \bottomrule
  \end{tabular}
  \label{tab:optim}
\end{table}

\subsection{Learning Rate Schedule}

Two observations motivate our choice of learning rate schedule.

\paragraph{Observation 1: EMA is ubiquitous but costly.}
Standard practice in diffusion model training---from the original Stable Diffusion~\cite{rombach2021highresolution} to SD~3.5~\cite{esser2024scaling}---is to maintain an exponential moving average (EMA) of model weights in parallel with the optimiser:
\[
  \theta_{\text{ema}}^{(t)} \;=\; \beta\,\theta_{\text{ema}}^{(t-1)} + (1-\beta)\,\theta^{(t)},
\]
with $\beta$ typically $0.9999$. The EMA weights are discarded during training and only used at inference. This requires storing a full extra copy of the model, doubling the weight memory footprint and adding $\mathcal{O}(P)$ work per step---a permanent tax on a resource that is consumed only once the run finishes.

\paragraph{Observation 2: Fixed LR decay limits training flexibility.}
The widely-used Warmup-Stable-Decay (WSD) schedule requires committing up front to when and for how long to anneal the learning rate. Mid-run decisions---extending training, switching data mixtures, or continuing from a released checkpoint---force an expensive re-decay from scratch. This rigidity is particularly painful for image generation, where data curricula (resolution, quality filtering, aesthetic scoring) are often iterated on during a run.

\paragraph{Connecting EMA to checkpoint merging.}
Both observations resolve to the same solution once one notices that \textbf{online EMA is mathematically equivalent to a geometric weighted average of past checkpoints}. Unrolling the EMA recurrence from step $n$ over $k$ subsequent steps, with the initialisation $\theta_{\text{ema}}^{(n)} = \theta^{(n)}$, yields
\[
  \theta_{\text{ema}}^{(n+k)} \;=\; \beta^{k}\,\theta^{(n)} \;+\; \sum_{j=1}^{k}(1-\beta)\,\beta^{k-j}\,\theta^{(n+j)}.
\]
Defining checkpoint weights $c_0 = \beta^k$ and $c_j = (1-\beta)\beta^{k-j}$ for $j \geq 1$, this is exactly a weighted checkpoint average $\hat{\theta} = \sum_{j=0}^{k} c_j\,\theta^{(n+j)}$ with $\sum_j c_j = 1$. The weights are geometric: among the non-base checkpoints, newer ones receive strictly higher weight; the base checkpoint receives $\beta^{k}$.

Via WSM Theorem~3.1~\cite{tian2025wsm}, any such weighted checkpoint average is equivalent to training from $\theta^{(n)}$ with a synthetic gradient decay schedule whose effective per-step learning rates are
\[
  w_j \;=\; \sum_{m=j}^{k} c_m \;=\; 1 - \beta^{\,k-j+1}, \qquad j = 1,\ldots,k.
\]
For the geometric (EMA) weights this produces a ``late-drop'' profile: the effective LR stays near~1 for most of the merge window and falls to $1-\beta$ at the final step. With $\beta=0.9999$ the floor is $10^{-4}$, matching the deep decay used in practice. Setting $w_j = \beta^j$ exactly recovers EMA, confirming that EMA during training and exponential-weight checkpoint merging after training are the same operation---one online, one offline. Because the merge weights $\{c_j\}$ are free parameters, choosing a better decay profile than exponential improves over EMA; in practice an inverse-square-root profile outperforms both mean averaging and EMA-based merging~\cite{tian2025wsm}.

The effective lookback horizon of EMA is $1/(1-\beta)$ steps. Saving checkpoints every $\Delta$ steps and merging the last $k \gg 1/\bigl(\Delta(1-\beta)\bigr)$ of them closely approximates the EMA---the residual weight on the base checkpoint decays as $\beta^{k\Delta}$---with zero memory overhead during training.

\paragraph{The WSM schedule.}
Nucleus Image adopts the Warmup-Stable-Merge (WSM) schedule~\cite{tian2025wsm}:
\[
  \eta(t) = \begin{cases}
    \eta_{\text{peak}} \cdot t\,/\,T_{\text{warmup}} & t < T_{\text{warmup}},    \\[4pt]
    \eta_{\text{peak}}                               & t \geq T_{\text{warmup}}.
  \end{cases}
\]
No learning rate decay is applied during training. We warm up linearly for 1{,}000 steps to $\eta_{\text{peak}} = 1 \times 10^{-4}$ and hold the learning rate constant thereafter, saving checkpoints every 2{,}500 steps. The final model is produced by merging the last $N$ checkpoints using inverse-square-root weights derived from Theorem~3.1:
\[
  c_j \;\propto\; 1 - \sqrt{j\,/\,N}, \qquad j = 0,\ldots,N-1 \quad\text{(oldest to newest)}.
\]
Mean averaging (corresponding to linear decay) is used as a simpler baseline and performs comparably. This design eliminates both costs simultaneously:
\begin{itemize}[nosep]
  \item \textbf{No EMA shadow copy} is maintained during training---the merge provides the same noise-smoothing through temporal parameter averaging, with the additional freedom to choose non-geometric weight distributions.
  \item \textbf{No upfront decay commitment}---the merge scheme and window size $N$ can be chosen, or re-run with different weights, entirely after training ends without touching the optimisation trajectory.
  \item \textbf{Continual training is trivial}---extend the run at constant LR and re-merge the latest $N$ checkpoints to produce a new release model at any point.
\end{itemize}

On GenEval (1024$\times$1024, 50 steps, CFG~\cite{ho2022cfg}\,=\,8.0), the merged checkpoint using the inverse-square-root scheme with $N=16$ gains \textbf{+3.2 points} over the equivalent unmerged checkpoint. This improvement is consistent across all task categories and persists across CFG and inference step sweeps.

\subsection{Initialization}
All weight matrices are initialized with truncated normal distributions ($\sigma = 0.02$). Modulation parameters---the scale and shift projections in the adaptive layer norm conditioning---are initialized to zero, so that each transformer block computes an identity transformation at the start of training. Router gate weights are initialized with a tighter truncated normal ($\sigma = 0.006$) to keep initial routing logits near-uniform, ensuring balanced expert utilization before the router has learned meaningful preferences.

\subsection{Parallelism}
\label{sec:parallelism}
Nucleus-Image's 17B-parameter MoE architecture is trained using FSDP2~\cite{zhao2023pytorch} combined with Expert Parallelism for the sparse MoE layers. The MoE layers distribute their 64 experts across all 64 GPUs, with each device hosting exactly one expert. This Expert Parallelism (EP) strategy, combined with Expert-Choice routing, ensures perfect load balance at the parameter level while requiring \texttt{all-to-all} communication to route tokens between devices. All other parameters are sharded within each node and replicated across nodes to minimize the \texttt{all-gather} communication across nodes during forward pass. To further reduce the communication overhead, we implement a selective resharding policy where the final transformer block and output projection layers retain the gathered parameters to make them available immediately for the backward pass. This design achieves high hardware utilization while maintaining the memory and communication efficiency necessary for large-scale training.

We use standard PyTorch distributed primitives for Expert Parallelism. In a forward pass, an \texttt{all-to-all} collective communicates the number of tokens each device will send to each expert. A second \texttt{all-to-all} collective moves the actual tokens to their destination expert. Since all the tokens on one GPU are accumulated for exactly one expert in our setting, we fallback to highly optimized \texttt{torch.matmul} instead of \texttt{torch.\_grouped\_mm} for expert forwards. After expert computation, a symmetric \texttt{all-to-all} collective returns processed tokens to their original devices.

Nucleus-Image training uses 8 nodes, each equipped with 8 NVIDIA H100 GPUs interconnected via NVLink. Inter-node communication uses GPUDirect-TCPXO over high-bandwidth RDMA networking. The GPUDirect-TCPXO stack enables direct GPU-to-GPU communication across nodes without CPU involvement, minimizing latency for the all-to-all collectives that dominate expert parallelism communication. We found all-to-all collectives scale well on our setup upto 16 nodes, beyond which we start to incur inefficiencies which can be solved by frameworks like DeepEP~\cite{deepep2025}.

\subsection{Multi-Stage Training Curriculum}
\label{sec:training-stages}
Nucleus-Image training follows a progressive resolution curriculum that increases the base resolution in three stages: 256, 512, and 1024 pixels. At every stage, the dataloader generates multiple aspect-ratio buckets from the stage's base resolution (Section~2.7), so the model sees diverse spatial layouts throughout training rather than training on square crops first and introducing aspect-ratio variation later.

\begin{table}[H]
  \caption{Multi-stage training curriculum. Each stage trains with multiple aspect ratios derived from its base resolution. The capacity factor is progressively reduced as expert specialization matures.}
  \centering
  \begin{tabular}{lllll}
    \toprule
    Stage & Base Resolution & Global Batch Size & Capacity Factor       & Notes                                           \\
    \midrule
    1     & 256             & 4{,}096           & 8.0 (uniform)         & Dense gradients; multiple aspect ratios         \\
    2     & 512             & 1{,}024           & 4.0 (uniform)         & Moderate sparsity; larger token sequences       \\
    3     & 1024            & 256               & 4.0 / 2.0 (per-layer) & Full progressive schedule; wavelet loss enabled \\
    \bottomrule
  \end{tabular}
  \label{tab:training-stages}
\end{table}

\textbf{Stage 1 (256 resolution).} Training begins at 256-pixel base resolution with a global batch size of 4{,}096 and a uniform capacity factor of 8 across all MoE layers. At this resolution, each image produces only $\sim\!256$ tokens (text tokens are not processed by the MoE layers, participating only via cross-attention). With 64 experts, a capacity factor of 8 ensures each expert selects $\lceil 8 \times 256 / 64 \rceil = 32$ tokens per sample, providing dense gradient coverage. A lower capacity factor would leave each expert with too few tokens for stable gradient estimates during this critical early phase. The large batch size compensates for the short per-sample token sequences, maintaining a high effective token throughput per step. Even at this low resolution, aspect-ratio bucketing produces a range of non-square crop sizes (e.g.\ $256\!\times\!128$, $192\!\times\!256$), exposing the model to varied spatial structure from the start.

\textbf{Stage 2 (512 resolution).} The base resolution is increased to 512 pixels, yielding $\sim\!1{,}024$ image tokens per sample, and the global batch size is reduced to 1{,}024. The capacity factor is lowered to 4; at this token count each expert still selects $\lceil 4 \times 1024 / 64 \rceil = 64$ tokens---sufficient for stable gradients---while beginning to enforce selective expert utilization.

\textbf{Stage 3 (1024 resolution).} The final stage trains at 1024-pixel base resolution with a global batch size of 256 and introduces the per-layer progressive capacity factor schedule (Section~\ref{sec:progressive-cf}): layers 3--4 retain a capacity factor of 4, while layers 5--31 operate at a capacity factor of 2. The smaller batch size reflects the $\sim\!16\times$ increase in per-sample token count relative to Stage~1, keeping memory and compute per step tractable. The wavelet loss (Section~4.1.3) is activated during this stage to provide explicit high-frequency supervision, improving fine-grained detail at high resolution.

\textbf{Rationale.} The core constraint is gradient density. Because only image tokens pass through the MoE layers, the per-expert token count is directly proportional to image resolution and capacity factor. At low resolutions the token budget is small, so a high capacity factor is necessary to ensure every expert receives enough tokens for stable gradient estimates. As resolution increases and token sequences grow, the capacity factor can be safely reduced---\emph{progressive sparsification}---without starving experts of gradient signal. This coupling also provides a natural compute curriculum: early stages process shorter sequences at lower cost per step, allowing the model to learn coarse structure efficiently before committing full compute to high-resolution detail.

\subsection{Source and Quality Aware Conditioning}
\label{subsec:system-prompt}

Each training example is prepended with a system prompt that communicates the data category and quality expectations to the diffusion backbone. The motivation is to provide clear training signals that prevent the model from conflating distinct data roles. For instance, synthetic text-rendering data is included to teach typographic capabilities, and its supervision should not pollute the model's treatment of incidental text in real-world photographs. Similarly, because the training mixture includes images across a wide quality range from the outset, the model must distinguish between lower-quality images that contribute broad world knowledge and high-aesthetic images that teach visually polished output without confusing the two training objectives.

Sytem Prompts are designed based on two axes. The first axis is \emph{source type}: real photographs, diffusion-generated images, digital illustrations and UI designs, and synthetic text-rendering data each receive a prompt tailored to the expected visual domain. The second axis is \emph{quality}: for real images, aesthetic scores and downstream quality signals determine whether the prompt emphasizes photorealistic, ultra-high-quality output or adopts a more general realism framing; synthetic images are differentiated using non-aesthetic signals such as caption quality and provenance. System prompts are formatted using the ChatML template native to the Qwen3-VL text encoder~\cite{yang2025qwen3}. Because the system prompt is derived deterministically from existing metadata, it adds no new annotation cost and functions as a lightweight conditioning signal that lets the model distinguish generation modes within a single architecture.

\subsection{Hyperparameters}
Table~\ref{tab:hparams} summarizes the key model and training hyperparameters captured in the current implementation. Items not yet finalized are marked as pending.

\begin{table}[H]
  \caption{Model and training hyperparameters.}
  \centering
  \begin{tabular}{ll}
    \toprule
    Parameter                     & Value                              \\
    \midrule
    Hidden dimension              & 2048                               \\
    Number of layers              & 32                                 \\
    Attention heads (Q / KV)      & 16 / 4                             \\
    Number of experts             & 64 + shared                        \\
    Experts per token (effective) & \textasciitilde2                   \\
    Expert hidden dimension       & 1344                               \\
    Text encoder                  & Qwen3-VL-8B-Instruct                        \\
    Image tokenizer               & Qwen-Image VAE (16ch)                        \\
    Learning rate (peak)          & $1\times10^{-4}$                   \\
    Learning rate schedule        & Warmup-Stable-Merge(WSM)           \\
    Weight decay                  & 0.01                               \\
    Precision                     & bf16                               \\
    Batch size (global)           & 4{,}096 / 1{,}024 / 256 (by stage) \\
    Total training steps          & 1.7M                            \\
    \bottomrule
  \end{tabular}
  \label{tab:hparams}
\end{table}

\section{Inference Optimizations}
Nucleus-Image is designed with inference efficiency as a first-class concern. Several architectural decisions made during training translate directly into latency and memory savings at inference time without requiring post-hoc modifications.

\subsection{Text Embedding Cache Across Denoising Steps}
\label{sec:inference-text-kv-cache}
Three architectural choices jointly ensure that no text-side computation needs to repeat across denoising steps.

\textbf{Text tokens are not timestep-conditioned.} Adaptive modulation via the timestep embedding is applied exclusively to image tokens. Text tokens receive no scale or gate parameters derived from $t_{\text{emb}}$, so their intermediate representations carry no timestep dependence.

\textbf{Text tokens do not generate query vectors.} Since text tokens contribute only key and value projections to each attention layer, the entire text processing pipeline --- the block-level projection $\mathbf{W}_{\text{enc}}$, the key and value projections $\mathbf{W}_K$ and $\mathbf{W}_V$, key normalization, and text RoPE application --- is a pure function of the text encoder output $\mathbf{C}$, which is fixed for a given prompt.

%
%
%

\subsection{Grouped Query Attention (GQA)}
We use Grouped Query Attention with 16 query heads and 4 key-value heads (4:1 ratio). At inference, the active KV cache for image tokens is:

\[
  \text{KV cache size} = 2 \times L \times S_i \times H_{\text{kv}} \times d_h \times \mathrm{sizeof(dtype)}
\]

where $H_{\text{kv}} = 4$ rather than 16. This yields a $4\times$ reduction in key-value cache memory for image tokens relative to standard multi-head attention, with a corresponding reduction in KV memory bandwidth.

\subsection{Text Tokens Excluded from MLP and MoE}
In dual-stream architectures such as Flux~\cite{flux2024} and SD3~\cite{esser2024scaling}, text tokens are full participants in transformer sub-layers. In Nucleus-Image, text tokens participate only as KV contributors in attention. They do not:

\begin{itemize}
  \item generate query vectors,
  \item pass through any MLP or MoE layer,
  \item receive adaptive modulation from the timestep embedding.
\end{itemize}

This design eliminates MoE routing overhead for text tokens entirely. Combined with text KV caching (Section~\ref{sec:inference-text-kv-cache}), text processing cost at inference reduces to a single text encoder forward pass plus one-time per-layer projections.

\subsection{Joint Attention over Image and Text}
Rather than separate attention operations for image self-attention and image-to-text cross-attention, we perform a single joint attention where image queries attend to concatenated image and text KV:

\begin{verbatim}
k_joint = cat([k_img, k_txt], dim=1)   # (B, S_i + S_t, H_kv, d_h)
v_joint = cat([v_img, v_txt], dim=1)
out = flash_attention(q=q_img, k=k_joint, v=v_joint)
\end{verbatim}

This requires only one Flash Attention kernel call per block rather than two. With text KV cached, the per-step concatenation is lightweight, and attention FLOPs scale as $O(S_i \cdot (S_i + S_t))$ rather than $O((S_i + S_t)^2)$ as in bidirectional joint attention.

\FloatBarrier

\section{Evaluation}
We evaluate Nucleus-Image on three standard text-to-image benchmarks: GenEval~\cite{ghosh2023geneval}, DPG-Bench~\cite{hu2024ella}, and OneIG-Bench~\cite{chang2025oneig}. These benchmarks provide complementary coverage: GenEval and DPG-Bench focus on compositional faithfulness, while OneIG-Bench measures broader capabilities including text rendering, reasoning, style, and diversity. All Nucleus-Image results are generated at $1024\times1024$ resolution, 50 inference steps, and CFG scale 8.0. Notably, all reported results are from the base model without any reinforcement learning, DPO, or preference tuning.

\subsection{GenEval}
GenEval~\cite{ghosh2023geneval} evaluates whether models correctly render specified objects, attributes, and spatial relations across six categories: single object, two objects, counting, colors, position, and attribute binding.

\begin{table}[H]
  \caption{GenEval results.}
  \centering
  \scriptsize
  \resizebox{\textwidth}{!}{%
    \begin{tabular}{lccccccc}
      \toprule
      Model                                      & Single        & Two  & Count & Colors & Position      & Attr. & Overall       \\
      \midrule
      Show-o~\cite{xie2024show}                  & 0.95          & 0.52 & 0.49  & 0.82   & 0.11          & 0.28  & 0.53          \\
      PixArt-$\alpha$~\cite{chen2024pixartalpha} & 0.98          & 0.50 & 0.44  & 0.80   & 0.08          & 0.07  & 0.48          \\
      Emu3-Gen~\cite{wang2024emu3}               & 0.98          & 0.71 & 0.34  & 0.81   & 0.17          & 0.21  & 0.54          \\
      SD3 Medium~\cite{esser2024scaling}         & 0.98          & 0.74 & 0.63  & 0.67   & 0.34          & 0.36  & 0.62          \\
      FLUX.1 Dev~\cite{flux2024}                 & 0.98          & 0.81 & 0.74  & 0.79   & 0.22          & 0.45  & 0.66          \\
      SD3.5 Large~\cite{esser2024scaling}        & 0.98          & 0.89 & 0.73  & 0.83   & 0.34          & 0.47  & 0.71          \\
      JanusFlow~\cite{ma2025janusflow}           & 0.97          & 0.59 & 0.45  & 0.83   & 0.53          & 0.42  & 0.63          \\
      Janus-Pro-1B~\cite{chen2025janus}          & ---           & 0.87 & 0.67  & ---    & ---           & 0.62  & 0.73          \\
      Janus-Pro-7B~\cite{chen2025janus}          & 0.99          & 0.89 & 0.59  & 0.90   & 0.79          & 0.66  & 0.80          \\
      HiDream-I1-Full~\cite{cai2025hidream}      & \textbf{1.00} & 0.98 & 0.79  & 0.91   & 0.60          & 0.72  & 0.83          \\
      \midrule
      Seedream 3.0~\cite{gao2025seedream}        & 0.99          & 0.96 & 0.91  & 0.93   & 0.47          & 0.80  & 0.84          \\
      Qwen-Image~\cite{qwen-image}               & 0.99          & 0.92 & 0.89  & 0.88   & 0.76          & 0.77  & 0.87          \\
      GPT Image 1 High~\cite{gptimage}           & 0.99          & 0.92 & 0.85  & 0.92   & 0.75          & 0.61  & 0.84          \\
      \midrule
      \textbf{Nucleus-Image}                     & 0.99          & 0.95 & 0.78  & 0.92   & \textbf{0.85} & 0.71  & \textbf{0.87} \\
      \bottomrule
    \end{tabular}
  }
  \label{tab:geneval-results}
\end{table}

\begin{figure}[t]
    \centering
    \includegraphics[width=\linewidth,height=0.40\textheight,keepaspectratio]{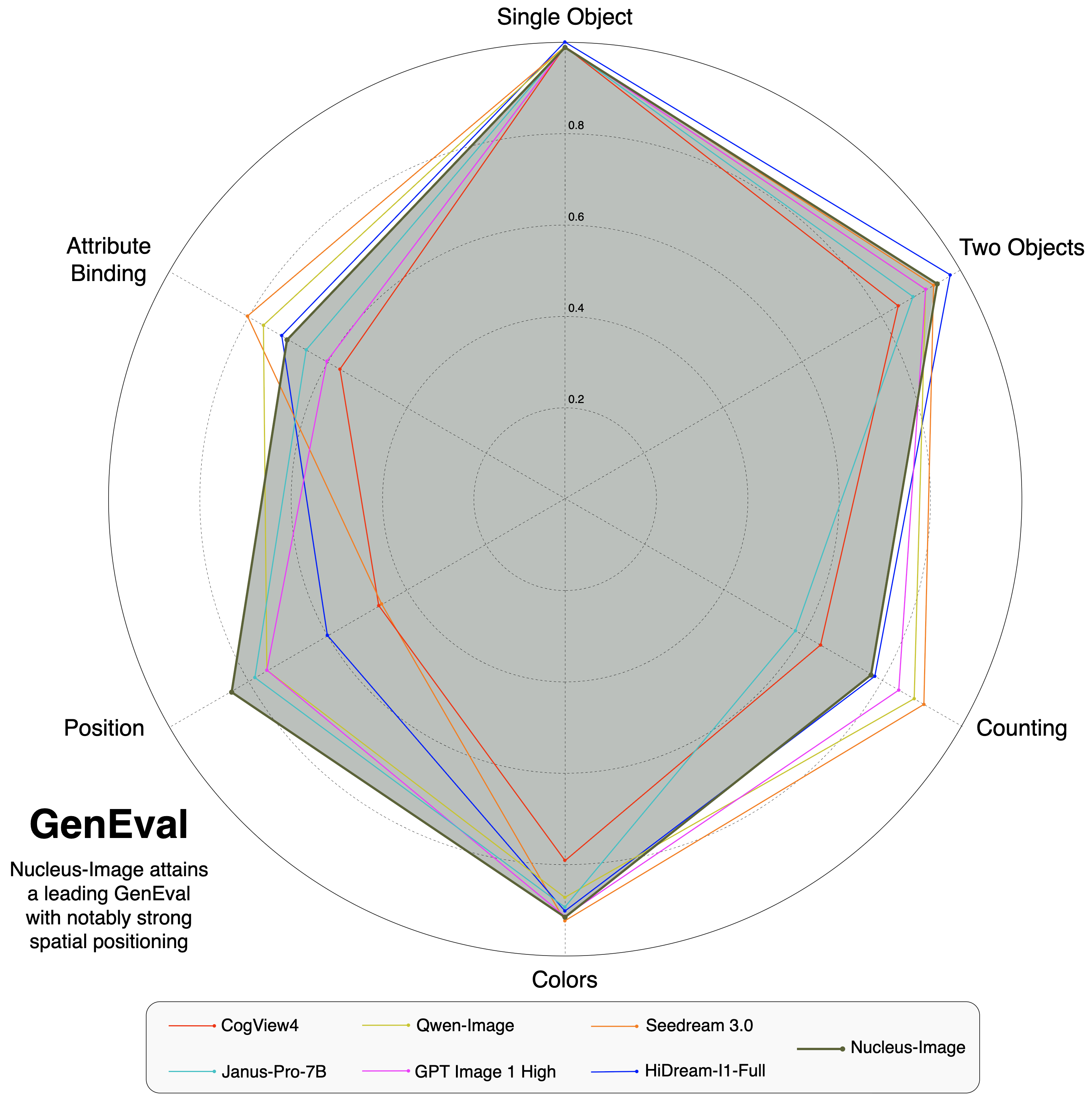}
    \caption{GenEval overall scores for top-performing models. Nucleus-Image matches Qwen-Image at 0.87 and leads all models on spatial position understanding.}
    \label{fig:geneval}
  \end{figure}

As shown in Table~\ref{tab:geneval-results}, Nucleus-Image achieves an overall score of 0.865 (reported as 0.87), matching Qwen-Image and surpassing all other reported models including GPT Image 1 High (0.84) and Seedream 3.0 (0.84). Notably, Nucleus-Image leads all models on spatial position (0.85), a category where even strong baselines like SD3.5 Large (0.34) and FLUX.1 Dev (0.22) struggle, demonstrating the model's strong understanding of spatial layout from text. Two-object (0.95) and colors (0.92) are also highly competitive, with counting (0.78) the primary area for improvement.

\subsection{DPG-Bench}
DPG-Bench~\cite{hu2024ella} evaluates dense prompt following on 1,065 multi-requirement prompts. Each prompt is decomposed into dependency-structured VQA checks spanning entity presence, attributes, relations, and counts.

\begin{figure}[H]
  \centering
  \includegraphics[width=\linewidth,height=0.30\textheight,keepaspectratio]{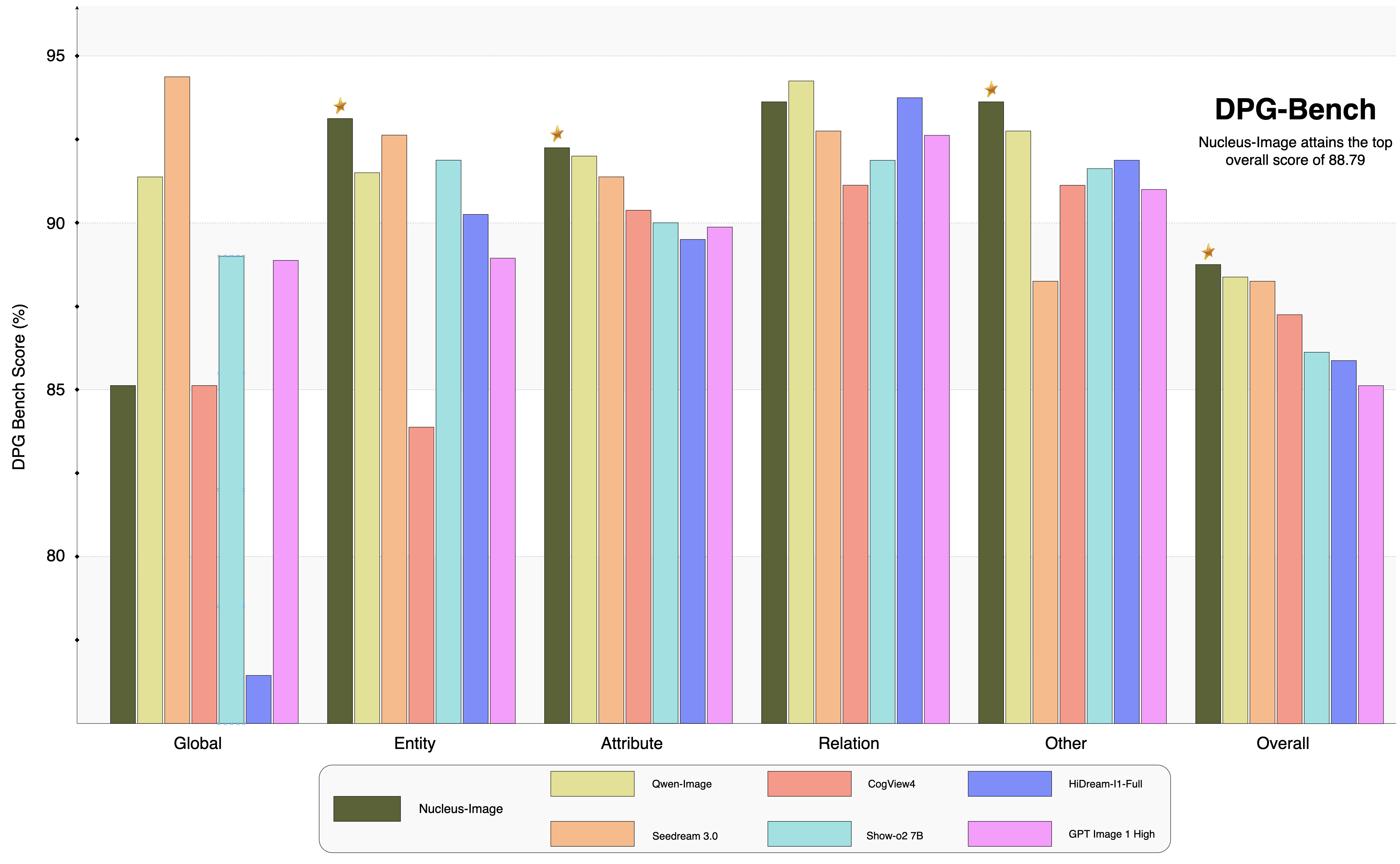}
  \caption{DPG-Bench overall scores for top-performing models. Nucleus-Image achieves the highest overall score of 88.79, leading in entity, attribute, and other subcategories.}
  \label{fig:dpg-bench}
\end{figure}

\begin{table}[H]
  \caption{DPG-Bench results.}
  \centering
  \scriptsize
  \resizebox{\textwidth}{!}{%
    \begin{tabular}{lcccccc}
      \toprule
      Model                                      & Global         & Entity         & Attribute      & Relation       & Other          & Overall        \\
      \midrule
      SD v1.5~\cite{rombach2021highresolution}   & 74.63          & 74.23          & 75.39          & 73.49          & 67.81          & 63.18          \\
      PixArt-$\alpha$~\cite{chen2024pixartalpha} & 74.97          & 79.32          & 78.60          & 82.57          & 76.96          & 71.11          \\
      Lumina-Next~\cite{zhuo2024luminanext}      & 82.82          & 88.65          & 86.44          & 80.53          & 81.82          & 74.63          \\
      SDXL~\cite{podell2023sdxl}                 & 83.27          & 82.43          & 80.91          & 86.76          & 80.41          & 74.65          \\
      Playground v2.5~\cite{li2024playground}    & 83.06          & 82.59          & 81.20          & 84.08          & 83.50          & 75.47          \\
      HunyuanDiT~\cite{li2024hunyuandit}         & 84.59          & 80.59          & 88.01          & 74.36          & 86.41          & 78.87          \\
      Janus~\cite{wu2025janus}                   & 82.33          & 87.38          & 87.70          & 55.46          & 86.41          & 79.68          \\
      PixArt-$\Sigma$~\cite{chen2024pixartsigma} & 86.89          & 82.89          & 88.94          & 86.59          & 87.68          & 80.54          \\
      Emu3-Gen~\cite{wang2024emu3}               & 85.21          & 86.68          & 86.84          & 90.22          & 83.15          & 80.60          \\
      Janus-Pro-1B~\cite{chen2025janus}          & 87.58          & 88.63          & 88.17          & 88.98          & 88.30          & 82.63          \\
      DALL-E 3~\cite{openai2023dalle3}           & 90.97          & 89.61          & 88.39          & 90.58          & 89.83          & 83.50          \\
      FLUX.1 Dev~\cite{flux2024}                 & 74.35          & 90.00          & 88.96          & 90.87          & 88.33          & 83.84          \\
      SD3 Medium~\cite{esser2024scaling}         & 87.90          & 91.01          & 88.83          & 80.70          & 88.68          & 84.08          \\
      Janus-Pro-7B~\cite{chen2025janus}          & 86.90          & 88.90          & 89.40          & 89.32          & 89.48          & 84.19          \\
      HiDream-I1-Full~\cite{cai2025hidream}      & 76.44          & 90.22          & 89.48          & 93.74          & 91.83          & 85.89          \\
      Lumina-Image 2.0~\cite{lumina-image-2}     & ---            & 91.97          & 90.20          & \textbf{94.85} & ---            & 87.20          \\
      \midrule
      Seedream 3.0~\cite{gao2025seedream}        & \textbf{94.31} & 92.65          & 91.36          & 92.78          & 88.24          & 88.27          \\
      Qwen-Image~\cite{qwen-image}               & 91.32          & 91.56          & 92.02          & 94.31          & 92.73          & 88.32          \\
      GPT Image 1 High~\cite{gptimage}           & 88.89          & 88.94          & 89.84          & 92.63          & 90.96          & 85.15          \\
      \midrule
      \textbf{Nucleus-Image}                     & 85.10          & \textbf{93.08} & \textbf{92.20} & 93.56          & \textbf{93.62} & \textbf{88.79} \\
      \bottomrule
    \end{tabular}
  }
  \label{tab:dpg-results}
\end{table}

As shown in Table~\ref{tab:dpg-results}, Nucleus-Image achieves the highest overall score of 88.79, narrowly outperforming Qwen-Image (88.32) and Seedream 3.0 (88.27). The model ranks first in entity (93.08), attribute (92.20), and other (93.62) subcategories, reflecting its strong ability to ground fine-grained prompt conditions. Global consistency (85.10) is comparatively lower, suggesting room for improvement in holistic scene coherence.

\subsection{OneIG-Bench}
OneIG-Bench~\cite{chang2025oneig} is a multi-dimensional benchmark that evaluates five axes: alignment, text rendering, reasoning, style, and diversity. The final score is the mean of all five.

\begin{figure}[H]
  \centering
  \includegraphics[width=\linewidth,height=0.15\textheight,keepaspectratio]{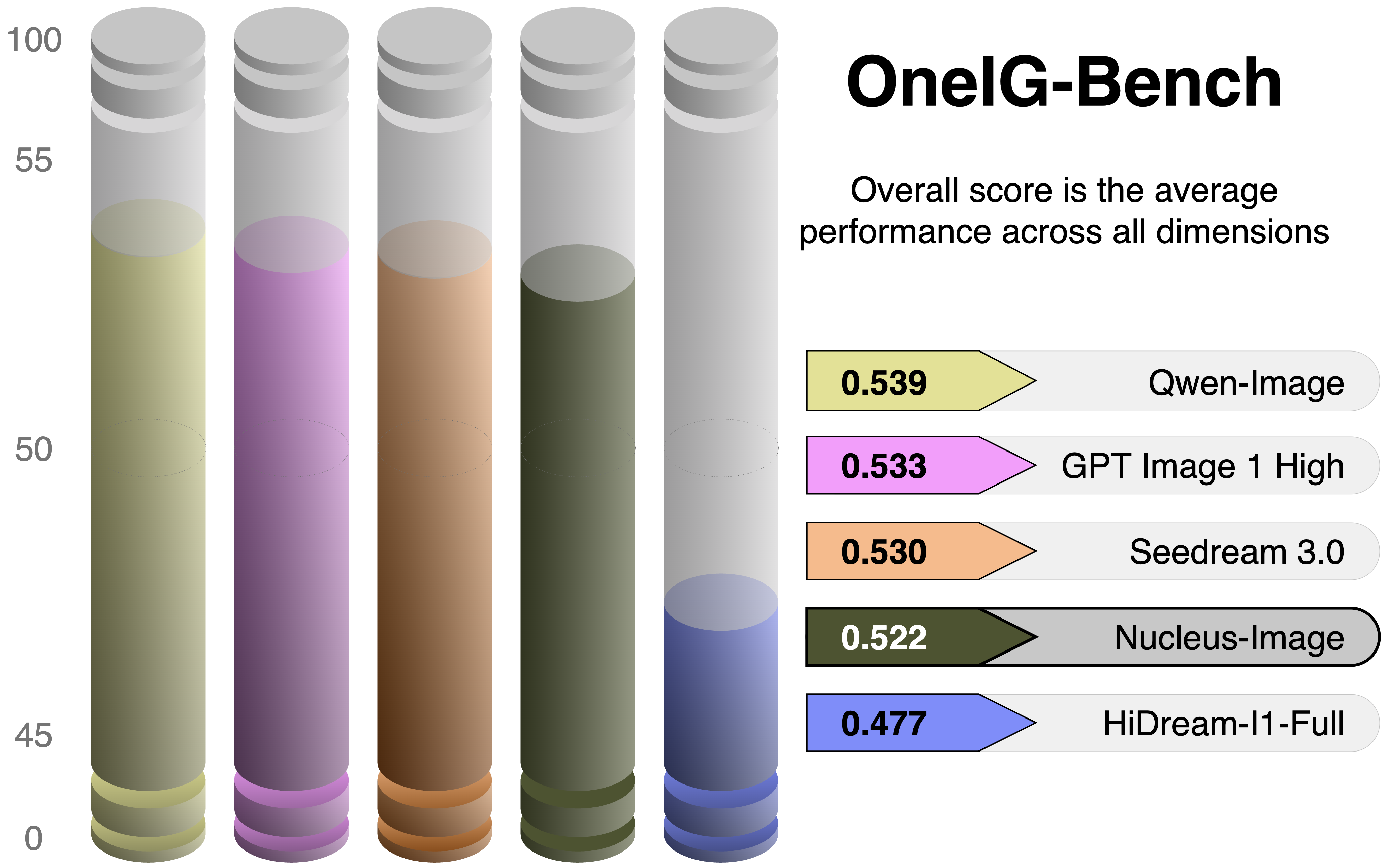}
  \caption{OneIG-Bench overall scores for top-performing models. Nucleus-Image scores 0.522, surpassing Imagen4 and Recraft V3, with particularly strong style and text rendering performance.}
  \label{fig:oneig-bench}
\end{figure}

\begin{table}[H]
  \caption{OneIG-Bench (EN) results.}
  \centering
  \scriptsize
  \resizebox{\textwidth}{!}{%
    \begin{tabular}{lcccccc}
      \toprule
      Model                                   & Alignment      & Text           & Reasoning      & Style          & Diversity      & Overall        \\
      \midrule
      Janus-Pro~\cite{chen2025janus}          & 0.553          & 0.001          & 0.139          & 0.276          & 0.365          & 0.267          \\
      BLIP3-o~\cite{chen2025blip3}            & 0.711          & 0.013          & 0.223          & 0.361          & 0.229          & 0.307          \\
      Show-o2-1.5B~\cite{xie2025show}         & 0.798          & 0.002          & 0.219          & 0.317          & 0.186          & 0.304          \\
      Show-o2-7B~\cite{xie2025show}           & 0.817          & 0.002          & 0.226          & 0.317          & 0.177          & 0.308          \\
      SD 1.5~\cite{rombach2021highresolution} & 0.565          & 0.010          & 0.207          & 0.383          & \textbf{0.429} & 0.319          \\
      SDXL~\cite{podell2023sdxl}              & 0.688          & 0.009          & 0.237          & 0.332          & 0.296          & 0.316          \\
      SANA-1.5 1.6B PAG~\cite{xie2025sana1d5} & 0.762          & 0.054          & 0.209          & 0.387          & 0.222          & 0.327          \\
      SANA-1.5 4.8B PAG~\cite{xie2025sana1d5} & 0.765          & 0.069          & 0.217          & 0.401          & 0.216          & 0.334          \\
      BAGEL~\cite{deng2025bagel}              & 0.769          & 0.044          & 0.173          & 0.367          & 0.251          & 0.361          \\
      BAGEL+CoT~\cite{deng2025bagel}          & 0.793          & 0.020          & 0.206          & 0.390          & 0.209          & 0.324          \\
      Lumina-Image 2.0~\cite{lumina-image-2}  & 0.819          & 0.106          & 0.270          & 0.354          & 0.216          & 0.353          \\
      Kolors 2.0~\cite{Kolors2}               & 0.820          & 0.427          & 0.262          & 0.360          & 0.300          & 0.434          \\
      FLUX.1 Dev~\cite{flux2024}              & 0.786          & 0.523          & 0.253          & 0.368          & 0.238          & 0.434          \\
      OmniGen2~\cite{wu2025omnigen2}          & 0.804          & 0.680          & 0.271          & 0.377          & 0.242          & 0.475          \\
      CogView4~\cite{Cogview4}                & 0.786          & 0.641          & 0.248          & 0.353          & 0.205          & 0.446          \\
      SD3.5 Large~\cite{esser2024scaling}     & 0.809          & 0.629          & 0.294          & 0.353          & 0.225          & 0.462          \\
      Imagen3~\cite{Imagen3}                  & 0.843          & 0.343          & 0.313          & 0.359          & 0.188          & 0.409          \\
      HiDream-I1-Full~\cite{cai2025hidream}   & 0.829          & 0.707          & 0.317          & 0.347          & 0.186          & 0.477          \\
      Recraft V3~\cite{recraftv3}             & 0.810          & 0.795          & 0.323          & 0.378          & 0.205          & 0.502          \\
      Imagen4~\cite{imagen4}                  & 0.857          & 0.805          & 0.338          & 0.377          & 0.199          & 0.515          \\
      \midrule
      Seedream 3.0~\cite{gao2025seedream}     & 0.818          & 0.865          & 0.275          & 0.413          & \textbf{0.277} & 0.530          \\
      Qwen-Image~\cite{qwen-image}            & \textbf{0.882} & \textbf{0.891} & 0.306          & 0.418          & 0.197          & \textbf{0.539} \\
      GPT Image 1 High~\cite{gptimage}        & 0.851          & 0.857          & \textbf{0.345} & \textbf{0.462} & 0.151          & 0.533          \\
      \midrule
      \textbf{Nucleus-Image}                  & 0.861          & 0.825          & 0.299          & 0.430          & 0.193          & 0.522          \\
      \bottomrule
    \end{tabular}
  }
  \label{tab:oneig-results}
\end{table}

As shown in Table~\ref{tab:oneig-results}, Nucleus-Image achieves an overall score of 0.522, placing it among the top tier of open and proprietary models and ahead of Imagen4 (0.515) and Recraft V3 (0.502). It performs strongest in style (0.430), where it surpasses Seedream 3.0 (0.413) and FLUX.1 Dev (0.368), and shows competitive text rendering (0.825), trailing only Qwen-Image (0.891) and Seedream 3.0 (0.865) in that category. Diversity (0.193) remains a known limitation relative to models like SD 1.5 and Seedream 3.0.

\subsection{Summary}
Table~\ref{tab:eval-summary} consolidates Nucleus-Image's performance across all three benchmarks. Averaging the normalized scores yields an overall score of 76.00, reflecting strong and consistent performance across compositional faithfulness (GenEval), dense prompt following (DPG-Bench), and multi-dimensional generation quality (OneIG-Bench). Nucleus-Image achieves state-of-the-art or near state-of-the-art results on all three benchmarks despite activating only $\sim$2B of its 17B parameters per forward pass, demonstrating that sparse MoE scaling is an effective path to high-quality text-to-image generation.

\begin{table}[H]
  \caption{Nucleus-Image benchmark summary.}
  \centering
  \begin{tabular}{lll}
    \toprule
    Benchmark                                  & Score (0--100) & Configuration    \\
    \midrule
    GenEval                                    & \textbf{87.00} & $1024\times1024$ \\
    DPG-Bench                                  & \textbf{88.79} & $1024\times1024$ \\
    OneIG-Bench                                & \textbf{52.20} & $1024\times1024$ \\
    Overall performance (avg. of normalized 3) & \textbf{76.00} & ---              \\
    \bottomrule
  \end{tabular}
  \label{tab:eval-summary}
\end{table}

\FloatBarrier

\section{Expert Allocation Analysis}
\label{sec:expert-analysis}

To understand how the expert-choice router distributes computation across spatial locations and denoising timesteps, we instrument the MoE layers at inference time. For each of the 29 MoE layers (layers 3--31, with layers 0--2 and the final layer kept dense), we record the gate logits, compute routing scores via softmax, and reconstruct the expert-choice top-$C$ selection. We then project the per-token statistics back onto the two-dimensional latent grid and bilinearly upsample them to pixel resolution, producing spatial heatmap overlays on the generated image. We analyze three representative generations spanning distinct visual domains: stylized text rendering, photorealistic scene composition, and portrait photography.

\subsection{Normalized Expert Allocation}

The \emph{normalized expert allocation} overlay visualizes, for each spatial position, the average number of experts that claim that token across all 29 MoE layers and all 50 denoising steps, normalized to $[0,1]$. High values (bright regions) indicate tokens that attract many competing experts; low values (dark regions) indicate tokens that require fewer experts.

Several patterns are consistent across the three images:
\begin{itemize}
  \item \textbf{Semantic saliency drives allocation.} Object boundaries, fine-grained textures, and compositionally important regions consistently receive higher expert counts. In the text image, the letter strokes and decorative vines are highlighted; in the helicopter scene, the aircraft body and rotor blades dominate; in the portrait, the eyes, lips, and hair boundary are the brightest regions.
  \item \textbf{Backgrounds are efficiently under-served.} Uniform or low-frequency regions---the white background in the text image, the sky in the helicopter scene, and the blurred wall in the portrait---consistently show low allocation values, indicating that the router avoids wasting expert capacity on tokens that carry little semantic information.
  \item \textbf{Edge and transition regions are prioritized.} The heatmaps trace object contours rather than filling entire object interiors, suggesting that the router identifies boundary tokens as requiring the most specialized processing.
\end{itemize}

\begin{figure}[t]
  \centering
  \setlength{\tabcolsep}{2pt}
  \begin{tabular}{ccc}
    \small Text Rendering & \small Scene Composition & \small Portrait \\[2pt]
    \includegraphics[width=0.31\linewidth]{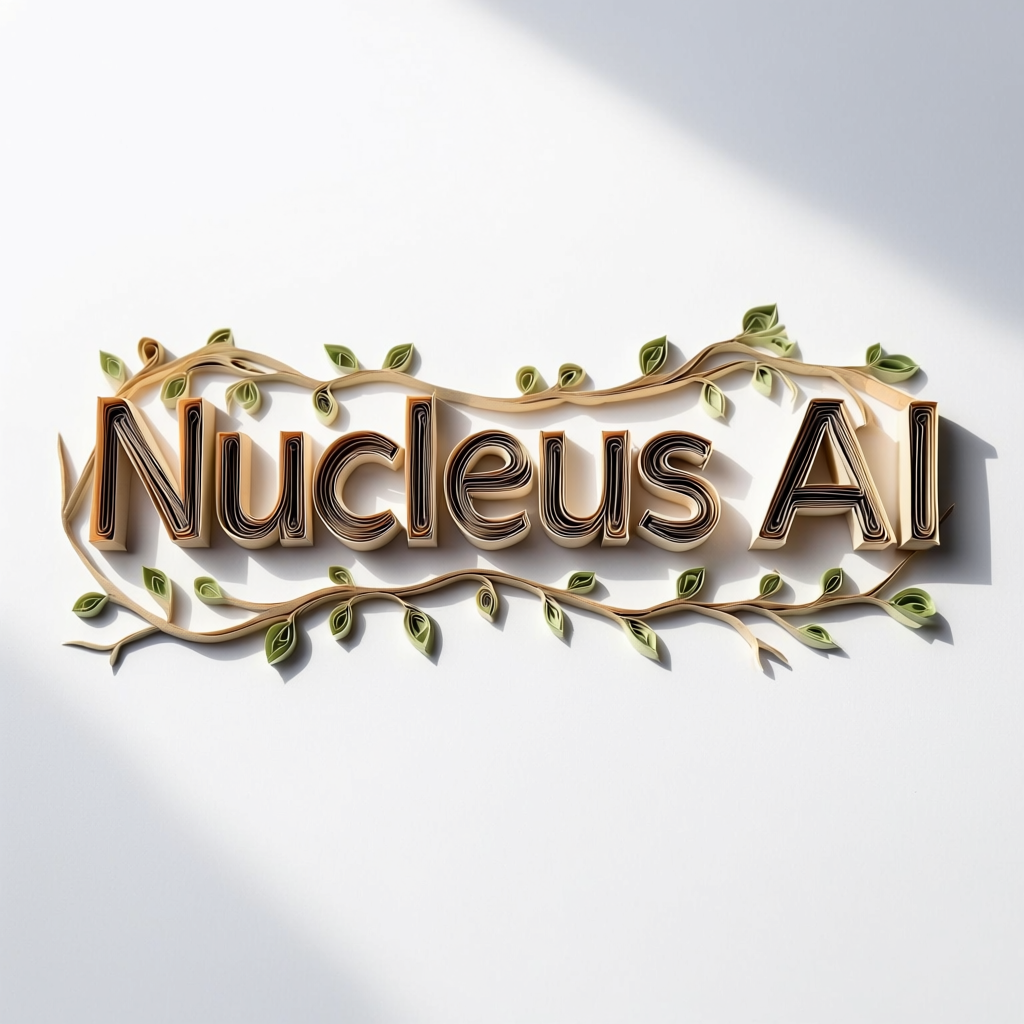} &
    \includegraphics[width=0.31\linewidth]{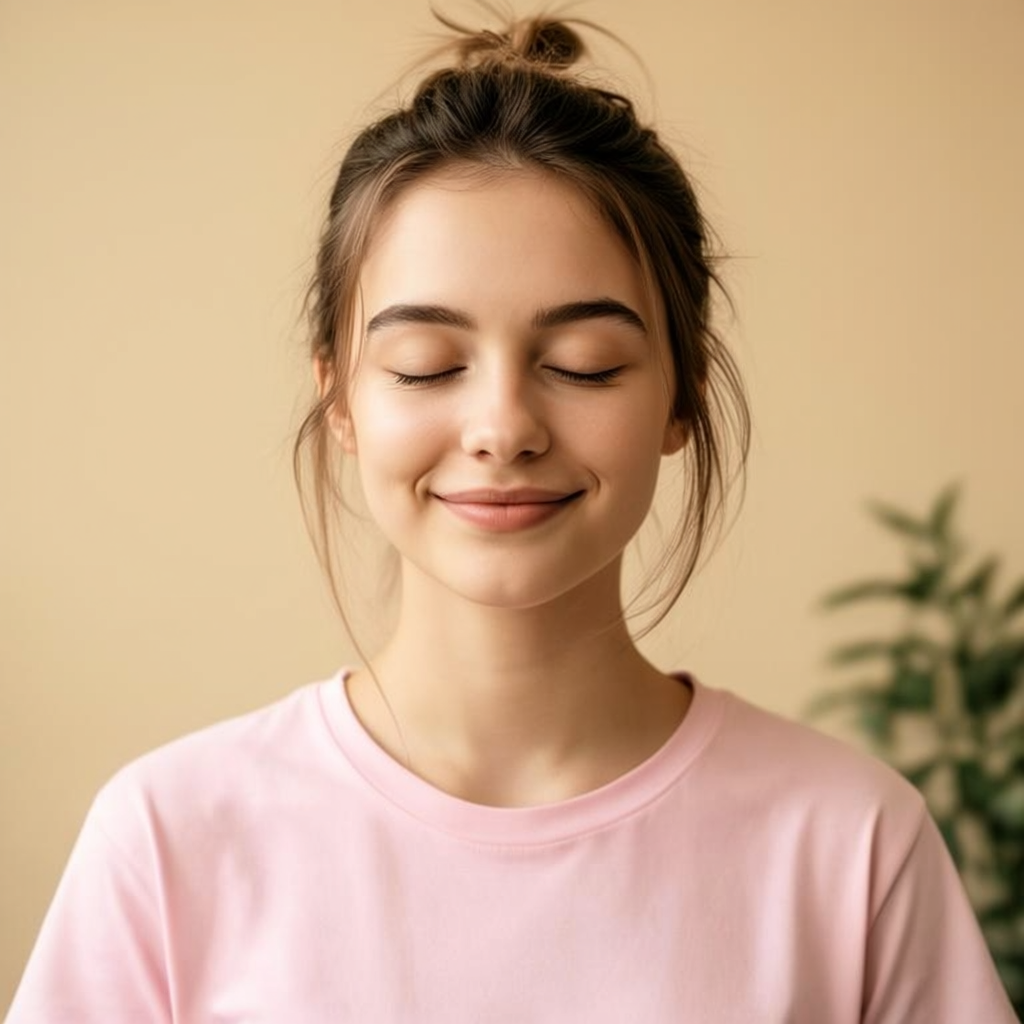} &
    \includegraphics[width=0.31\linewidth]{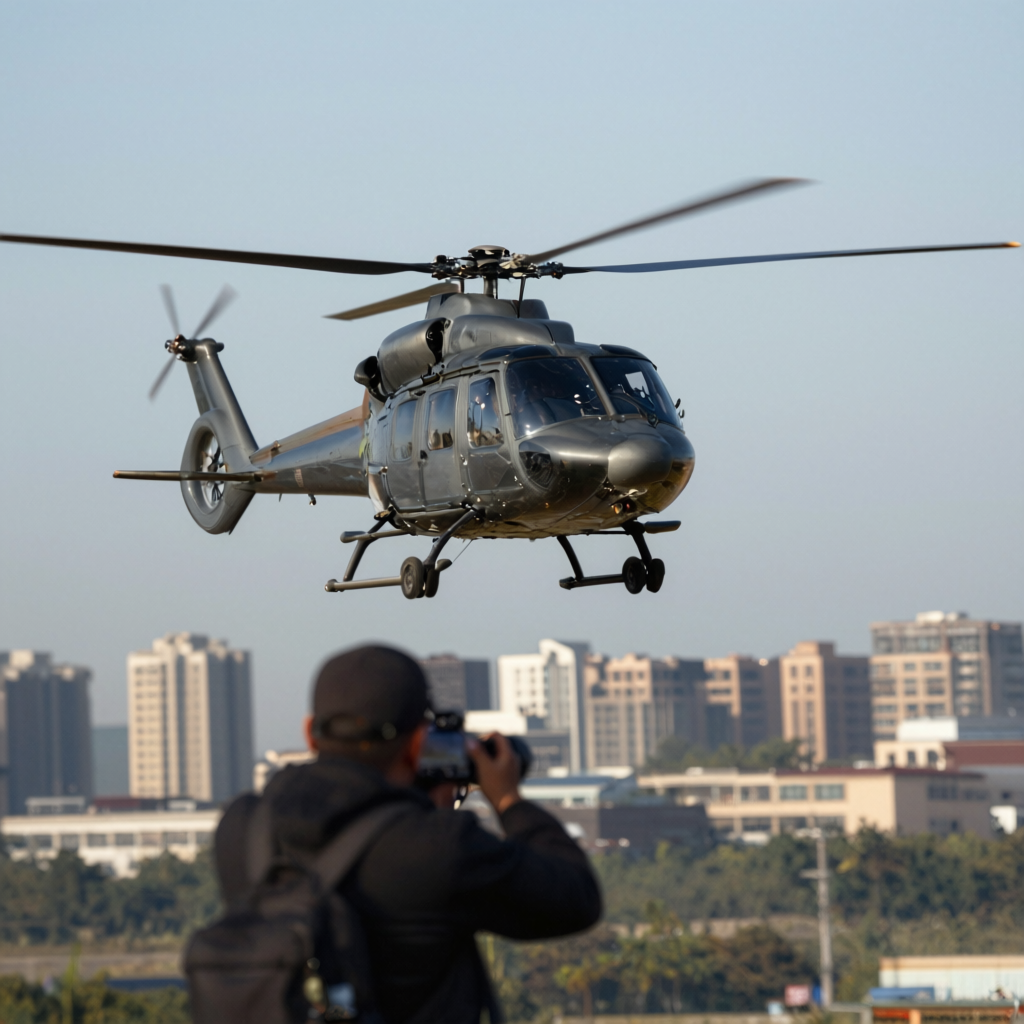} \\[2pt]
    \multicolumn{3}{c}{\textbf{Normalized Expert Allocation (\ensuremath{\uparrow})}} \\[2pt]
    \includegraphics[width=0.31\linewidth]{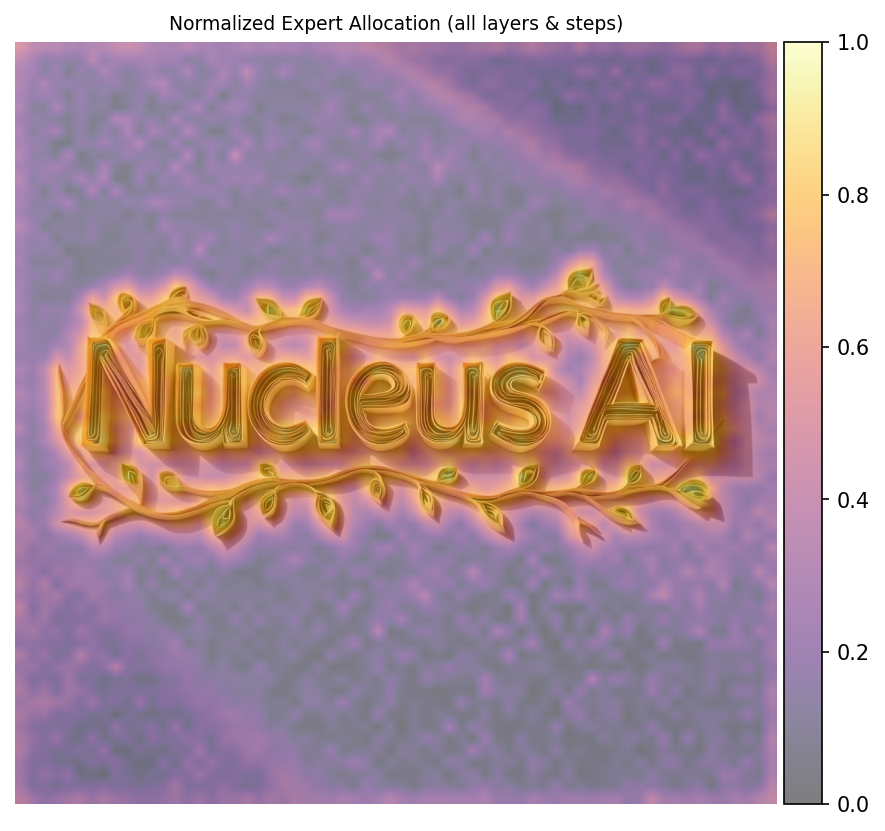} &
    \includegraphics[width=0.31\linewidth]{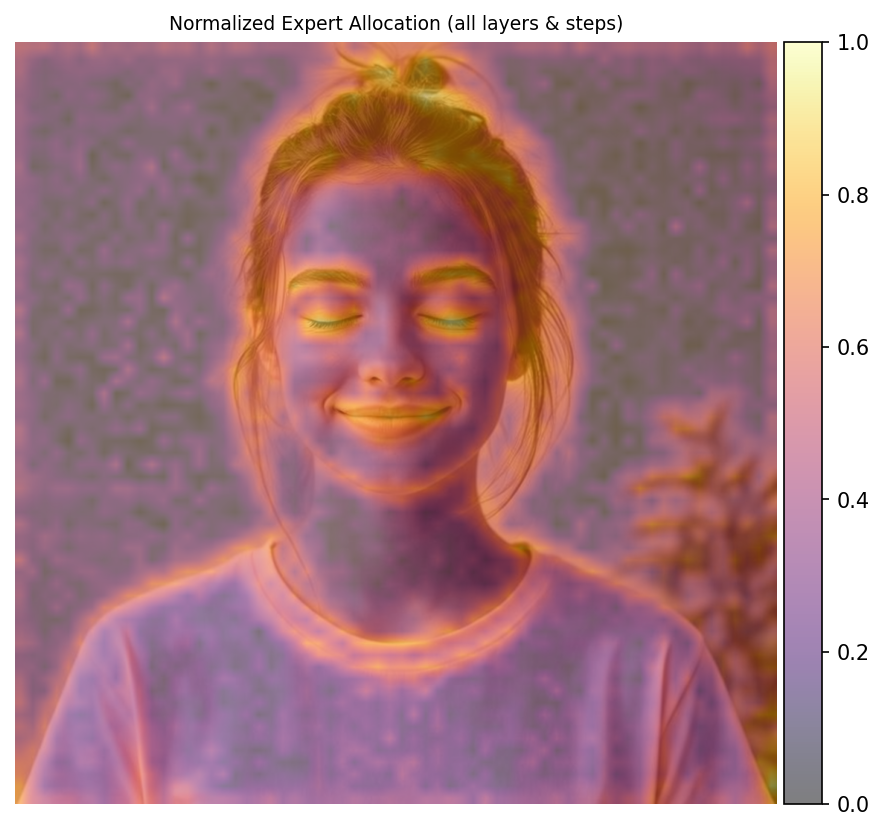} &
    \includegraphics[width=0.31\linewidth]{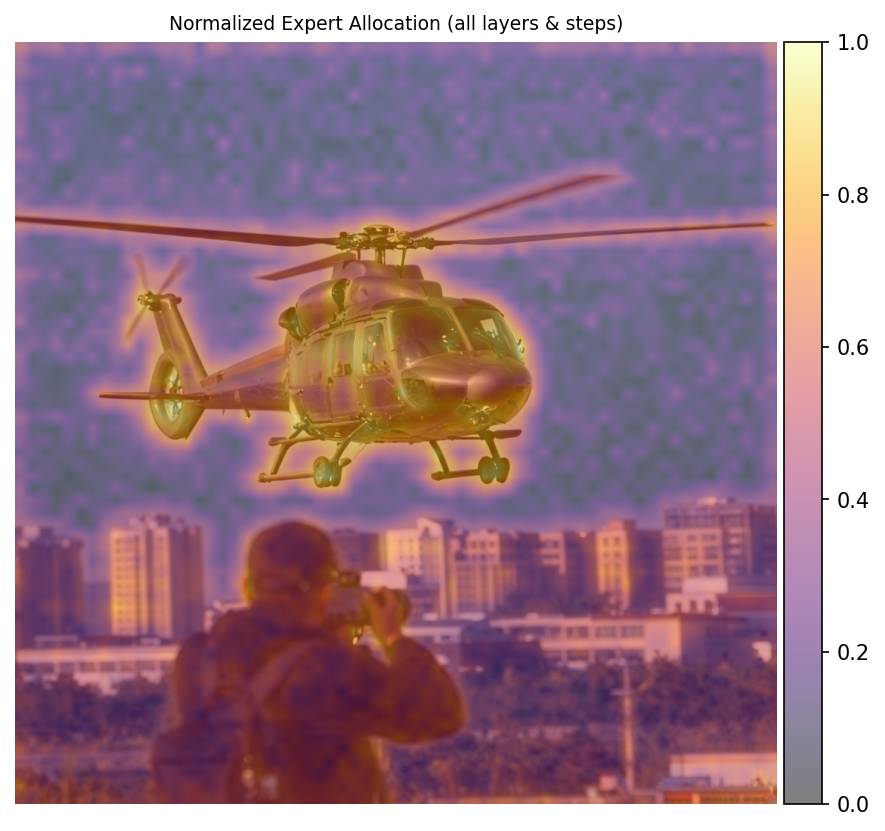} \\[2pt]
    \multicolumn{3}{c}{\textbf{Expert Diversity (\ensuremath{\downarrow})}} \\[2pt]
    \includegraphics[width=0.31\linewidth]{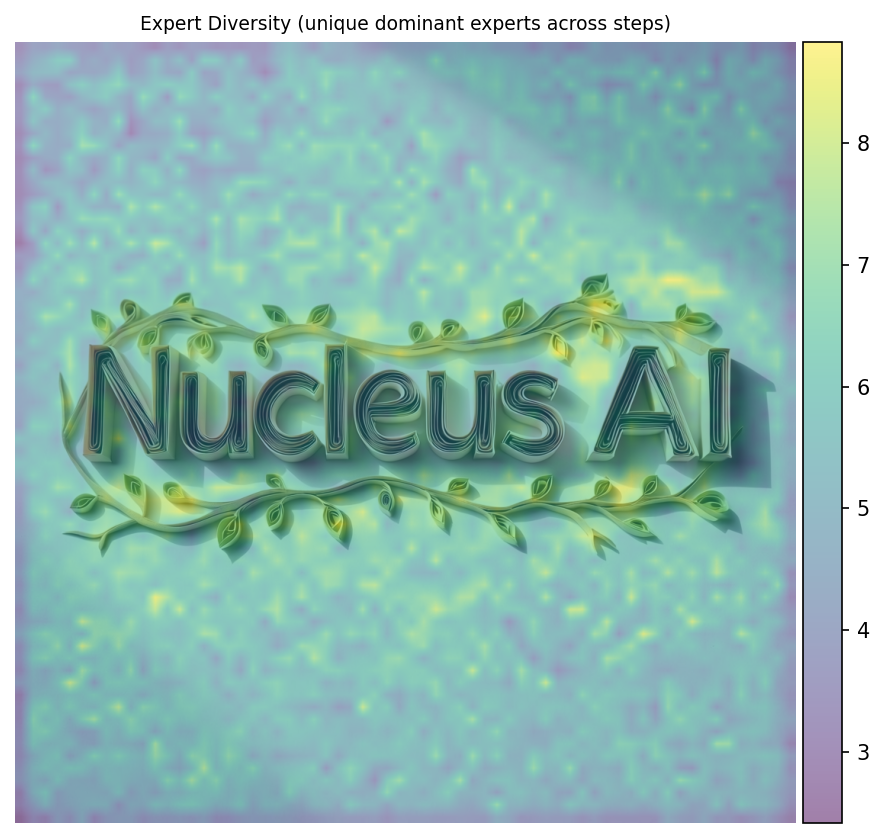} &
    \includegraphics[width=0.31\linewidth]{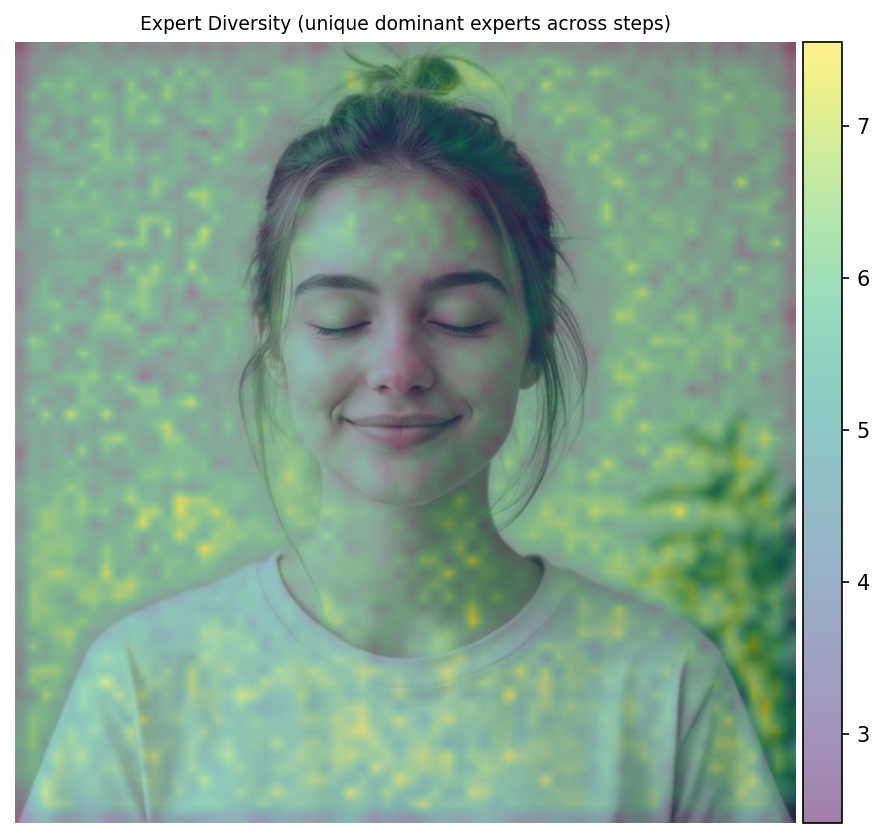} &
    \includegraphics[width=0.31\linewidth]{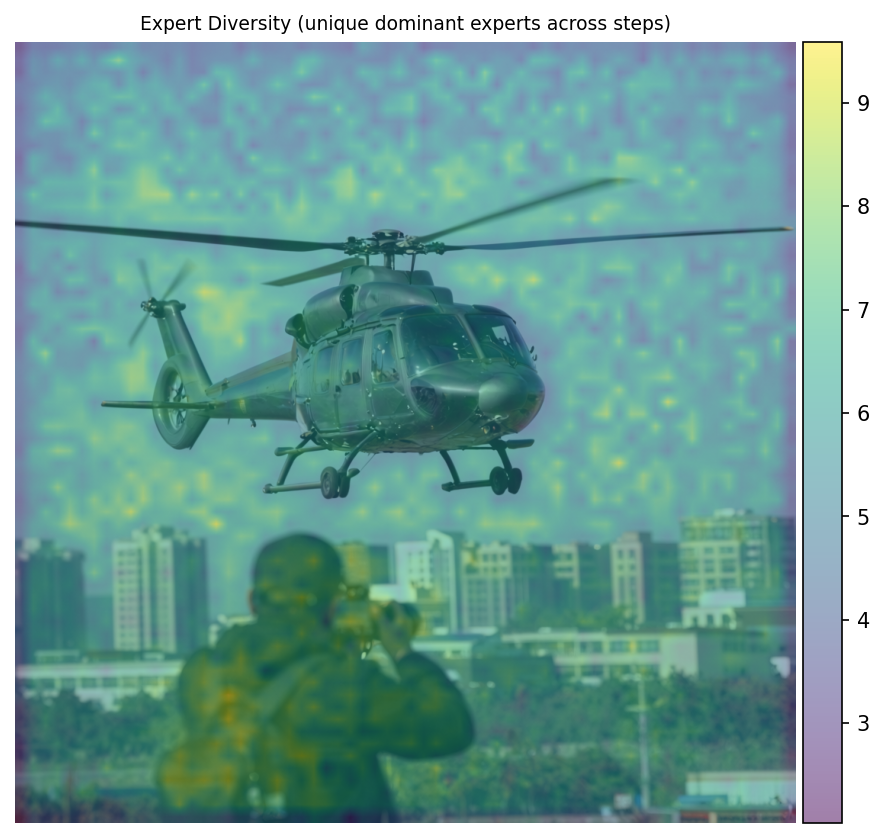} \\
  \end{tabular}
  \caption{\textbf{Expert allocation and diversity across three domains.} Each column shows a different generation: stylized text rendering (left), photorealistic scene composition (center), and portrait photography (right). \emph{Top:} generated images. \emph{Middle:} normalized expert allocation aggregated over all 29 MoE layers and 50 denoising steps---bright regions attract more expert capacity. \emph{Bottom:} expert diversity, measuring the number of unique dominant experts per spatial position across all timesteps. Across all three domains the router concentrates computation on semantically salient structures (letterforms, helicopter fuselage, facial features) while assigning fewer experts to uniform backgrounds. Foreground regions exhibit lower diversity (stable expert specialization), whereas backgrounds show higher diversity (fluctuating, uncommitted assignments).}
  \label{fig:expert-allocation-grid}
\end{figure}

\subsection{Expert Diversity}

The \emph{expert diversity} overlay (right column of Figure~\ref{fig:expert-allocation-grid}) measures, for each spatial position, how many unique experts serve as the dominant expert (highest softmax score) for that token across the 50 denoising steps. Low diversity means that the same expert consistently handles a token throughout the entire denoising trajectory; high diversity means that the dominant expert changes frequently as the latent evolves.

The diversity maps reveal a complementary pattern to the allocation overlays:
\begin{itemize}
  \item \textbf{Foreground regions show lower diversity.} In all three images, the primary subjects---text and vines, helicopter, face---tend toward the lower end of the diversity scale. This indicates that the router assigns a stable set of specialist experts to semantically important tokens, maintaining consistent processing throughout denoising.
  \item \textbf{Backgrounds show higher diversity.} Sky, walls, and uniform surfaces exhibit the highest diversity values, meaning their dominant expert changes frequently across timesteps. Because these regions carry little structural information, no single expert specializes for them, and the dominant assignment fluctuates based on noise-level-dependent routing cues.
  \item \textbf{The diversity range is moderate (3--9 unique experts out of 50 steps).} This confirms that routing is neither degenerate (always the same expert) nor chaotic (a different expert every step), but rather exhibits structured temporal coherence with gradual transitions as the denoising trajectory progresses.
\end{itemize}

\subsection{Temporal Evolution of Expert Routing}

To study how expert allocation evolves over the denoising trajectory, we visualize the number-of-experts-per-token heatmap at layer~17 (a mid-depth MoE layer) across 10 evenly sampled timesteps from $t{=}0$ (pure noise) to $t{=}49$ (clean image).

\begin{figure}[t]
  \centering

  \begin{minipage}[b]{\linewidth}
    \centering
    \includegraphics[width=0.95\linewidth]{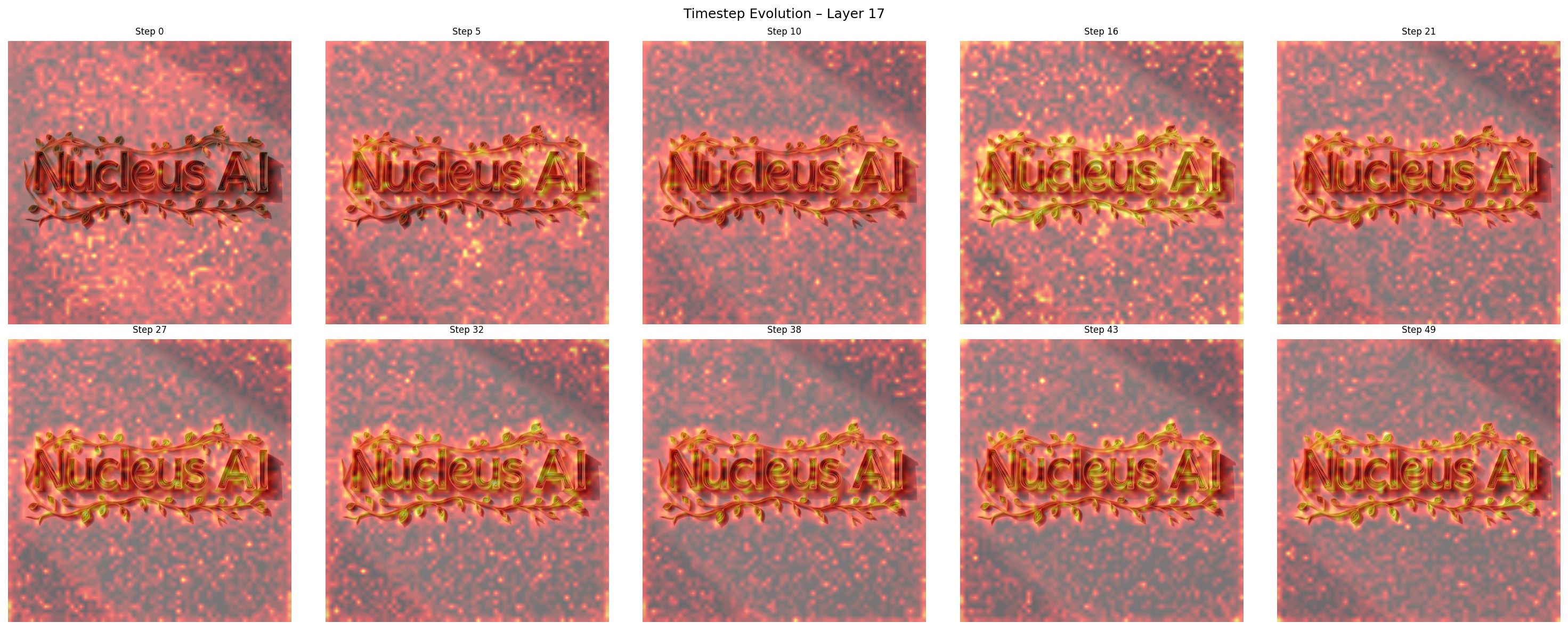}
    \subcaption{Stylized text rendering}
    \label{fig:ts-text}
  \end{minipage}

  \vspace{0.2cm}

  \begin{minipage}[b]{\linewidth}
    \centering
    \includegraphics[width=0.95\linewidth]{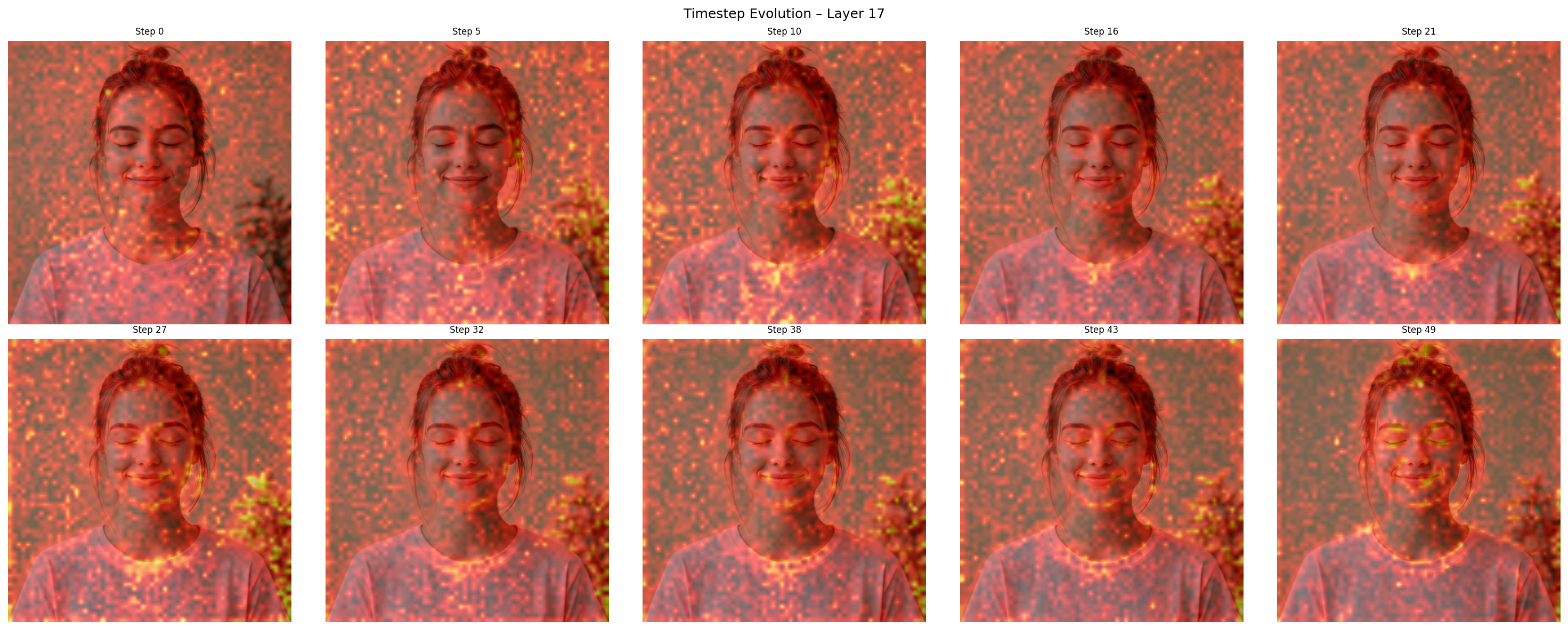}
    \subcaption{Photorealistic scene composition}
    \label{fig:ts-heli}
  \end{minipage}

  \vspace{0.2cm}

  \begin{minipage}[b]{\linewidth}
    \centering
    \includegraphics[width=0.95\linewidth]{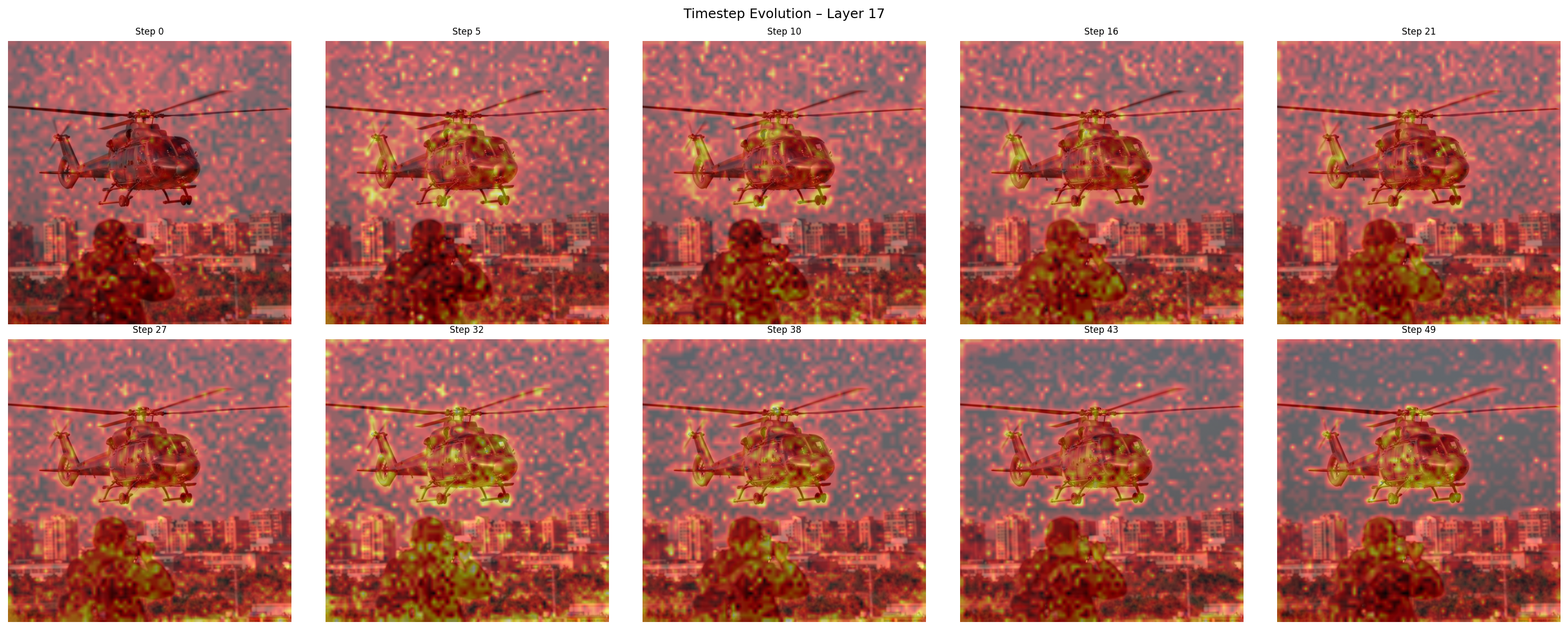}
    \subcaption{Portrait photography}
    \label{fig:ts-portrait}
  \end{minipage}

  \caption{\textbf{Timestep evolution of expert allocation at layer~17.} Each row shows the number-of-experts-per-token heatmap overlaid on the generated image, sampled at 10 evenly spaced denoising steps (left to right, top to bottom: steps 0, 5, 11, 17, 22, 27, 33, 38, 44, 49). Early steps exhibit diffuse, spatially unstructured allocation; mid-steps develop coarse semantic structure; late steps produce the sharpest, most content-aligned allocation maps.}
  \label{fig:timestep-evolution}
\end{figure}

The temporal progression exhibits a consistent three-phase pattern across all three images:

\begin{enumerate}
  \item \textbf{Early steps ($t < 10$):} The allocation heatmap is relatively uniform with scattered hot spots. At this stage the latent is dominated by noise and the router cannot yet distinguish meaningful spatial structure, so expert assignment is diffuse and weakly correlated with the final image content.

  \item \textbf{Mid steps ($10 \leq t < 35$):} Coarse spatial structure emerges in the heatmap. The router begins to differentiate foreground from background---in the text image, letter-shaped regions start to concentrate experts; in the helicopter scene, the aircraft silhouette becomes visible in the allocation map; in the portrait, the face outline appears. This phase coincides with the formation of large-scale compositional structure in the latent.

  \item \textbf{Late steps ($t \geq 35$):} The heatmap becomes sharpest and most strongly correlated with semantic content. Expert allocation tightly follows object boundaries, fine details (rotor blades, individual letters, eyelashes), and texture transitions. The router has effectively learned to route computational resources to the regions that benefit most from specialized expert processing during the final refinement phase.
\end{enumerate}

This progression confirms that the timestep-conditioned router input (which concatenates the timestep embedding with the unmodulated hidden state) enables the router to adapt its allocation strategy as a function of the denoising stage. The router does not merely rely on a static spatial prior but dynamically reallocates expert capacity as the signal-to-noise ratio of the latent improves.

The expert allocation analysis reveals that the expert-choice routing mechanism in Nucleus-Image exhibits three desirable properties:
\begin{enumerate}
  \item \textbf{Content-aware sparsity:} Experts concentrate on semantically salient regions (object boundaries, fine details, text) while efficiently under-serving uniform backgrounds, achieving adaptive compute allocation without explicit spatial supervision.
  \item \textbf{Temporal coherence:} Individual experts maintain stable specialization for specific spatial tokens across the denoising trajectory, as evidenced by moderate diversity scores, while still permitting smooth transitions as the latent content evolves.
  \item \textbf{Noise-adaptive routing:} The allocation pattern transitions from diffuse and unstructured at high noise levels to sharp and semantically aligned at low noise levels, demonstrating that the timestep-conditioned gate successfully modulates routing decisions based on the current stage of the denoising process.
\end{enumerate}

These properties arise naturally from the combination of expert-choice routing, timestep-conditioned gate inputs, and the load-balancing losses described in the Loss Functions section, without requiring any explicit spatial or semantic routing supervision.

\section{Conclusion}
Nucleus-Image demonstrates that sparse mixture-of-experts scaling is a viable and efficient path to high-quality text-to-image generation. By activating only approximately 2B of its 17B total parameters per forward pass, the model matches or exceeds models with significantly larger active parameter budgets on GenEval, DPG-Bench, and OneIG-Bench. To our knowledge, Nucleus-Image is the first openly available MoE-based diffusion model to reach this quality tier, and it does so entirely through pre-training---without any reinforcement learning, direct preference optimization, or human preference tuning.

This report details several contributions that collectively enable this result. On the architecture side, we introduce decoupled routing that separates timestep-aware expert assignment from timestep-conditioned computation, preventing modulation-scale collapse in MoE diffusion transformers. Text tokens are excluded from the transformer backbone entirely---participating only as key-value contributors in joint attention---which eliminates MoE routing overhead for text and enables full text KV caching across denoising steps. On the training side, we pair the Muon optimizer with a parameter grouping recipe tailored for diffusion MoE, apply orthogonal regularization on router weights to maintain expert diversity, and adopt progressive sparsification of the capacity factor across resolution stages. The Warmup-Stable-Merge learning rate schedule replaces online EMA with post-hoc checkpoint merging, eliminating the memory cost of a shadow weight copy while preserving the flexibility to extend training without committing to a decay schedule. On the systems side, custom Triton kernels fuse the repeated modulation-normalization-residual patterns into single-pass operations, eliminating memory-bandwidth bottlenecks in expert computation under expert parallelism. On the data side, we construct a 700M-image corpus with 1.5B multi-granularity caption pairs and train with multi-aspect-ratio bucketing from the outset at every resolution stage.

By pushing simultaneously on architecture, optimization, data curation, and systems-level efficiency, Nucleus-Image shows that sparse MoE is not merely a parameter-efficient alternative but a highly effective scaling axis for image generation and diffusion models in general. To accelerate progress in efficient image generation, we publicly release the model weights, training code, and dataset alongside this report.

\section*{Authors}
Core Contributors\footnote{Equal contribution}: Chandan Akiti, Sai Ajay Modukuri, Murali Nandan Nagarapu

Contributors: Gunavardhan Akiti\footnote{CMU, work done while interning at NucleusAI}, Haozhe Liu\footnote{KAUST}

\section*{Acknowledgments}
This work was supported in part by the NucleusAI research and infrastructure teams.
We thank Mithril Cloud~\cite{mithrilcloud} for providing us with compute resources for extensive experiments, and Google for their support of the main training run.
We are grateful to NVIDIA for the NeMo Curator~\cite{nemocurator} toolkit, which we used for large-scale data preprocessing, and NVIDIA DALI~\cite{dali} for efficient data loading.
We also thank the TorchTitan~\cite{torchtitan2024} project for their Mixture-of-Experts training recipes.

\bibliographystyle{unsrt}
\bibliography{references}

@article{zhou2022expertchoice,
  title={Mixture-of-experts with expert choice routing},
  author={Zhou, Yanqi and Lei, Tao and Liu, Hanxiao and Du, Nan and Huang, Yanping and Zhao, Vincent and Dai, Andrew M and Le, Quoc V and Laudon, James and others},
  journal={Advances in Neural Information Processing Systems},
  volume={35},
  pages={7103--7114},
  year={2022},
}

@article{ainslie2023gqa,
  title={{GQA}: Training Generalized Multi-Query Transformer Models from Multi-Head Checkpoints},
  author={Ainslie, Joshua and Lee-Thorp, James and de Jong, Michiel and Zemlyanskiy, Yury and Lebr{\'o}n, Federico and Sanghai, Sumit},
  journal={arXiv:2305.13245},
  year={2023},
}

@article{shazeer2020glu,
  title={{GLU} Variants Improve Transformer},
  author={Shazeer, Noam},
  journal={arXiv:2002.05202},
  year={2020},
}

@article{ho2022cfg,
  title={Classifier-Free Diffusion Guidance},
  author={Ho, Jonathan and Salimans, Tim},
  journal={arXiv:2207.12598},
  year={2022},
}

@inproceedings{peebles2023dit,
  title={Scalable Diffusion Models with Transformers},
  author={Peebles, William and Xie, Saining},
  booktitle={ICCV},
  year={2023},
}

@misc{deepep2025,
  title={{DeepEP}: an efficient expert-parallel communication library},
  author={{DeepSeek-AI}},
  year={2025},
  url={https://github.com/deepseek-ai/DeepEP},
}

@misc{sun2024ecdit,
  title={{EC-DIT}: Scaling Diffusion Transformers with Adaptive Expert-Choice Routing},
  author={Sun, Haotian and Lei, Tao and Zhang, Bowen and Li, Yanghao and Huang, Haoshuo and Pang, Ruoming and Dai, Bo and Du, Nan},
  year={2024},
  eprint={2410.02098},
  archivePrefix={arXiv},
  primaryClass={cs.CV},
  url={https://arxiv.org/abs/2410.02098}
}

@misc{fei2024ditmoe,
  title={Scaling Diffusion Transformers to 16 Billion Parameters},
  author={Fei, Zhengcong and Fan, Mingyuan and Yu, Changqian and Li, Debang and Huang, Junshi},
  year={2024},
  eprint={2407.11633},
  archivePrefix={arXiv},
  primaryClass={cs.CV},
  url={https://arxiv.org/abs/2407.11633}
}

@misc{jordan2024muon,
  title={Muon: An optimizer for hidden layers in neural networks},
  author={Jordan, Keller},
  year={2024},
  url={https://github.com/KellerJordan/Muon}
}

@misc{liu2025muonscalable,
  title={Muon is Scalable for {LLM} Training},
  author={Liu, Jingyuan and Zeng, Jianlin and others},
  year={2025},
  eprint={2502.16982},
  archivePrefix={arXiv},
  primaryClass={cs.LG},
  url={https://arxiv.org/abs/2502.16982}
}

@misc{torchtitan2024,
  title={{TorchTitan}: One-stop {PyTorch} native solution for production ready {LLM} pre-training},
  author={Huang, Wanchao and Luk, Zuchao and Bl\"{o}baum, Patrick and Zeng, Shiyang and Ge, Tian and Deng, Peng and Chauhan, Himanshu and Li, Jian and Lim, Deven and Lai, Helen and Deng, Will and Bom, Vignesh and Roh, Boyuan and others},
  year={2024},
  eprint={2410.06511},
  archivePrefix={arXiv},
  primaryClass={cs.LG},
  url={https://arxiv.org/abs/2410.06511}
}

@misc{hsu2024liger,
  title={Liger Kernel: Efficient Triton Kernels for {LLM} Training},
  author={Pin-Lun Hsu and Yun Dai and Vignesh Kothapalli and Qingquan Song and Shao Tang and Siyu Zhu and Steven Shimizu and Shivam Sahni and Haowen Ning and Yanning Chen},
  year={2024},
  eprint={2410.10989},
  archivePrefix={arXiv},
  primaryClass={cs.LG},
  url={https://arxiv.org/abs/2410.10989}
}

@article{tian2025wsm,
  title={{WSM}: Decay-Free Learning Rate Schedule via Checkpoint Merging for {LLM} Pre-training},
  author={Tian, Changxin and Wang, Peng and others},
  journal={arXiv preprint arXiv:2507.17634},
  year={2025}
}

@misc{radford2021learning,
  title={Learning Transferable Visual Models From Natural Language Supervision},
  author={Alec Radford and Jong Wook Kim and Chris Hallacy and Aditya Ramesh and Gabriel Goh and Sandhini Agarwal and Girish Sastry and Amanda Askell and Pamela Mishkin and Jack Clark and Gretchen Krueger and Ilya Sutskever},
  year={2021},
  eprint={2103.00020},
  archivePrefix={arXiv},
  primaryClass={cs.CV},
  url={https://arxiv.org/abs/2103.00020}
}

@misc{metaclip,
      title={Meta CLIP 2: A Worldwide Scaling Recipe}, 
      author={Yung-Sung Chuang and Yang Li and Dong Wang and Ching-Feng Yeh and Kehan Lyu and Ramya Raghavendra and James Glass and Lifei Huang and Jason Weston and Luke Zettlemoyer and Xinlei Chen and Zhuang Liu and Saining Xie and Wen-tau Yih and Shang-Wen Li and Hu Xu},
      year={2025},
      eprint={2507.22062},
      archivePrefix={arXiv},
      primaryClass={cs.CV},
      url={https://arxiv.org/abs/2507.22062}, 
}

@misc{hinz2020,
      title={Improved Techniques for Training Single-Image GANs}, 
      author={Tobias Hinz and Matthew Fisher and Oliver Wang and Stefan Wermter},
      year={2020},
      eprint={2003.11512},
      archivePrefix={arXiv},
      primaryClass={cs.CV},
      url={https://arxiv.org/abs/2003.11512}, 
}

@misc{ren2025,
      title={Unveiling and Mitigating Memorization in Text-to-image Diffusion Models through Cross Attention}, 
      author={Jie Ren and Yaxin Li and Shenglai Zeng and Han Xu and Lingjuan Lyu and Yue Xing and Jiliang Tang},
      year={2025},
      eprint={2403.11052},
      archivePrefix={arXiv},
      primaryClass={cs.CV},
      url={https://arxiv.org/abs/2403.11052}, 
}

@misc{zhang2025,
      title={CAPTAIN: Semantic Feature Injection for Memorization Mitigation in Text-to-Image Diffusion Models}, 
      author={Tong Zhang and Carlos Hinojosa and Bernard Ghanem},
      year={2025},
      eprint={2512.10655},
      archivePrefix={arXiv},
      primaryClass={cs.AI},
      url={https://arxiv.org/abs/2512.10655}, 
}

@misc{sugiura2025,
      title={WAON: Large-Scale and High-Quality Japanese Image-Text Pair Dataset for Vision-Language Models}, 
      author={Issa Sugiura and Shuhei Kurita and Yusuke Oda and Daisuke Kawahara and Yasuo Okabe and Naoaki Okazaki},
      year={2025},
      eprint={2510.22276},
      archivePrefix={arXiv},
      primaryClass={cs.CV},
      url={https://arxiv.org/abs/2510.22276}, 
}

@misc{taylor2026,
      title={The Algorithmic Gaze of Image Quality Assessment: An Audit and Trace Ethnography of the LAION-Aesthetics Predictor}, 
      author={Jordan Taylor and William Agnew and Maarten Sap and Sarah E. Fox and Haiyi Zhu},
      year={2026},
      eprint={2601.09896},
      archivePrefix={arXiv},
      primaryClass={cs.HC},
      url={https://arxiv.org/abs/2601.09896}, 
}

@misc{lumina-image-2,
      title={Lumina-Image 2.0: A Unified and Efficient Image Generative Framework}, 
      author={Qi Qin and Le Zhuo and Yi Xin and Ruoyi Du and Zhen Li and Bin Fu and Yiting Lu and Jiakang Yuan and Xinyue Li and Dongyang Liu and Xiangyang Zhu and Manyuan Zhang and Will Beddow and Erwann Millon and Victor Perez and Wenhai Wang and Conghui He and Bo Zhang and Xiaohong Liu and Hongsheng Li and Yu Qiao and Chang Xu and Peng Gao},
      year={2025},
      eprint={2503.21758},
      archivePrefix={arXiv},
      primaryClass={cs.CV},
      url={https://arxiv.org/abs/2503.21758}, 
}

@misc{playground-v3,
      title={Playground v3: Improving Text-to-Image Alignment with Deep-Fusion Large Language Models}, 
      author={Bingchen Liu and Ehsan Akhgari and Alexander Visheratin and Aleks Kamko and Linmiao Xu and Shivam Shrirao and Chase Lambert and Joao Souza and Suhail Doshi and Daiqing Li},
      year={2024},
      eprint={2409.10695},
      archivePrefix={arXiv},
      primaryClass={cs.CV},
      url={https://arxiv.org/abs/2409.10695}, 
}

@misc{qwen-image,
      title={Qwen-Image Technical Report}, 
      author={Chenfei Wu and Jiahao Li and Jingren Zhou and Junyang Lin and Kaiyuan Gao and Kun Yan and Sheng-ming Yin and Shuai Bai and Xiao Xu and Yilei Chen and Yuxiang Chen and Zecheng Tang and Zekai Zhang and Zhengyi Wang and An Yang and Bowen Yu and Chen Cheng and Dayiheng Liu and Deqing Li and Hang Zhang and Hao Meng and Hu Wei and Jingyuan Ni and Kai Chen and Kuan Cao and Liang Peng and Lin Qu and Minggang Wu and Peng Wang and Shuting Yu and Tingkun Wen and Wensen Feng and Xiaoxiao Xu and Yi Wang and Yichang Zhang and Yongqiang Zhu and Yujia Wu and Yuxuan Cai and Zenan Liu},
      year={2025},
      eprint={2508.02324},
      archivePrefix={arXiv},
      primaryClass={cs.CV},
      url={https://arxiv.org/abs/2508.02324}, 
}

@article{wan2025wan,
  title={Wan: Open and advanced large-scale video generative models},
  author={Wan, Team and Wang, Ang and Ai, Baole and Wen, Bin and Mao, Chaojie and Xie, Chen-Wei and Chen, Di and Yu, Feiwu and Zhao, Haiming and Yang, Jianxiao and others},
  journal={arXiv preprint arXiv:2503.20314},
  year={2025}
}

@article{yang2025qwen3,
  title={Qwen3 technical report},
  author={Yang, An and Li, Anfeng and Yang, Baosong and Zhang, Beichen and Hui, Binyuan and Zheng, Bo and Yu, Bowen and Gao, Chang and Huang, Chengen and Lv, Chenxu and others},
  journal={arXiv preprint arXiv:2505.09388},
  year={2025}
}

@article{liu2022flow,
  title={Flow straight and fast: Learning to generate and transfer data with rectified flow},
  author={Liu, Xingchao and Gong, Chengyue and Liu, Qiang},
  journal={arXiv preprint arXiv:2209.03003},
  year={2022}
}

@article{rafailov2023direct,
  title={Direct preference optimization: Your language model is secretly a reward model},
  author={Rafailov, Rafael and Sharma, Archit and Mitchell, Eric and Manning, Christopher D and Ermon, Stefano and Finn, Chelsea},
  journal={Advances in neural information processing systems},
  volume={36},
  pages={53728--53741},
  year={2023}
}

@article{zhao2023pytorch,
  title={Pytorch fsdp: experiences on scaling fully sharded data parallel},
  author={Zhao, Yanli and Gu, Andrew and Varma, Rohan and Luo, Liang and Huang, Chien-Chin and Xu, Min and Wright, Less and Shojanazeri, Hamid and Ott, Myle and Shleifer, Sam and others},
  journal={arXiv preprint arXiv:2304.11277},
  year={2023}
}

@inproceedings{esser2024scaling,
  title={Scaling Rectified Flow Transformers for High-Resolution Image Synthesis},
  author={Esser, Patrick and Kulal, Sumith and Blattmann, Andreas and Entezari, Rahim and M{\"u}ller, Jonas and Saini, Harry and Levi, Yam and Lorenz, Dominik and Sauer, Axel and Boesel, Frederic and others},
  booktitle={ICML},
  year={2024}
}

@article{zhang2019root,
  title={Root mean square layer normalization},
  author={Zhang, Biao and Sennrich, Rico},
  journal={Advances in neural information processing systems},
  volume={32},
  year={2019}
}

@misc{dali,
  title={{NVIDIA DALI}: A GPU-accelerated data loading library},
  author={Nvidia},
  year={2024},
  url={https://github.com/NVIDIA/DALI}
}

@misc{li2024hunyuandit,
      title={Hunyuan-DiT: A Powerful Multi-Resolution Diffusion Transformer with Fine-Grained Chinese Understanding}, 
      author={Zhimin Li and Jianwei Zhang and Qin Lin and Jiangfeng Xiong and Yanxin Long and Xinchi Deng and Yingfang Zhang and Xingchao Liu and Minbin Huang and Zedong Xiao and Dayou Chen and Jiajun He and Jiahao Li and Wenyue Li and Chen Zhang and Rongwei Quan and Jianxiang Lu and Jiabin Huang and Xiaoyan Yuan and Xiaoxiao Zheng and Yixuan Li and Jihong Zhang and Chao Zhang and Meng Chen and Jie Liu and Zheng Fang and Weiyan Wang and Jinbao Xue and Yangyu Tao and Jianchen Zhu and Kai Liu and Sihuan Lin and Yifu Sun and Yun Li and Dongdong Wang and Mingtao Chen and Zhichao Hu and Xiao Xiao and Yan Chen and Yuhong Liu and Wei Liu and Di Wang and Yong Yang and Jie Jiang and Qinglin Lu},
      year={2024},
      eprint={2405.08748},
      archivePrefix={arXiv},
      primaryClass={cs.CV}
}

@misc{nemocurator,
  title={{NeMo Curator}: {GPU}-Accelerated Data Curation for Large Language Models},
  author={Jennings, Joseph and Bhandwaldar, Mostofa Patwary and Elazar, Vibhu Jawa and Ryan, Ayush Dattagupta and Zeng, Jiwei Liu and Nithin, Shankar Rao and Casper, Jared and Gonzalez, Ashwath Aithal and others},
  year={2024},
  eprint={2407.10638},
  archivePrefix={arXiv},
  primaryClass={cs.CL},
  url={https://github.com/NVIDIA-NeMo/Curator}
}

@misc{mithrilcloud,
  title={{Mithril Cloud}},
  author={{Mithril Cloud}},
  year={2025},
  url={https://mithril.ai}
}

@article{flash3,
  author={Jay Shah and Ganesh Bikshandi and Ying Zhang and Vijay Thakkar and Pradeep Ramani and Tri Dao},
  title={FlashAttention-3: Fast and Accurate Attention with Asynchrony and Low-precision},
  journal={arXiv:2407.08608},
  year={2024},
}

@article{layernorm,
  author={Lei Jimmy Ba and Jamie Ryan Kiros and Geoffrey E. Hinton},
  title={Layer Normalization},
  journal={arXiv:1607.06450},
  year={2016},
}

@inproceedings{adamw,
  author={Ilya Loshchilov and Frank Hutter},
  title={Decoupled Weight Decay Regularization},
  booktitle={ICLR},
  year={2019},
}

@misc{flux2024,
    author={BlackForest},
    title={FLUX},
    year={2024},
    howpublished={\url{https://github.com/black-forest-labs/flux}},
}

@article{gao2025seedream,
  title={Seedream 3.0 technical report},
  author={Gao, Yu and Gong, Lixue and Guo, Qiushan and Hou, Xiaoxia and Lai, Zhichao and Li, Fanshi and Li, Liang and Lian, Xiaochen and Liao, Chao and Liu, Liyang and others},
  journal={arXiv preprint arXiv:2504.11346},
  year={2025}
}

@misc{rombach2021highresolution,
      title={High-Resolution Image Synthesis with Latent Diffusion Models}, 
      author={Robin Rombach and Andreas Blattmann and Dominik Lorenz and Patrick Esser and Björn Ommer},
      year={2021},
      eprint={2112.10752},
      archivePrefix={arXiv},
      primaryClass={cs.CV}
}

@inproceedings{chen2024pixartalpha,
  title={PixArt-$\alpha$: Fast Training of Diffusion Transformer for Photorealistic Text-to-Image Synthesis},
  author={Chen, Junsong and Yu, Jincheng and Ge, Chongjian and Yao, Lewei and Xie, Enze and Wang, Zhongdao and Kwok, James T and Luo, Ping and Lu, Huchuan and Li, Zhenguo},
  booktitle={ICLR},
  year={2024}
}

@article{podell2023sdxl,
  title={Sdxl: Improving latent diffusion models for high-resolution image synthesis},
  author={Podell, Dustin and English, Zion and Lacey, Kyle and Blattmann, Andreas and Dockhorn, Tim and M{\"u}ller, Jonas and Penna, Joe and Rombach, Robin},
  journal={arXiv preprint arXiv:2307.01952},
  year={2023}
}

@article{li2024playground,
  title={Playground v2. 5: Three insights towards enhancing aesthetic quality in text-to-image generation},
  author={Li, Daiqing and Kamko, Aleks and Akhgari, Ehsan and Sabet, Ali and Xu, Linmiao and Doshi, Suhail},
  journal={arXiv preprint arXiv:2402.17245},
  year={2024}
}

@inproceedings{wu2025janus,
  title={Janus: Decoupling visual encoding for unified multimodal understanding and generation},
  author={Wu, Chengyue and Chen, Xiaokang and Wu, Zhiyu and Ma, Yiyang and Liu, Xingchao and Pan, Zizheng and Liu, Wen and Xie, Zhenda and Yu, Xingkai and Ruan, Chong and others},
  booktitle={Proceedings of the Computer Vision and Pattern Recognition Conference},
  pages={12966--12977},
  year={2025}
}

@inproceedings{chen2024pixartsigma,
  title={Pixart-$\sigma$: Weak-to-strong training of diffusion transformer for 4k text-to-image generation},
  author={Chen, Junsong and Ge, Chongjian and Xie, Enze and Wu, Yue and Yao, Lewei and Ren, Xiaozhe and Wang, Zhongdao and Luo, Ping and Lu, Huchuan and Li, Zhenguo},
  booktitle={European Conference on Computer Vision},
  pages={74--91},
  year={2024},
  organization={Springer}
}

@article{wang2024emu3,
  title={Emu3: Next-token prediction is all you need},
  author={Wang, Xinlong and Zhang, Xiaosong and Luo, Zhengxiong and Sun, Quan and Cui, Yufeng and Wang, Jinsheng and Zhang, Fan and Wang, Yueze and Li, Zhen and Yu, Qiying and others},
  journal={arXiv preprint arXiv:2409.18869},
  year={2024}
}

@article{chen2025janus,
  title={Janus-pro: Unified multimodal understanding and generation with data and model scaling},
  author={Chen, Xiaokang and Wu, Zhiyu and Liu, Xingchao and Pan, Zizheng and Liu, Wen and Xie, Zhenda and Yu, Xingkai and Ruan, Chong},
  journal={arXiv preprint arXiv:2501.17811},
  year={2025}
}

@misc{openai2023dalle3,
  title        = {{DALL{\textperiodcentered}E 3}},
  author       = {{OpenAI}},
  year         = {2023},
  month        = sep,
  howpublished = {https://openai.com/research/dall-e-3},
}

@article{cai2025hidream,
  title={HiDream-I1: A High-Efficient Image Generative Foundation Model with Sparse Diffusion Transformer},
  author={Cai, Qi and Chen, Jingwen and Chen, Yang and Li, Yehao and Long, Fuchen and Pan, Yingwei and Qiu, Zhaofan and Zhang, Yiheng and Gao, Fengbin and Xu, Peihan and others},
  journal={arXiv preprint arXiv:2505.22705},
  year={2025}
}

@article{hu2024ella,
  title={Ella: Equip diffusion models with llm for enhanced semantic alignment},
  author={Hu, Xiwei and Wang, Rui and Fang, Yixiao and Fu, Bin and Cheng, Pei and Yu, Gang},
  journal={arXiv preprint arXiv:2403.05135},
  year={2024}
}

@article{ghosh2023geneval,
  title={Geneval: An object-focused framework for evaluating text-to-image alignment},
  author={Ghosh, Dhruba and Hajishirzi, Hannaneh and Schmidt, Ludwig},
  journal={Advances in Neural Information Processing Systems},
  volume={36},
  pages={52132--52152},
  year={2023}
}

@article{xie2024show,
  title={Show-o: One single transformer to unify multimodal understanding and generation},
  author={Xie, Jinheng and Mao, Weijia and Bai, Zechen and Zhang, David Junhao and Wang, Weihao and Lin, Kevin Qinghong and Gu, Yuchao and Chen, Zhijie and Yang, Zhenheng and Shou, Mike Zheng},
  journal={arXiv preprint arXiv:2408.12528},
  year={2024}
}

@inproceedings{ma2025janusflow,
  title={Janusflow: Harmonizing autoregression and rectified flow for unified multimodal understanding and generation},
  author={Ma, Yiyang and Liu, Xingchao and Chen, Xiaokang and Liu, Wen and Wu, Chengyue and Wu, Zhiyu and Pan, Zizheng and Xie, Zhenda and Zhang, Haowei and Yu, Xingkai and others},
  booktitle={Proceedings of the Computer Vision and Pattern Recognition Conference},
  pages={7739--7751},
  year={2025}
}

@article{deng2025bagel,
  title={Emerging properties in unified multimodal pretraining},
  author={Deng, Chaorui and Zhu, Deyao and Li, Kunchang and Gou, Chenhui and Li, Feng and Wang, Zeyu and Zhong, Shu and Yu, Weihao and Nie, Xiaonan and Song, Ziang and others},
  journal={arXiv preprint arXiv:2505.14683},
  year={2025}
}

@article{wu2025omnigen2,
  title={OmniGen2: Exploration to Advanced Multimodal Generation},
  author={Wu, Chenyuan and Zheng, Pengfei and Yan, Ruiran and Xiao, Shitao and Luo, Xin and Wang, Yueze and Li, Wanli and Jiang, Xiyan and Liu, Yexin and Zhou, Junjie and others},
  journal={arXiv preprint arXiv:2506.18871},
  year={2025}
}

@misc{Kolors2,
  author       = {Kuaishou Kolors team },
  title        = {Kolors2.0}, 
  howpublished = {https://app.klingai.com/cn/},
  year         = 2025,
}

@misc{gptimage,
  title={GPT-Image-1},
  author={OpenAI},
  year={2025},
  url={https://openai.com/index/introducing-4o-image-generation/},
}

@article{xie2025sana1d5,
  title={Sana 1.5: Efficient scaling of training-time and inference-time compute in linear diffusion transformer},
  author={Xie, Enze and Chen, Junsong and Zhao, Yuyang and Yu, Jincheng and Zhu, Ligeng and Wu, Chengyue and Lin, Yujun and Zhang, Zhekai and Li, Muyang and Chen, Junyu and others},
  journal={arXiv preprint arXiv:2501.18427},
  year={2025}
}

@article{zhuo2024luminanext,
  title={Lumina-next: Making lumina-t2x stronger and faster with next-dit},
  author={Zhuo, Le and Du, Ruoyi and Xiao, Han and Li, Yangguang and Liu, Dongyang and Huang, Rongjie and Liu, Wenze and Zhu, Xiangyang and Wang, Fu-Yun and Ma, Zhanyu and others},
  journal={Advances in Neural Information Processing Systems},
  volume={37},
  pages={131278--131315},
  year={2024}
}

@misc{recraftv3,
  author = {Recraft},
  title = {Recraft v3},
  year = {2024},
  howpublished = {\url{https://www.recraft.ai/}}
}

@misc{imagen4,
  title={Imagen},
  author={Google},
  year={2025},
  url={https://deepmind.google/models/imagen/}
}

@article{chang2025oneig,
  title={OneIG-Bench: Omni-dimensional Nuanced Evaluation for Image Generation},
  author={Chang, Jingjing and Fang, Yixiao and Xing, Peng and Wu, Shuhan and Cheng, Wei and Wang, Rui and Zeng, Xianfang and Yu, Gang and Chen, Hai-Bao},
  journal={arXiv preprint arXiv:2506.07977},
  year={2025}
}

@article{chen2025blip3,
  title={Blip3-o: A family of fully open unified multimodal models-architecture, training and dataset},
  author={Chen, Jiuhai and Xu, Zhiyang and Pan, Xichen and Hu, Yushi and Qin, Can and Goldstein, Tom and Huang, Lifu and Zhou, Tianyi and Xie, Saining and Savarese, Silvio and others},
  journal={arXiv preprint arXiv:2505.09568},
  year={2025}
}

@article{xie2025show,
  title={Show-o2: Improved Native Unified Multimodal Models},
  author={Xie, Jinheng and Yang, Zhenheng and Shou, Mike Zheng},
  journal={arXiv preprint arXiv:2506.15564},
  year={2025}
}

@misc{Cogview4,
    title = {Cogview4},
    author={THUKEG Z.ai},
    url = {https://github.com/THUDM/CogView4},
    year = {2025}
}

@misc{Imagen3,
    title = {Imagen 3},
    author={Imagen Team Google},
    year = {2024}
}

@misc{zoph2022stmoe,
  title         = {{ST-MoE}: Designing Stable and Transferable Sparse Expert Models},
  author        = {Barret Zoph and Irwan Bello and Sameer Kumar and Nan Du and
                   Yanping Huang and Jeff Dean and Noam Shazeer and William Fedus},
  year          = {2022},
  eprint        = {2202.08906},
  archivePrefix = {arXiv},
  primaryClass  = {cs.LG},
  url           = {https://arxiv.org/abs/2202.08906}
}

@misc{wang2025diffusion4k,
  title         = {Diffusion-4K: Ultra-High-Resolution Image Synthesis with Latent Diffusion Models},
  author        = {Jinjin Zhang and Qiuyu Huang and Junjie Liu and Xiefan Guo and Di Huang},
  year          = {2025},
  eprint        = {2503.18352},
  archivePrefix = {arXiv},
  primaryClass  = {cs.CV},
  url           = {https://arxiv.org/abs/2503.18352}
}

@misc{guo2025advancing,
  title         = {Advancing Expert Specialization for Better {MoE}},
  author        = {Hongcan Guo and Haolang Lu and Guoshun Nan and Bolun Chu and
                   Jialin Zhuang and Yuan Yang and Wenhao Che and Xinye Cao and
                   Sicong Leng and Qimei Cui and Xudong Jiang},
  year          = {2025},
  eprint        = {2505.22323},
  archivePrefix = {arXiv},
  primaryClass  = {cs.LG},
  url           = {https://arxiv.org/abs/2505.22323}
}

@misc{ernie2025report,
  title         = {{ERNIE} 4.5 Technical Report},
  author        = {{Baidu-ERNIE-Team}},
  year          = {2025},
  howpublished  = {\url{https://ernie.baidu.com/blog/publication/ERNIE_Technical_Report.pdf}},
  url           = {https://ernie.baidu.com/blog/publication/ERNIE_Technical_Report.pdf}
}

\end{document}